\documentclass[msom,nonblindrev]{informs3}
\OneAndAHalfSpacedXI
\usepackage{makecell}
\usepackage{multirow}
\usepackage{natbib}
 \bibpunct[, ]{(}{)}{,}{a}{}{,}%
 \def\bibfont{\small}%

\usepackage[utf8]{inputenc} % allow utf-8 input
\usepackage[T1]{fontenc}    % use 8-bit T1 fonts
\usepackage{hyperref}       % hyperlinks
\usepackage{url}            % simple URL typesetting
\usepackage{booktabs}       % professional-quality tables
\usepackage{amsfonts}       % blackboard math symbols
\usepackage{nicefrac}       % compact symbols for 1/2, etc.
\usepackage{microtype}      % microtypography
\usepackage{xcolor}   
\usepackage{footnote}% colors
\usepackage{tablefootnote}
\usepackage{floatrow}
\usepackage{algorithm}
\usepackage{algpseudocode}
\usepackage{amsmath,amssymb,amsfonts}
\usepackage{caption}

\usepackage{graphicx}
\usepackage{enumitem}

 \newcommand{\negin}[1]{\noindent{\textcolor{red}{\{{\bf NG:} #1\}}}}
  \newcommand{\jason}[1]{\noindent{\textcolor{blue}{\{{\bf JL:} #1\}}}}

\usepackage{algorithm}
\usepackage{algorithmicx}
\usepackage{algpseudocode}

\newtheorem{proposition}{Proposition}
\newtheorem{definition}{Definition}
\newtheorem{theorem}{Theorem}

\newtheorem{lemma}{Lemma}
\newtheorem{assumption}{Assumption}
\newtheorem{remark}{Remark}

\newcommand{\N}{\mathbb{N}}
\newcommand{\contex}{X}

\newcommand{\I}{\mathbb{I}}
\newcommand{\R}{\mathbb{R}}
\newcommand{\prob}{\mathbb{P}}
\newcommand{\expect}{\mathbb{E}}

\newcommand{\E}{\mathbb{E}}

\newcommand{\norm}[1]{\lVert#1\rVert}

\newcommand{\cond}[1]{\left. \right|}
\def \rev{{\sf{\small{rev}}}}

\def \reg{{\sf{\small{Regret}}}}

\def \ce {c_1}

\def \ct{c_2}

\def \cth {c_3}

\begin{document}
%%%%%%%%%%%%%%%%

\TITLE{Incentive-aware Contextual Pricing with Non-parametric Market Noise}
\ARTICLEAUTHORS{%
\AUTHOR{Negin Golrezaei}
\AFF{Sloan School of Management, Massachusetts Institute of Technology, \EMAIL{golrezae@mit.edu} \URL{}}
\AUTHOR{Patrick Jaillet}
\AFF{Department of Electrical Engineering and Computer Science, Massachusetts Institute of Technology, \EMAIL{jaillet@mit.edu} \URL{}}
\AUTHOR{Jason Cheuk Nam Liang}
\AFF{Operations Research Center, Massachusetts Institute of Technology, \EMAIL{jcnliang@mit.edu} \URL{}}
% Enter all authors
} % end of the block

\ABSTRACT{We consider a dynamic pricing problem for repeated contextual second-price auctions with multiple strategic buyers who aim to maximize their long-term time discounted utility. The seller has limited information on buyers' overall demand curves which depends on a non-parametric market-noise distribution, and buyers
may potentially submit corrupted bids (relative to true valuations) to manipulate the seller's pricing policy for more favorable reserve prices in the future. We focus on designing the seller's learning policy to set contextual reserve prices where the seller's goal is to minimize regret compared to the revenue of a benchmark clairvoyant policy that has full information of buyers' demand. We propose a policy with a phased-structure that incorporates randomized ``isolation" periods, during which a buyer is randomly chosen to solely participate in the auction. We show that this design allows the seller to control the number of periods in which  buyers significantly corrupt their bids. We then prove that our policy enjoys a $T$-period regret of  $\widetilde{\mathcal{O}}(\sqrt{T})$ facing strategic buyers. Finally, we conduct numerical simulations to compare our proposed algorithm to standard pricing policies. Our  numerical results show that our algorithm outperforms these policies under various buyer bidding behavior. 
}

\maketitle

\section{Introduction}

% The problem of dynamic pricing in repeated auctions has attracted great attention in recent years due to the booming activities in online advertisement markets
\iffalse
\jason{The massive amount of real-time data in online advertisement markets drive advertisers to bid only on impressions that they value. Such targeting behavior results in a non trivial proportion of markets exhibiting a significant degree of homogeneity, that is, bidders in such markets tend to have similar preferences for the impressions auctioned. Therefore, the problem to maximize publisher's revenue in these targeted markets can be analyzed through the reserve price setting mechanism for repeated second-price auctions facing multiple buyers with a common degree of willingness-to-pay.}
\jasoncom{missing some industry background}

\fi

We study the problem of  designing pricing policies for highly heterogeneous items against strategic agents. The motivation comes from the availability  of massive amount of real-time data in online platforms and in particular, online advertising markets, where the seller has access to detailed information about item features/contexts. In such environments, designing optimal policies involves learning buyers' demand (which is a mapping from item features and offered prices to the likelihood of the item being sold) under limited understanding of buyers' behavior. Our key goal is to develop effective and robust dynamic pricing polices that facilitate such a complex learning process for very general non-parametric contextual demand curves facing strategic buyers.
% The pricing policies should overcome the challenge of having limited information about buyers' willingness-to-pay, address the general learning problem for non-parametric demand curves, and resolve the issues that arise along with buyers' strategic behavior.

% Formally, we study the setting wherein any period $t$ over a finite time horizon $T$, the seller sells one item to buyers via running a second price auction with a reserve price.  The item is characterized by a $d$-dimensional context vector $x_t$, public to the seller and buyers. We consider an interdependent contextual valuation model in which the buyer's valuation for the item is the sum of common and private components. The common component, which is the same across all the buyers, is a function of the feature vector; and the private component, which captures buyers' idiosyncratic preferences, is independently sampled from an unknown \emph{non-parametric} noise distribution $F$. The common component determines the expected willingness-to-pay of the buyers and is the inner product of the feature vector and a fixed scaling factor $\beta$, which we call the ``mean vector". We note that such a linear valuation model is very common in the literature of dynamic pricing; e.g. see \cite{golrezaei2018dynamic, javanmard2016dynamic,kanoria2017dynamic} and \cite{javanmard2017perishability}. 

Formally, we study the setting wherein any period $t$ over a finite time horizon $T$, the seller sells one item to buyers via running a second price auction with a reserve price.  The item is characterized by a $d$-dimensional context vector $x_t$, public to the seller and buyers. We consider an interdependent contextual valuation model in which a buyer's valuation for the item is the sum of common and private components. The common component determines the expected willingness-to-pay of buyers and is the inner product of the feature vector and a fixed ``mean vector" $\beta$ that is homogeneous across buyers; the private component, which captures buyers' idiosyncratic preferences, is independently sampled from an unknown \emph{non-parametric} noise distribution $F$. We note that such a linear valuation model is very common in the literature of dynamic pricing; e.g. see \cite{golrezaei2018dynamic, javanmard2016dynamic,kanoria2017dynamic} and \cite{javanmard2017perishability}. 

Under this interdependent contextual valuation model, we study a \emph{strategic setting} where buyers intend to maximize  long-term discounted utility and may
consequently  submit \emph{corrupted}, i.e., untruthful, bids. The motivation of this strategic setting comes from the repeated buyer-seller interactions
when the seller does not possess full information on buyers' demand and aims to learn it using buyers' submitted bids. In a single-shot second price auction, where there is no repeated interactions between the seller and buyers,
bidding truthfully is a buyer's weakly dominant action. However, this is no longer the case in our repeated second price auction setting: repeated auctions may incentivize the buyers to submit corrupted bids, rather than their true valuations, in order to manipulate seller's future reserve prices; e.g. by underbidding, buyers may trick the seller to lower future reserve prices. 

%provide buyers with the opportunity to take advantage of the seller's lack of knowledge in buyers' valuations by submitting corrupted bids instead of their true valuations. In other words, since the seller does not know buyers' demand curves and aims to learn them using submitted bids, buyers are incentivized to leverage their private information and ``game the system'' by submitting (corrupted) bids in a strategic manner so that they can  manipulate future reserve prices.

% It is a well-known result that during a single-shot second price auction, bidding truthfully is a buyer's weakly dominant action. However, this salient feature no longer exists when the second price auctions are run repeatedly over time. This is because the repeated seller-buyer interactions provide buyers with the opportunity to take advantage of the seller's lack of knowledge in buyers' valuations by submitting corrupted bids instead of their true valuations: since the seller does not know buyers' demand curves and aims to learn them using submitted bids, buyers are incentivized to leverage their private information and ``game the system'' by submitting (corrupted) bids in an untruthful and strategic manner so that they can  manipulate future reserve prices.

In this work, we would like to design a reserve price policy for the seller  who does not know the mean vector $\beta$ and the  noise distribution $F$. The policy  
 dynamically learns/optimizes contextual reserve prices while being robust to  corrupted data (bids), submitted by strategic buyers.
In particular, our objective is to
minimize our policy's regret computed  against a clairvoyant benchmark policy that knows both $\beta $ and $F$. Designing low-regret policies in our setting  involves overcoming the following challenges: {(i) The demand curve is constantly shifting due to the change in contexts over time. (ii) The shape of the demand curve is unknown due to the lack of information on the market  noise  distribution $F$ which  may  not  enjoy  a  parametric  functional  form. Furthermore, we do not impose the \textit{Monotone Hazard Rate} (MHR)\footnote{Distribution $F$ is MHR if $\frac{f(z)}{1-F(z)}$ is non-decreasing in $z$, where $f$ is the corresponding pdf.} assumption on $F$.  While the MHR assumption  is common  in the related literature and can significantly simplify reserve price optimization (see e.g. Remark \ref{remark:MHR}), it has been shown to fail in  practice (see \cite{celis2014buy,golrezaei2017boosted}). (iii) As stated earlier, in our strategic setting, buyers may take advantage of the seller's lack of knowledge about  buyers' demand and   submit corrupted bids to manipulate future reserve prices.}

\textbf{Main contribution.} We develop a policy called \emph{Non-Parametric Contextual Policy against Strategic Buyers} (NPAC-S) that enables the seller to efficiently learn the optimal contextual reserve prices while being
robust against buyers' corrupted bids. Our policy design incorporates two simple yet effective features, namely a \emph{phased structure} and \emph{random isolation}.
% First, NPAC-S  partitions the entire time horizon into phases of increasing length, and estimates the mean vector and the distributions of the second-highest and highest valuations only using data from the previous phase. 
First, NPAC-S partitions the entire horizon into consecutive phases, and then
estimates the mean vector and the distributions of the second-highest and highest valuations only using data from the previous phase. This reduces the buyers' manipulating power on future reserve prices as past corrupted bids prior to the previous phase will not affect future pricing decisions. Second, the NPAC-S policy incorporates randomized isolation periods, that is, in each period with some probability the seller chooses a particular buyer at random and let her be the single participant of the auction during this period. In these isolation periods, the isolated buyer faces no competition from other buyers, and hence may incur large utility loss if a significantly corrupted bid is submitted.\footnote{In the isolation periods, when the valuation of the isolated buyer is greater than the reserve price, significantly underbidding may cause the item to not be allocated; when the valuation of the isolated buyer is lower than the reserve price, overbidding results in the buyer paying much higher prices (relative to valuation) to achieve the item. In either case, the isolated buyer will incur a significant utility loss compared to truthful bidding. }
%  \footnote{Buyers are aware of  the randomized isolation periods as the seller announces and commits to his pricing policy.}

For our main theoretical results, we show that 
in virtue of our isolation periods in our design of NPAC-S, the number of past periods with large corruptions is $\mathcal{O}(\log(t))$ for any period $t$ via leveraging the fact that buyers aim to maximize their long-term discounted utility. Furthermore, we present novel high probability bounds for our estimation errors in $\beta$ and $F$ which are estimated by ordinary least squares and empirical distributions, respectively, with the presence of corrupted bids. Finally, in Theorem \ref{thm:strat:Regret}, we show that the NPAC-S policy achieves a regret of $\tilde{\mathcal{O}}(d\sqrt{T})$ for general non-parametric distributions $F$ against a clairvoyant benchmark policy.

% The NPAC-S policy can still use OLS estimators and empirical distributions because, in virtue of our isolation periods, the number of past periods with large corruptions is $\mathcal{O}(\log(t))$ for any period $t$ with high probability.  Put differently, the number of outliers in the submitted bids is small and as a result, there is no need to redesign the estimation techniques used in NPAC-T for the strategic setting. \if false Moreover,  the isolation periods in the NPAC-S policy refrains  uncoordinated collusive behaviour between buyers to collectively shade their bids and drive the reserve prices to zero. \fi 

% \if false With randomized isolation periods in our design, we can break such (uncoordinated) collusive behaviour because  in these periods only a single buyer participates in the auction with a random reserve price. 

% In Theorem \ref{thm:strat:Regret}, we show that the NPAC-S policy achieves a regret of $\tilde{\mathcal{O}}\left(d\sqrt{T}\right)$ for general non-parametric distributions $F$. 

\textbf{Related literature.} 
Here we discuss related works that study dynamic pricing against strategic buyers with stochastic valuations, \footnote{The general theme of learning in the presence of strategic agents or corrupted information has also been studied in other applications; see, for example, \cite{chen2018markdown, birge2018dynamic, feng2019intrinsic}. There are also related works that study adversarial buyer valuations. For example, \cite{drutsa2019reserve} studies the seller's pricing problem for repeated second-price auctions facing multiple strategic buyers with private valuations fixed overtime. In addition, buyers in this work also seek to maximize cumulative discounted utility. The paper proposes an algorithm that achieves  $\mathcal{O}(\log\log(T))$ regret for worst-case (adversarial) valuations.} and refer readers to Appendix \ref{app:extendreview} for broader related works.

Both \cite{amin2013learning,amin2014repeated} study a dynamic pricing problem in a posted price auction against a single strategic buyer. \cite{amin2013learning} addresses the non-contextual stochastic valuation setting, where as \cite{amin2014repeated} studies a linear contextual valuation model, but 
with no market noise disturbance. \cite{amin2014repeated} proposes an  algorithm that achieves $\widetilde{\mathcal{O}}(T^{{2}/{3}})$ regret in contrast with our regret of $\widetilde{\mathcal{O}}(\sqrt{T})$ using the NPAC-S policy. We point out that this is because the seller in their setting only observes the outcome of the auction (i.e. bandit feedback), while in our setting we assume that seller can examine all submitted bids. Our setting is more complex compared to \cite{amin2013learning,amin2014repeated} as we handle the contextual pricing problem against multiple strategic buyers, and also deals with the issue of learning a non-parametric distribution function in the presence of strategic buyer behavior. {\cite{kanoria2017dynamic} consider a contextual buyer valuation model similar to ours (but with the MHR assumption on the market noise distribution) and proposes a pricing algorithm that sets personalized reserve prices for individual buyers. They argue that the design of their algorithm induces an equilibrium where buyers always bid truthfully, and then further assume buyers act according to this equilibrium. Our work distinguishes itself from two aspects. First, setting personalized reserve prices in \cite{kanoria2017dynamic} rely crucially on the MHR assumption, and in this paper we relax this assumption such that our methodology works for a larger class of market noise distributions. Second, we consider more general buyers who do not necessarily play any equilibrium and are forward looking.
}
\cite{golrezaei2018dynamic} study a similar interdependent contextual valuation model to ours, but with heterogeneous mean vector $\beta$ across agents.  Our work distinguishes itself from \cite{golrezaei2018dynamic} in two major ways. First, they focus on optimizing contextual reserve w.r.t. the worst-case distribution among a known class of MHR market noise distributions. In contrast, our work relaxes this constraint and does not require the seller to have any prior knowledge on the possibly non-parametric distribution.  Second, in their setting, the seller only utilizes the outcome of the auctions to learn buyer demand and results in a regret of $\widetilde{\mathcal{O}}(T^{{2}/{3}})$.\footnote{A recent work  \cite{deng2019robust} builds on the  result of \cite{golrezaei2018dynamic} by considering a stronger benchmark that knows future 
buyer valuation distributions (noise distribution and all the future  contextual information). They 
 design robust pricing schemes whose regret is $\mathcal{O}(T^{5/6})$  against the aforementioned benchmark, confirming the generalizability of pricing schemes in \cite{golrezaei2018dynamic}.} In our work, we exploit the information of all submitted bids by taking advantage of the  fact that buyers'  utility-maximizing behaviour constrains their degree of corruption on bids. This  eventually allows us to achieve an improved regret of $\widetilde{\mathcal{O}}(\sqrt{T})$. Nevertheless, our proposed algorithm cannot not handle heterogeneous $\beta$'s, and hence this will be an interesting future research direction. \cite{drutsa2020optimal} studies the posted price selling problem against a strategic agent with a non-linear (stochastic) contextual valuation model that satisfies some Lipschitz condition with no additive noise.
 
We summarize some key differences in the settings/results of the aforementioned works in Table \ref{table}. 

\begin{table*}[tbh]\footnotesize
\centering
\begin{tabular}{ccccccc}
     \toprule
Algorithm & \# buyers & Context & Noise/value dist. & Discount util.  & Regret \\
 \midrule
 Phased \cite{amin2013learning} & 1 & False & Lipschitz & True & Sublinear\tablefootnote{The Phased algorithm in \cite{amin2013learning} incurs a regret of $\mathcal{O}(T^{\alpha} + {1}/(1-\eta)^{1/\alpha})$ for any chosen $\alpha \in (0,1)$, where $\eta$ is the buyer-utility discount factor.}\\
 LEAP \cite{amin2014repeated} & 1 & True & No additive noise & True & $\mathcal{O}(T^{2/3})$\\
 PELS \cite{drutsa2020optimal}& $ 1 $& True &  No additive noise& True &$\mathcal{O}(T^{d/(d+1)})$\\
  HO-SERP \cite{kanoria2021incentive} & $\geq 2$ &  True &MHR & False & $\mathcal{O}(\sqrt{T})$\\
 SCORP \cite{golrezaei2018dynamic} & $\geq 2$ & True &MHR & True & $\mathcal{O}(T^{2/3})$ \\
NPAC-S (this work) & $\geq 2$ & True & Non-parametric & True &$\mathcal{O}(\sqrt{T})$\\
     \bottomrule
\end{tabular}
\vspace{0.1cm}
\caption{\footnotesize Summary of settings and results for seller algorithms that sell against strategic agents with stochastic valuations. Note that the Discount util. column indicates whether the algorithm deals with buyers who discount their long-term utilities. Note that HO-SERP\cite{kanoria2021incentive} and  SCORP \cite{golrezaei2018dynamic} set \emph{personalized reserve prices} for each buyer, whereas NPAC-S sets a single reserve for all buyers. PELS in \cite{drutsa2020optimal} learns a non-linear contextual valuation model and hence yields larger regret. Among all algorithms, only SCORP \cite{golrezaei2018dynamic} handles heterogeneous $\beta$ across buyers.}
\label{table}
\end{table*}

\section{Preliminaries} \label{Sec:model}
\textbf{Notation.} For $a\in \N^+$, denote $[a] = \{1,2,\dots, a\}$. %For any $z\in \R$, we denote $(z)^+ = \max\{0,z\}$. For a vector $x\in \R^d$, we denote $\norm{x}_1$,$\norm{x}_2$, and $\norm{x}_\infty$ as its $\ell_1$-norm, $\ell_2$-norm, and $\ell_\infty$-norm  respectively, and 
For two vectors $x,y\in \R^d$, denote $\langle x, y\rangle$ as their inner product. Finally, $\I\{\cdot\}$ is the indicator function: $\I\{\mathcal{A}\} =1$ if event $\mathcal{A}$ occurs and $0$ otherwise.

We consider a seller who runs  repeated second price auctions over a horizon with length $T$ that is unknown to the seller. In each auction $t\in [T]$, an item is sold to $N$ buyers, where the item is characterized by a $d$-dimensional   feature vector $x_t \in \mathcal{X} \subset \{x \in \R^d: \norm{x}_\infty \leq x_{\max} \}$ where $0< x_{\max}<\infty$.  We assume that $x_t$ is independently drawn from some distribution $\mathcal{D}$ unknown to the seller.  We define $\Sigma $ as the covariance matrix of distribution $\mathcal{D}$.\footnote{{The covariance matrix of a distribution $\mathcal{P}$ on $\R^d$ is defined as $\expect_{x\sim\mathcal{P}}[x x^\top] - \mu\mu^\top$, where $\mu = \expect_{x\sim \mathcal{P}}[x]$}. } We assume that $\Sigma$ is positive definite and unknown to the seller, and define the smallest eigenvalue  of $\Sigma$ to be $\lambda_0 > 0$.

%  The term $\langle \beta, x_t\rangle$ is the common value component that represents  the expected (mean) buyers' valuation and  $\epsilon_{i,t}$ is some mean zero market noise that captures the idiosyncratic tastes of buyers.
\textbf{Buyer valuation model.} We focus on an interdependent valuation model where the
valuation of  buyer
$i\in [N]$ at time $t\in [T]$% is the sum of the buyers’ common contextual willingness-to-pay, and their idiosyncratic taste. That is is, buyer's valuation for the item at time with a $d$-dimensional feature vector $x_t$ 
is given by  
$v_{i, t} =\langle \beta, x_t\rangle  + \epsilon_{i,t}$. Here, 
  $\beta$ is called the \textit{mean vector} and is fixed over time and unknown to the seller, while $\epsilon_{i,t}$ is idiosyncratic market noise sampled independently over time and across buyers from some time-invariant distribution $F$ with probability density function $f$, both unknown to the seller. Furthermore, $F$ has bounded support $ (-\epsilon_{\max}, \epsilon_{\max})$, in which its probability density function is bounded by $c_f  := \sup_{z \in [-\epsilon_{\max}, \epsilon_{\max}]} f(z) \geq \inf_{z \in [-\epsilon_{\max}, \epsilon_{\max}]} f(z) > 0$.  The support boundary $\epsilon_{\max}$ is not necessarily known to the seller. We assume there exist $v_{\max} > 0$ so that
  $v_{i, t}\in [0, v_{\max}]$
  for all $i\in [N]$, $t\in [T]$. 
 %\pj{I think that we need to say something about how restrictive and/or realistic this assumption is}.

% \begin{assumption} \label{assum:F}
% Market noise variables $\{\epsilon_{i,t}\}_{t\in [T], i\in[N]}$ have zero-mean and are drawn independently from the distribution $F$. Furthermore, $F$ has bounded support $ (-\epsilon_{\max}, \epsilon_{\max})$, in which pdf $f$ exists and is bounded by $c_f  := \sup_{z \in [-\epsilon_{\max}, \epsilon_{\max}]} f(z) \geq \inf_{z \in [-\epsilon_{\max}, \epsilon_{\max}]} f(z) > 0$. 
% \end{assumption}
% \negin{We don't need to use the ``Assumption" environment. We can say these in text. }

We highlight that our setting  does not enforce distribution $F$ to be parametric nor to satisfy the MHR assumption. This is because via analyzing real auction data sets, it has been shown that the MHR assumption does not necessarily hold in online advertising markets \cite{celis2014buy,golrezaei2017boosted}.

% We focus on this setting because it has been shown that the MHR assumption does not necessarily hold in real-world online advertising markets \citep{celis2014buy,golrezaei2017boosted}.  

%We assume that the common value component is always positive and bounded by some finite value $c>0$ unknown to the seller, i.e., $ {\color{blue}0< \langle x, \beta \rangle \leq c}$ for $\forall x\in \mathcal{D}$. 

%We highlight that distribution $F$ is not necessarily parametric. We only assume that distribution $F$ is sub-Gaussian. \negin{maybe } 
%, and the smallest eigenvalue $\lambda_0^2$ of the covariance matrix $\Sigma = \mathrm{Var}(x)$  for $x\sim \mathcal{D}$ is positive. 

%The parameters $\beta$ and $F$ remain constant over the entire time horizon $T$, but are unknown to the seller.
%\negin{Add a discussion why this is a difficult setting. We may remove this discussion to the introduction though.}

\textbf{Repeated contextual second price auctions with reserve.} The contextual second price auction with reserve is described as followed for $N\geq 2$ buyers: In any period $t\ge 1$,
a context vector $x_t \sim \mathcal D$ is revealed to the seller and buyers. The seller then computes reserve price $r_t$, while simultaneously each buyer $i\in [N]$ forms individual valuations $v_{i,t}$ and submits a bid $b_{i,t}$ to the seller.  Let $i^{\star} = \arg\max_{i \in [N]} b_{i, t}$ be the buyer who submitted the highest bid.\footnote{No ties will occur since we assume that no valuations and bids are the same.} If  $b_{i^{\star}, t}\ge r_{t}$, the item is allocated to buyer $i^{\star}$ and he is charged the maximum between the reserve price and second highest bid, i.e. buyer $i^{\star}$ pays $p_{i^{\star},t} = \max\{r_t, \max_{i\ne i^{\star}} b_{i, t}\}$. For any other buyer $i\ne  i^{\star}$, the payment $p_{i,t}=  0$. In the case where  $b_{i^{\star}, t}< r_{t}$, the item is not allocated and all payments are zero.

% \begin{enumerate}[leftmargin=0.8cm]
%     \item [(i)] The seller observes the context vector $x_t \sim \mathcal D$ and reveals it to the buyers. The seller then computes a reserve price $r_t$. 
%     \item [(ii)] Each buyer $i\in [N]$,  forms individual valuations $v_{i,t}$ and submit a bid $b_{i,t}$ to the seller. 
%     \item [(iii)] Let $i^{\star} = \arg\max_{i \in [N]} b_{i, t}$.\footnote{No ties will occur since we assume that no valuations and bids are the same.} If  $b_{i^{\star}, t}\ge r_{t}$, the item is allocated to buyer $i^{\star}$, i.e., the buyer with the  highest submitted bid, and he is charged the maximum between the reserve price and second highest bid; that is, the winner $i^{\star}$ pays $p_{i^{\star},t} = \max\{r_t, \max_{i\ne i^{\star}} b_{i, t}\}$. For any other buyer $i\ne  i^{\star}$, the payment $p_{i,t}=  0$. Note that if  $b_{i^{\star}, t}< r_{t}$, the item is not allocated and all the payments are zero.
%     % \item The seller observes all the bids, i.e., $v_{i, t}$, $i\in [N]$.
% \end{enumerate}

Here, the seller's reserve price $r_t$ can only depend on $x_t$ and the history set $\mathcal{H}_{t-1}:=\{(r_1, \{b_{i,1}\}_{i \in [N]},x_1), \ldots, (r_{t-1}, \{b_{i,t-1}\}_{i \in [N]},x_{t-1})\}$ which includes all information available to the seller up to period $t-1$.
% \begin{align*}
%     \mathcal{H}_{t-1}:=\{(r_1, b_1,x_1), (r_2, b_2,x_2), \ldots, (r_{t-1}, b_{t-1},x_{t-1})\}\,,
% \end{align*}

% For the special case when there is only $N=1$ buyer, the auction  only differs in step (iv) of the multiple-buyer mechanism: the item is allocated to the  buyer if her bid $b_t$ is greater than or equal to the reserve $r_t$, and her payment is $r_t$ if the item is allocated and $0$ otherwise. In this paper, we mainly focus on the case of $N\geq 2$ buyers as analyzing the multi-buyer case is more challenging, especially when bidders are strategic. Yet we point out that our proposed algorithms can be easily generalized to the single-buyer second price auction.

\textbf{Buyers' bidding behavior.}  In the setting where buyers are strategic, we assume that in any period $t$, each buyer $i\in [N]$ aims at maximizing his long-term discounted utility $U_{i,t}$:
  \begin{align}
     U_{i, t} ~ := ~ \sum_{\tau = t}^T \eta^{\tau} \expect\left[v_{i,\tau} w_{i,\tau} - p_{i,\tau} \right]\label{eq:strat:utility}\,,
  \end{align}
 where $\eta \in (0,1)$  is the discount factor, $w_{i,t} \in \{0,1\}$ indicates whether buyer $i$ wins the item, and the expectation is taken with respect to the randomness due to the noise distribution $F$, the context distribution $\mathcal{D}$, and buyers' bidding behavior. We point out that this discounted utility model illustrates the fact that buyers are less patient than the seller, and is a common framework in many dynamic pricing literature; see \cite{amin2013learning,amin2014repeated,golrezaei2018dynamic}, and \cite{liu2018learning}. The motivation lies in many applications in online advertisement markets wherein the user traffic is usually very uncertain and as a result, advertisers (buyers) would not like to miss out an opportunity of showing their ads to  targeted users. It is worth noting that \cite{amin2013learning} showed, in the case of a single buyer, it is not possible to obtain a no-regret policy when $\eta = 1$, that is, when the buyer is as patient as the seller.
 
% Additionally, buyers may from time to time submit ``corrupted bids'', i.e., underbid (shade their bid) or overbid with respect to their true valuations, sacrificing current utility   with the aim to lower future reserve prices and  increase their future long-term utility. We assume that seller announces his pricing policy to all buyers so that buyers  have full knowledge of the seller's learning and pricing algorithm and has the freedom to adopt any bidding strategy to maximize their long term utility.

% We now describe  the scope of feasible buyer bidding behavior in the strategic setting. Recall that the maximum possible valuation $v_{\max}$ is known to both buyers and the seller. Thus, buyers have no incentive to submit a bid greater than $v_{\max}$, i.e., $b_{i,t} \leq v_{\max}$ for all $i\in[N], t\in[T]$ as the seller  only sets reserve prices less than or equal to $v_{\max}$. 
% Furthermore, 

Furthermore, we assume buyers corrupt their true valuations in an additive manner:
\begin{align*}
 \forall i\in[N], t\in[T] \quad b_{i,t} =  v_{i,t} - a_{i,t} ~ \text{ where } ~  |a_{i,t}| \leq a_{\max}\,.
\end{align*}
Here, $a_{i,t}$ is called the degree of corruption, and we refer to the buyer behavior of submitting a bid $b_{i,t} \neq v_{i,t}$ (i.e., $a_{i,t}\neq 0$) as ``corrupted bidding''. 
Note that when $a_{i,t} >0$, the buyer shades her bid, and when $a_{i,t} < 0$, the buyer overbids. 
% Since the seller observes buyers' bids instead of their true valuations in the strategic setting, corrupted bids may deteriorate the estimation accuracy of buyers' demand, and as a result negatively impact pricing decisions in future periods.
Essentially, a buyer $i$'s strategic behavior is equivalent to deciding on a non-zero value of $a_{i,t}$.  In this work, we impose no restrictions on the degree of corruption $a_{i,t}$ for a buyer $i$ in period $t$ other than it is bounded. \footnote{A bound for the degree of corruption is natural as buyers always submit non negative bids and all bids are bounded by $v_{\max}$.}

\section{Benchmark and Seller's Regret} \label{Sec:benchmark}

The seller's revenue in period $t\in[T]$ is the sum of total payments 
from all buyers, and the expected revenue given context $x_t \in \mathcal{X}$ and reserve price $r_t$ is
\begin{align}
\label{eq:revenue}
\begin{aligned}
    & \rev_t(r_t) :=\E\Big[\sum_{i\in [N]} p_{i,t}~ \Big| ~ x_t, r_t \Big],\\
   & \text{where } ~ p_{i,t} ~ = ~
  \max\{b_t^-, r_t \}\I\{ b_{i,t} \geq \max\{b_t^+, r_t\}\}\,.
 \end{aligned}
\end{align}
Here, $b_t^-$ and $b_t^+$ are the second-highest and highest bids in period $t$, respectively; the expectation is taken with respect to the noise distribution in period $t$ and any randomness in the reserve price $r_t$ as well as bid values submitted by buyers in period $t$ (as buyers' bidding strategies may be randomized). 

% When the number of buyers $N= 1$, we set $b_t^-$ to zero.
 
%  We point out that in this paper, although we focus on the setting where the seller faces $N \geq 2$ buyers, we will also address the payment rule in the case where there is only $1$ buyer in a later section. \negin{Please either move this to the previous section or remove it. }

The seller's objective is to maximize his expected revenue over a fixed time horizon $T$ through optimizing contextual reserve prices $r_t$ for any $t\in [T]$. To evaluate any seller pricing policy, we compare its total revenue against that of a benchmark policy run by a clairvoyant seller who knows the mean vector $\beta$ and the non-parametric noise distribution $F$. 
This clairvoyant seller's benchmark policy  sets the ``optimal'' contextual reserve price in each period to obtain the maximum achievable revenue $\max_{r} \rev_{t}(r)$ in each period, and hence facing such a seller there will be no incentive for buyers' to corrupt their bids. To provide a more formal  definition for the revenue of the clairvoyant seller as well as  ``optimality'' in contextual reserve prices,  we rely on the following proposition that characterizes the seller's conditional expected revenue when buyers bid truthfully.
\begin{proposition}[Seller's Revenue with Truthful Buyers]\label{prop:benchmark}
Consider the case of $N\geq 2$ buyers who bid their true valuations, i.e., $v_{i,t} = b_{i,t}$ for any $ i\in[N]$ and $t\in[T]$. 
Conditioned on the reserve price $r_t$ and the current context $x_t \in \mathcal{X}$, the seller's single period expected revenue in Equation (\ref{eq:revenue}) is 
\begin{align}
\begin{aligned}
  \int_{-\infty}^{\infty} z dF^-(z) + \langle \beta,x_t \rangle  +\int_{0}^{r_{t}} F^{-}(z- \langle \beta,x_t \rangle )dz  - r_{t} \left( F^{+}(r_{t}- \langle \beta,x_t \rangle)\right) \,, 
\end{aligned}
\end{align}
where for any ${z} \in \R$, $F^-({z}) := NF^{N-1}({z}) - (N-1)F^N({z})$ and $F^+({z}) := F^N({z})$.  
\end{proposition}
The proof for this proposition is detailed in Appendix \ref{appsec:prop}.
In Proposition \ref{prop:benchmark}, $F^+(\cdot)$ and $F^-(\cdot)$ are the cumulative distribution functions of $\epsilon_t^+ :=v_t^+ -\langle \beta, x_t\rangle $ and $\epsilon_t^- :=v_t^- -\langle \beta, x_t\rangle $ respectively, where $v_t^+$ and $v_t^-$ are the highest and second highest valuations in period $t\in [T]$. 
% That is, $\epsilon_t^+$ and $\epsilon_t^-$ are the $N^{th}$ and $(N-1)^{th}$ order statistics of $N$ independent random samples from  distribution $F$. 

In light of Proposition \ref{prop:benchmark}, we define the benchmark policy of the clairvoyant seller as followed,
\begin{definition}[Benchmark Policy]\label{def:benchmark_policy}
The benchmark policy knows the mean vector $\beta$ and noise distribution $F$, and sets the reserve price for a context vector $x\in \mathcal{X}$ as
\begin{align}
\begin{aligned}
\label{eq:opt_reserve}
    r^{\star}(x) 
  ~ =~  \arg\max_{y\ge 0} 
    \int_{0}^y F^{-}(z- \langle \beta,x \rangle )dz - y \left( F^{+}(y- \langle \beta,x \rangle)\right)\,. 
    \end{aligned}
\end{align}
Therefore, the benchmark reserve price in period $t$, denoted by $r_{t}^{\star}$, is $r^{\star} (x_t)$, and the corresponding optimal revenue, denoted by $\text{REV}^\star_t$, is equal to
\begin{align*}
     \int_{-\infty}^{\infty} z dF^-(z) + \langle \beta,x_t \rangle  
  +\int_{0}^{r^{\star} (x_t)} F^{-}(z- \langle \beta,x_t \rangle )dz  - r^{\star} (x_t) \left( F^{+}(r^{\star} (x_t)- \langle \beta,x_t \rangle)\right)\,.
\end{align*}
% \[\text{REV}^\star_t = \int_{-\infty}^{\infty} z dF^-(z) + \langle \beta,x_t \rangle  +\int_{0}^{r^{\star} (x_t)} F^{-}(z- \langle \beta,x_t \rangle )dz - r^{\star} (x_t) \left( F^{+}(r^{\star} (x_t)- \langle \beta,x_t \rangle)\right)\,. \]
\end{definition}

\begin{remark}
\label{remark:MHR}
When distribution $F$ satisfies the MHR assumption, the objective function of the optimization problem in Equation (\ref{eq:opt_reserve}) is unimodal in the decision variable $y$, and according to \cite{golrezaei2018dynamic}, $r^{\star}(x)$ can be simplified as follows: 
$r^{\star}(x) = \arg\max_{y\ge 0}  y (1-F(y -\langle \beta,x \rangle))$. In words, the MHR assumption decouples the reserve price optimization problem for multiple agents to the much simpler monopolistic pricing for each individual agent.
\end{remark}

We observe this benchmark  provides an optimal mapping from the feature vector $x_t$ to reserve price $r^{\star}(x_t)$, which remains unchanged over time as the mean vector $\beta$ and noise distribution $F$ are time-invariant.
This echoes our earlier point that pricing is challenging in our contextual setting since we would need to approximate or learn the optimal mapping $r^{\star}(\cdot)$,  whereas in non-contextual environments it is sufficient to learn a single optimal reserve price value.

% We observe this benchmark sets the reserve price that maximizes the expected revenue under truthful buyer behavior (Proposition \ref{prop:benchmark}). In fact, the benchmark provides an optimal mapping from the feature vector $x_t$ to reserve price $r^{\star}(x_t)$, which remains unchanged over time as the mean vector $\beta$ and noise distribution $F$ are time-invariant.
% This echoes our earlier point that pricing is challenging in our contextual setting since we would need to approximate or learn the optimal mapping $r^{\star}(\cdot)$,  whereas in non-contextual environments it is sufficient to learn a single optimal reserve price value. 

 We now proceed to define the regret of a policy $\pi$ (possibly random) when the regret is measured against the benchmark policy.   Suppose that in any period $t$, policy $\pi$ selects reserve price $r_t^{\pi}$.
%  , where $r_t^{\pi}$ is a function of the context vector $x_t$ and may or may not depend on the history $\mathcal{H}_{t-1}$. 
%  Then, the revenue of policy $\pi$ in period $t$ is \[\Rev^{\pi}_t =\E[\max \{r_t^{\pi}, b_t^{-}\}\cdot \I\{b_t^+ \geq r_t^{\pi} \}]\,,\]
% where the expectation is w.r.t. the noise distribution $F$ , feature distribution $\mathcal D$, and buyers valuations.  
Then, the regret of policy $\pi$ in period $t$ and its cumulative $T$-period regret are defined as:
% \begin{align}
% \label{eq:regret}
%    \reg_t^\pi  = \expect\left[\text{REV}^{\star}_t - \rev_t(r_t^\pi)\right]~~~\text{and}~~~ \reg^\pi(T) = \sum_{t\in [T]}\reg_t^\pi\,,
% \end{align}
\begin{align}
\label{eq:regret}
    \reg^\pi(T) = \sum_{t\in [T]}\expect\left[\text{REV}^{\star}_t - \rev_t(r_t^\pi)\right]\,,
\end{align}
where the optimal revenue $\text{REV}^{\star}_t$ is given in Definition \ref{def:benchmark_policy}, and the expectation is taken with respect to the context distribution $\mathcal{D}$ as well as the possible randomness in the actual reserve price $r_t^\pi$. Our goal is to design a policy that obtains a low regret for any $\beta$, $F$, and context distribution $\mathcal{D}$.

\section{The NPAC-S Policy}
\label{Sec:strat}

% In this section, we develop a seller strategy that aims to extract as much revenue as possible from all buyers, and equivalently minimize cumulative regret against the clairvoyant benchmark described in Definition \ref{def:benchmark_policy} that sets the optimal contextual reserve price defined in Equation (\ref{eq:opt_reserve}).

% It is a well-known result that during a single-shot second price auction, bidding truthfully is a buyer's weakly dominant action. However, this salient feature no longer exists when the second price auctions are run repeatedly over time. This is because the repeated seller-buyer interactions provide buyers with the opportunity to take advantage of the seller's lack of knowledge in buyers' valuations by submitting corrupted bids instead of their true valuations: since the seller does not know buyers' demand curves and aims to learn them using submitted bids, buyers are incentivized to leverage their private information and ``game the system'' by submitting (corrupted) bids in an untruthful and strategic manner so that they can  manipulate future reserve prices.

In this section, we first propose a policy called \emph{Non-Parametric Contextual Policy against Strategic Buyers} (NPAC-S) to maximize seller’s expected revenue in our strategic setting. Then, we provide insights into how our design in NPAC-S 
makes the policy robust to buyer strategic behavior, and in turn allows the policy to learn the mean vector $\beta$ and noise distribution $F$ efficiently.  Finally, we present theoretical regret guarantees for NPAC-S  against the clairvoyant benchmark described in Definition \ref{def:benchmark_policy} that sets the optimal contextual reserve price defined in Equation (\ref{eq:opt_reserve}). 

% we point out that because $|E_{\ell+1}| \geq |E_{\ell}| $ for all $\ell \geq 1$ and for some $\tilde{\ell} \geq \log_2(\log_2(T))$ it holds true that $|E_{\tilde{\ell}}| = T^{1-2^{-\tilde{\ell}}} \geq T/2$, 

\textbf{The NPAC-S policy.} The detailed NPAC-S policy is shown in Algorithm \ref{algo:strat}, and consists of three main components.
\emph{(i) Phased Structure: } NPAC-S partitions $T$ into consecutive phases, where each phase $\ell \geq 1$, denoted as $E_{\ell}$, has length $T^{1-2^{-\ell}}$. This implies $|E_1| = \sqrt{T}$ and $|E_{\ell}|/\sqrt{|E_{\ell-1}|} = \sqrt{T}$. Here, we can establish that the total number of phases can be upper bounded by $\lceil\log_2(\log_2(T)) \rceil + 1$. \emph{(ii) Estimation for $\beta$, $F^{-}$ and $F^{+}$: } At the end of each phase, NPAC-S uses the submitted bids from the pervious phase and employs Ordinary Least Squares (OLS) and empirical distributions to estimate the mean vector $\beta$ as well as $F$, respectively. \emph{(iii) Random isolation: } NPAC-S incorporates random isolation periods in which a single buyer is chosen at random, and the item is auctioned to this isolated buyer (i.e. the seller only considers the bid of the isolated buyer and ignores bid from other buyers).\footnote{The seller discloses her commitment to the random isolation protocol to all buyers at $t = 0$, and it is not necessary for the seller to reveal, during an isolation period, which buyer is being isolated.} Note that  when a buyer $i$ is isolated, the buyer wins the item if and only if his bid is greater than the reserve price, and pays the reserve price if he wins. Here, the seller's pricing policy is announced to all buyers (at $t=0$) so that buyers examine the policy and have the freedom to adopt any bidding strategy to  maximize their long term discounted utility.
% Other buyers who are not isolated do not participate in the auction (i.e. their submitted bids are ignored). 

% More specifically,
% during each period in phase $\ell$, with probability $1/|E_{\ell}|$, the seller isolates one of the $N$ buyers uniformly at random, and sets reserve $r_t = r_t^u \sim \text{Uniform}(0,v_{\max})$, where $v_{\max}$ is the maximum possible buyer valuation. Note that 
%  when a buyer $i$ is isolated, the buyer wins the item if and only if his bid is greater than the reserve price, and pays the reserve price if he wins. Other buyers who are not isolated are not eligible to participate in the auction. Hence, one can think of a randomized isolation period in NPAC-S as running a second price auction with a single buyer.
% On the other hand, with probability $1- 1/|E_{\ell}|$, the seller sets the reserve price $r_t$ based on estimates of $\beta$, $F^-$, and $F^+$ calculated at the end of the previous phase $\ell-1$ using the data from $E_{\ell-1}$.  We point out that the seller's pricing policy is announced to all buyers so that buyers examine the policy and have the freedom to adopt any bidding strategy to  maximize their long term (discounted) utility.

\begin{savenotes}
 \begin{algorithm}[tbh]
    \centering
    \caption{Non-Parametric Contextual Policy against Strategic Buyers (NPAC-S)}	\label{algo:strat}
    \footnotesize
    \begin{algorithmic}[1]
  \State Initialize $\widehat{\beta}_1 = 0$,  and $\widehat{F}_1^-(z) =\widehat{F}_1^+(z) = 0$ for $\forall z \in \R$.
\For{phase $\ell \geq 1$}
      \For{$t\in E_{\ell}$}
      \State\textbf{Isolation}: With probability $1/|E_{\ell}|$, choose one buyer uniformly at random and offer price 
         \begin{align}
         \label{eq:strat:reserve_price_0}
             r_t^u \sim \text{Uniform}(0,v_{\max})\,.
         \end{align}
         \State\textbf{No Isolation}: With probability $1- 1/|E_{\ell}|$, set reserve price for all buyers as
         \begin{align}
         \label{eq:strat:reserve_price_1}
         \begin{aligned}
                 \widehat{r}_t =  \arg\max_{y\in [0, v_{\max}]}  
               & \int_{0}^y \widehat F^{-}_\ell(z- \langle \widehat\beta_\ell,x_t \rangle )dz - y\cdot \widehat F^{+}_\ell(y- \langle \widehat \beta_\ell,x_t \rangle)\,.
                 \end{aligned}
            \end{align}
         \State \textbf{Observe all bids $\{b_{i,t}\}_{i\in[N]}$}
        \EndFor
    \State \textbf{Update estimate of the mean vector $\beta$:} \footnote{For a matrix $A$, $A^\dag$ represents its pseudo inverse, so if $A$ is invertible,  we have $A^\dag = A^{-1}$. In Lemma \ref{lemma:strat:bias} of Appendix \ref{appsec:stratbuyers}, we show that with high probability $\sum_{\tau } x_\tau x_\tau^\top$ is positive definite, and hence invertible.}
    \begin{align} 
    \label{eq:strat:betaestimate}
            &  \widehat{\beta}_{\ell+1}  = ( \sum_{\tau \in E_{\ell}} x_\tau x_\tau^\top )^\dag \cdot ( \sum_{\tau \in E_{\ell}} x_\tau \Bar{b}_\tau )\,,
        \end{align}
        where $\bar b_{\tau}  = \frac{1}{N} \sum_{i\in [N]} b_{i, \tau}$.
    \State  \textbf{Update the estimate of $F^+$ and $F^-$:} 
            \begin{align}
            \begin{aligned}
              \label{eq:strat:F-F+estimate}
              &  \widehat{F}_{\ell+1}^-({z})  = N\widehat{F}_{\ell+1}^{N-1}({z}) - (N-1)\widehat{F}_{\ell+1}^N({z})~\text{ and }~ \widehat{F}_{\ell+1}^+({z}) = \widehat{F}_{\ell+1}^N({z}) 
              \,.
              \end{aligned}
        \end{align}
        where $\widehat{F}_{\ell+1}(z) $ is defined as
         \begin{align} 
         \label{eq:strat:Festimate}
                &  \widehat{F}_{\ell+1}(z)  =
              \frac{1}{N|E_{\ell}|} \sum_{\tau \in E_{\ell}}\sum_{i\in [N]} \I(b_{i,\tau} - \langle \widehat{\beta}_{\ell+1}, x_\tau \rangle \leq  {z}),    
              \end{align}
    \EndFor
    \end{algorithmic}
\end{algorithm}
\end{savenotes}

\begin{remark}
Here, we comment on how one can solve the reserve price  optimization problem in Equation (\ref{eq:strat:reserve_price_1}). The key observation is that for any period $t$, $\widehat{F}_{\ell}(\cdot)$ is a step function with jumps at points in the finite set $\mathcal{C}_{\ell}:= \{b_{i,\tau} - \langle \widehat{\beta}_{\ell} , x_{\tau}\rangle \}_{i\in [N], \tau\in E_{\ell-1}}$.  This implies that in order to solve for $r_t$ in Equation (\ref{eq:strat:reserve_price_1}), it suffices to conduct a grid search for $\forall {y} \in \mathcal{C}_{\ell}$. More specifically, we let $ \{z^{(0)}, z^{(1)}, \dots z^{(M)}\}$ be the ordered list (in increasing order) of all elements in $\mathcal{C}_{\ell} \cup \{0\}$, where $z^{(0)} := 0$ and $M := |\mathcal{C}_{\ell}|$ (here, we assumed that $0\notin \mathcal{C}_{\ell}$ without loss of generality). Hence, $r_t$ is equal to
\begin{align*}
     \arg\max_{m \in [M]} & ~
    \sum_{j=1}^m \widehat{F}^{-}_{\ell} (z^{(j)} - \langle \widehat{\beta}_{\ell},x_t \rangle )\cdot(z^{(j)} - z^{(j-1)})  - z^{(m)}  \widehat{F}^{+}_{\ell}(z^{(m)}- \langle \widehat{\beta}_{\ell},x_t \rangle)\,.
\end{align*}
This shows that the complexity to solve Equation (\ref{eq:strat:reserve_price_1}) is $\mathcal{O}(M^2)$. More detailed discussions and efficient algorithms regarding related problems can be found in \cite{mohri2016learning}.
\end{remark}

% Here, we comment on how one can solve the reserve price  optimization problem in Equation (\ref{eq:strat:reserve_price_1}). The key observation is that for any period $t$, $\widehat{F}_t(\cdot)$ is a step function with jumps at points in the finite set $\mathcal{C}_{t}:= \{b_{i,\tau} - \langle \widehat{\beta}_t , x_{\tau}\rangle \}_{i\in [N], \tau\in[t]}$. Furthermore, $\widehat{F}_t({z}) = 0$ for any ${z} < \min \mathcal{C}_{t}$ and $\widehat{F}_t({z}) = 1$ for any ${z}\geq  \max \mathcal{C}_{t}$. This implies that in order to solve for $r_t$ in Equation (\ref{eq:strat:reserve_price_1}), it suffices to conduct a grid search for $\forall {y} \in \mathcal{C}_{t}$. More specifically, we let $ \{z^{(0)}, z^{(1)}, \dots z^{(M)}\}$ be the ordered list (in increasing order) of all elements in $\mathcal{C}_{t} \cup \{0\}$, where $z^{(0)} := 0$ and $M := |\mathcal{C}_{t}|$ (here, we assumed that $0\notin \mathcal{C}_{t}$ without loss of generality). Hence,

\textbf{Motivation for design of NPAC-S.} Here
we provide some insights into the design of the NPAC-S policy, particularly the phased structure and the incorporation of random isolation periods.

Due to the phased structure of the algorithm, our estimates for $\beta$, $F^-$, and $F^+$ only depend on the bids and contextual features in the previous phase. Thus, corrupted bids submitted by buyers in past periods will have no impact on future estimates as well as pricing decisions. One can think of this as erasing  all memory prior to the previous phase and restarting the algorithm, which can potentially reduce buyers' manipulating power on our estimates and reserve prices.

We now discuss the impact of having isolation  periods. As all buyers are aware of the randomized isolation protocol, the presence of isolation periods restricts buyers from significantly corrupting their bids too often as by doing so they may suffer a substantial utility loss when they are isolated. To illustrate this point with an example, compare the following scenarios: (i) if there are no isolation periods, a buyer having the lowest valuation among all buyers may submit a bid by adding large corruption, but still ending up not being the second highest or highest bidder. Assuming that other buyers  bid truthfully, such a scenario will not lead to any changes in utility of any buyer, but introduces a large outlier to the set of data points used in our estimations. In words, when no isolation occurs, buyers may be able to distort the seller's learning process without facing unfavorable consequences; (ii) during an isolation period when a buyer is isolated, corrupting her bid may perhaps result in significant utility loss, e.g., losing the item by underbidding when her true valuation is greater than the reserve price, or winning the item by overbidding when her true valuation is less than the reserve price. Therefore, randomized isolation incentivizes utility-maximizing buyers to reduce the frequency of corrupting their bids. Mathematically, we characterize this statement in the following Lemma \ref{lemma:strat:boundedindividualbiglies}.

% As all buyers are aware of the randomized isolation protocol, the presence of isolation periods restricts buyers from significantly corrupting their bids too often as by doing they may suffer a substantial utility loss when they are isolated. To illustrate this point, consider the following: if no isolation occurs, a buyer may submit a bid that is far from her true valuation but face no consequences since her bid may not change the outcome nor payment of the auction. An example may be a buyer having the lowest valuation among all buyers, and submits a bid by adding a large corruption to her valuation, but still ending up not being the second highest or highest bidder. Assuming that other buyers  bid truthfully, such a scenario will not lead to any changes in utility of any buyer, but  introduces a large outlier to the set of data points used in our estimations. In words, when no isolation occurs, buyers may be able to distort the seller's revenue without losing anything. However, during an isolation period when a buyer is chosen, she is isolated from the influence of other buyers, and corrupting her bid may perhaps yield a significant utility loss, e.g., losing the item by underbidding when her true valuation is greater than the reserve price, or winning the item by overbidding when her true valuation is less than the reserve price. Therefore, randomized isolation incentivizes utility-maximizing buyers to reduce the frequency of corrupting their bids. Mathematically, we characterize this statement in
% the following Lemma \ref{lemma:strat:boundedindividualbiglies}.

\begin{lemma}[Bounding number of significantly corrupted bids]
\label{lemma:strat:boundedindividualbiglies}
For $i \in [N]$ and phase $\ell \geq 1$ define
\begin{align}
\label{eq:strat:defindividualbigcorruptions}
\begin{aligned}
    &  \mathcal{S}_{i,\ell} := \left\{t \in E_\ell: |a_{i,t}|\geq \frac{1}{|E_\ell|} \right\}~\text{ and }~  L_\ell := {\log\left({v_{\max}^2N|E_{\ell}|^4} -1 \right)}/{\log({1}/{\eta})}\,,
  \end{aligned}
\end{align}
where $\mathcal{S}_{i,\ell}$ is the set of all periods in phase $E_\ell$ during which buyer $i$ significantly corrupts his bids. Then, we have $\prob\left(\left| \mathcal{S}_{i,\ell}\right| > L_\ell \right) \leq {1}/{|E_{\ell}|} $.
\end{lemma}
The proof of this lemma is shown in Appendix \ref{appsec:strat:lem1}.

\textbf{Bounding the regret of NPAC-S.} 
Here, we first present the regret of NPAC-S. Then we introduce several key results that are crucial to proving the regret bound of NPAC-S and also comment on how they resolve challenges that arise due to buyers' strategic behavior.

\begin{theorem}[Regret of NPAC-S Policy]\label{thm:strat:Regret}
Suppose that the length of the horizon  $T \geq \max\{ \big(\frac{8x_{\max}^2}{\lambda_0^2}\big)^4, 9 \}$ where $\lambda_0^2$ is the minimum eigenvalue of covariance matrix $\Sigma$. Then, in the strategic setting,  the T-period  regret of the NPAC-S policy is in the order of ${\mathcal{O}\Big(c_f \sqrt{dN^3 \log(T) }\cdot \log\left(\log(T)\right)\big( \sqrt{T} + \frac{\sqrt{N^3 \log(T)}T^{\frac{1}{4}}}{\log\left(1/\eta\right)} \big)\Big)} $, where regret is computed against the benchmark policy in Definition \ref{def:benchmark_policy} that knows the mean vector $\beta$ and noise distribution $F$. Here, recall $c_f = \sup_{z \in [-\epsilon_{\max}, \epsilon_{\max}]} f(z)> 0$ where $f$ is the the pdf of $F$.
\end{theorem}
\begin{remark}
The proof of this theorem is presented in Appendix \ref{appsec:stratbuyers}. In the regret of NPAC-S, the factor $1/{\log(1/\eta)}$ serves as a worse case guarantee for the amount of corruption that buyers' can apply to their bids throughout the entire horizon $T$. As buyers get less patient, i.e., as  $\eta$ decreases,  buyers are less willing to forgo current utility in the current period. Thus, in the presence of randomized isolation periods, impatient buyers are less likely to significantly corrupt bids, which translates into lower regret. The $\log(\log(T))$ factor  corresponds to the information loss due to the policy's phased structure, which ``restarts'' the algorithm at the beginning of each of $\mathcal{O}\left(\log(\log(T))\right)$ phases and relies only on the information of the previous phase. 
\end{remark}

The regret of NPAC-S can be decomposed into two parts: (i) the estimation errors in $\beta$, $F^-$ and $F^+$, which result in the posted reserve price $r_t$  deviating from the optimal reserve price $r_t^\star$, and hence incur a revenue loss compared to the clairvoyant benchmark; and (ii) the revenue loss due to \emph{allocation mismatch} in the auction outcome because  of buyers' strategic bidding behaviour. Here, allocation mismatch refers to the phenomenon where a bidder would have won (lost) the auctioned item had she bid truthfully, but instead lost (won) the item as she submitted a corrupted bid in reality.

We first comment on several challenges with respect to bounding the estimation errors in $\beta$, $F^-$ and $F^+$. First, the OLS estimator and empirical distributions to estimate the mean vector and distributions $F^-$ and $F^+$, respectively are extremely vulnerable to corrupted data (outliers), and hence standard high probability bounds are  invalid for our setting. Additionally, there exists a complication in terms of bounding the estimation errors in $F^-$ and $F^+$ because estimation errors for $\beta$ will further propagate into the estimation errors in  $F$ and consequently impacting the estimates for $F^-$ and $F^+$. To illustrate this point, consider the ideal scenario where all bids are truthful (i.e. $v_{i,t} = b_{i,t}$ for all $i \in [N]$ and $t\in [T]$). Even in this scenario, the terms $v_{i,\tau} - \langle \widehat{\beta}_{\ell},x_\tau \rangle $ in the expressions for $\widehat{F}_{\ell}(\cdot) $ are not realizations of $\epsilon_{i,\tau}$ due to estimation errors in the mean vector $\widehat{\beta}_{\ell}$. Hence, the estimate $\widehat{F}_{\ell}(\cdot) $ evaluated at any point ${z} \in\R$ is biased, i.e. $\expect[\widehat{F}_{\ell}(z- \langle \widehat{\beta}_{\ell},x_t \rangle )] \neq F(z- \langle \widehat{\beta}_{\ell+1},x_t \rangle )$. Furthermore, the estimates $\widehat{F}_{\ell}^+(\cdot) $ and $\widehat{F}_{\ell}^-(\cdot)$ are evaluated at points which may be random variables since $\widehat{\beta_{\ell}}$  is a random variable that depends on the history of the previous phase.

%  we cannot naively apply CDF-based concentration inequalities (e.g. the Dvoretzky-Kiefer-Wolfowitz Inequality) to bound the distribution estimation errors, because 

% This also sheds light on the more subtle and challenging issue:  the estimates $\widehat{F}_{\ell}^+(\cdot) $ and $\widehat{F}_{\ell}^-(\cdot)$ are evaluated at points which may be random variables that since $\widehat{\beta_{\ell}}$  is a random variable that depends on the entire history up to the current period. 

In light of such challenges in bounding estimation errors, as one of our main contributions, the following Lemma \ref{lemma:shortenedestimatebounds} provides good estimation error guarantees for $\beta$, $F^-$ and $F^+$ in the presence of corrupted bids and the aforementioned error propagation phenomena. 
\begin{lemma}[Bounding estimation errors in $\beta$, $F^{-}$ and $F^{+}$]
\label{lemma:shortenedestimatebounds} 
For any phase $E_{\ell}$, with probability at least $1- \Theta\left(1/{|E_{\ell}|}\right)$, the following events hold: (i) $\norm{\widehat{\beta}_{\ell+1} - \beta}_1 = \mathcal{O} (\frac{1}{\sqrt{|E_\ell|}}+\frac{\log(|E_\ell|)}{\log(1/\eta){|E_\ell|}})$; (ii) for any ${z}\in \R$, $ |\widehat{F}_{\ell+1}^-({z}) - F^-({z})| = \mathcal{O} (\frac{N^2}{\sqrt{|E_\ell|}}+\frac{N^2\log(|E_\ell|)}{\log(1/\eta){|E_\ell|}} )$ and $|\widehat{F}_{\ell+1}^+({z}) - F^+({z})| = \mathcal{O} \big(\frac{N}{\sqrt{|E_\ell|}}+\frac{N\log(|E_\ell|)}{\log(1/\eta){|E_\ell|}} \big)$.
Here, recall the discount factor $\eta \in (0,1)$.
% \begin{align}
% \norm{\widehat{\beta}_{\ell+1} - \beta}_1 = \mathcal{O} \left(\frac{1}{\sqrt{|E_\ell|}}+\frac{\log(|E_\ell|)}{\log(1/\eta){|E_\ell|}} \right) \quad \left|\widehat{F}_{\ell+1}^-({z}) - F^-({z})\right| = \mathcal{O} \left(\frac{N^2}{\sqrt{|E_\ell|}}+\frac{N^2\log(|E_\ell|)}{\log(1/\eta){|E_\ell|}} \right) \text{ and} \left|\widehat{F}_{\ell+1}^+({z}) - F^+({z})\right| = \mathcal{O} \left(\frac{N}{\sqrt{|E_\ell|}}+\frac{N\log(|E_\ell|)}{\log(1/\eta){|E_\ell|}} \right)
% \end{align}
\end{lemma}
We refer readers to Lemma \ref{lemma:strat:bias}  and Lemma \ref{lemma:strat:boundF-F+estimate} in Appendix \ref{appsec:strat:otherlemmas}  for  more detailed statements on our high probability bounds regarding estimation errors in $\beta$, $F^{-}$ and $F^{+}$.

In addition to inaccurate estimates for $\beta$, $F^{-}$ and $F^{+}$, the allocation mismatch phenomenon due to strategic bidding also contributes to the regret of NPAC-S. For example, suppose that the highest valuation is  greater than the reserve price. In that case,  if buyers were truthful,  the item would be allocated and the seller would gain positive revenue. Now, if buyers shade their bids, the auctioned item may  not get allocated, resulting in zero revenue for the seller. In the following Lemma \ref{lemma:strat:boundedoutcomechanges}, we show that the number of allocation mismatch periods for each buyer is bounded with high probability. 
\begin{lemma}[Bounding allocation mismatch periods]\label{lemma:strat:boundedoutcomechanges} 
Define the following two sets of time periods:
\begin{align}
\begin{aligned}
\label{eq:strat:boundedoutcomechanges0}
    & \mathcal{B}_{i,\ell}^s = \left\{t\in E_\ell: v_{i,t} \geq D_{t}   ~ , ~ b_{i,t} \leq D_{t}   \right\} ~~~\text{and} \\
    & \mathcal{B}_{i,\ell}^o = \left\{t\in E_\ell: v_{i,t} \leq D_{t}  ~ , ~  b_{i,t} \geq D_{t}  \right\}  \\
    & \text{where } D_{t} = \max\{b_{-i,t}^+, \widehat{r}_t\}\,.
\end{aligned}
\end{align}
Here, $b_{-i,t}^+$ is the highest among all bids excluding that submitted by buyer $i$, and $\widehat{r}_t$ is the reserve price offered to all buyers if no isolation occurs (defined in Equation (\ref{eq:strat:reserve_price_1})).  Then, for $\mathcal{B}_{i,\ell} := \mathcal{B}_{i,\ell}^s \cup \mathcal{B}_{i,\ell}^o$, we have $ \prob\left(|\mathcal{B}_{i,\ell}| \leq 2L_\ell + 4c_f + 8\log(|E_\ell|) \right) ~ \geq ~ 1 -  {4}/{|E_\ell|}$, and $L_{\ell}$ is defined in Equation (\ref{eq:strat:defindividualbigcorruptions}).
% \begin{align*}
%       \prob\left(|\mathcal{B}_{i,\ell}| \leq 2L_\ell + 4c_f + 8\log(|E_\ell|) \right) ~ \geq ~ 1 -  \frac{4}{|E_\ell|} \,.
% \end{align*} 
Here, the probability is taken with respect to the randomness in $\{ (x_\tau, \epsilon_{i,\tau}, a_{i,\tau})\}_{\tau \in E_\ell, i\in [N]}$.
\end{lemma}
Note that $\mathcal{B}_{i,\ell}^s$ represents the set of all periods in phase $\ell$ during which buyer $i$ should have won the item if she bid truthfully, but in reality lost due to shading her bid (i.e. allocation mismatch due to shading), while similarly $\mathcal{B}_{i,\ell}^o$ is the periods of allocation mismatch due to overbidding. Therefore, $\mathcal{B}_{i,\ell} := \mathcal{B}_{i,\ell}^s \cup \mathcal{B}_{i,\ell}^o$ can be interpreted as the set of all periods in phase $\ell$ when an allocation mismatch occurs for buyer $i$. The detailed proof is provided in Appendix \ref{appsec:strat:lem2}.

\textbf{NPAC-S against Truthful Buyers. } 
% \label{sec:lazyNPACS}
Here, we make a remark that in a hypothetical world where buyers are truthful (i.e. $v_{i,t} = b_{i,t}$ or equivalently the degree of corruption $a_{i,t} = 0$ for all $i \in [N]$, $t\in[T]$), our proposed NPAC-S policy achieves a regret of $\mathcal{O}(c_f \sqrt{dN^3T\log(T) }\cdot \log\log(T) )$ compared to the clairvoyant benchmark policy in Definition \ref{def:benchmark_policy}. Intuitively, this is easy to see because the set of all periods in phase $E_\ell$ during which a buyer $i$ significantly corrupts his bids, namely $\mathcal{S}_{i,\ell}$ defined in Lemma \ref{lemma:strat:boundedindividualbiglies}, will be empty. As a result, there will be no allocation mismatch periods, and the $1/\log(1/\eta)$ terms in the estimation errors in $\beta$, $F^{-}$, $F^{+}$ will vanish (see Lemma \ref{lemma:shortenedestimatebounds}). The proof for the regret bounds of NPAC-S against truthful buyers would thus be a simplification to the proof of Theorem \ref{thm:strat:Regret}, and hence will be omitted.

\section{Numerical Study}
Here, we present numerical simulations to compare the performance of NPAC-S with several baseline seller policies. In particular, consider the following baseline policies: (i) \textsc{Naive} which always sets a 0 reserve price; (ii) \textsc{ContHEDGE} which runs an independent version of the \textsc{HEDGE} algorithm for every distinct context vector (see an introduction of \textsc{HEDGE} for the adversarial multi-arm bandit problem in \cite{auer1995gambling}). The ``arms'' of \textsc{HEDGE}  correspond to potential reserve price options. Note that \textsc{HEDGE} is a special case of the well-known EXP3 algorithm which is a simple off-the-shelf algorithm that not only has good theoretical guarantees, but has also been applied (or its variations/generalizations have been adopted) in many areas in online advertising (see e.g. \cite{zimmert2019optimal,balseiro2019learning,han2020learning}). (iii) \textsc{HO-SERP}, which sets personalized reserve prices for each buyer using ``rolling window'' estimates of $\beta$ and $F$ w.r.t other buyers' submitted bids (see \cite{kanoria2021incentive}). Here we consider \textsc{HO-SERP} as a baseline because among all seller algorithms in related works that study pricing in a contextual, stochastic, and strategic buyer setting similar to ours (see Table \ref{table}), \textsc{HO-SERP} achieves nearly the best theoretical performance. Note  \textsc{HO-SERP} requires the noise distribution to be MHR.

To model buyers' strategic behavior, instead of restricting buyers to bid according to a specific strategy to maximizes long-term discounted utility, we instead mimic the outcome of some general class of such strategies (parameterized by $\eta$) via randomly selecting periods over the entire horizon and have buyers significantly corrupt bids in these periods.  We will refer to these randomly selected periods as \emph{corruption periods}. When this randomization procedure is repeated over many trials, we believe the average bidding outcome would serve as a relatively accurate approximation to the outcomes of a general class of strategies for utility-discounting buyers. Furthermore, inspired by Lemma \ref{lemma:strat:boundedindividualbiglies} which suggests that the number of periods when a buyer significantly corrupts her bid is bounded, we let the selected number of corruption periods be $L_{\ell}$ defined in Equation (\ref{eq:strat:defindividualbigcorruptions}) . Note that $L_{\ell}$ is increasing in $\eta$ and represents the fact that more patient buyers (i.e. larger $\eta$) value long term utility more and hence would be willing to corrupt bids more frequently with the aim of achieving higher future utility.

Out detailed experimental setup is as followed. We consider a horizon of length $T = 5,000$, $N = 2$ buyers, context vectors of dimension $d =4$, $v_{\max} = 10$ and $v_{\min} = 0$.  For each $\eta \in \{0.2,0.4,0.6,0.8\}$, repeat the following procedure for $n = 50$ trials, each including $T$ periods:

For each phase $E_{\ell}$ ($\ell \geq 1$), \footnote{For fixed $T$, since length of phase $\ell \geq 1$ is $T^{1-2^{-\ell}}$, in our case when $T = 5,000$ we have 4 phases whose phase lengths are $70, 594, 1724, 2612$, respectively, where the last phase is truncated.} sample $L_{\ell}$ corruption periods uniformly at random. Then, regarding buyer's valuations, we generate $\beta \in [0,1]^{d}$, where each entry is sampled independently according to a uniform distribution on [0,1], i.e., $U(0,1)$. We further normalize $\beta$ with the sum of all entries so that $\norm{\beta}_{2} = 1$. We then generate 10 distinct contexts vectors $\mathcal{\contex}= \{\contex^{j}\}_{j \in [10]}$, where each entry for any distinct context vector is sampled independently from $U\left(\frac{v_{\max}}{3},\frac{2v_{\max}}{3}\right)$. Then, for every period $t\in[T]$, sample $x_{t}$ uniformly at random from $\mathcal{\contex}$, and sample $\epsilon_{i,t}$ for all $i \in [N]$ independently from $U\left(-\frac{v_{\max}}{3},\frac{v_{\max}}{3}\right)$. Note that our construction guarantees $v_{i,t} = \langle\beta,x_{t} \rangle + \epsilon_{i,t} \in [v_{\min},v_{\max}]$, and the noise distribution is uniform which satisfies the MHR assumption (so the application of the \textsc{HO-SERP} is valid). If $t$ is a corruption period, we let buyers submit a bid of value 0 to model the behavior of significant bid-shading; otherwise, we let buyers bid their true valuations.\footnote{We remark that our numerical experiments focus on buyers' bid-shading behavior. This is mainly because empirical studies found that shading is prevalent in repeated auctions on modern online advertising platforms and theoretical works have demonstrated various versions of bid-shading strategies can help buyers achieve near-optimal performances in a variety of practical settings, such as buyers being constrained by a limited budget or target return on investment (see e.g. \cite{zeithammer2007research,golrezaei2018auction,balseiro2019learning}). }

\begin{figure}[h]
\includegraphics[width=0.7\textwidth]{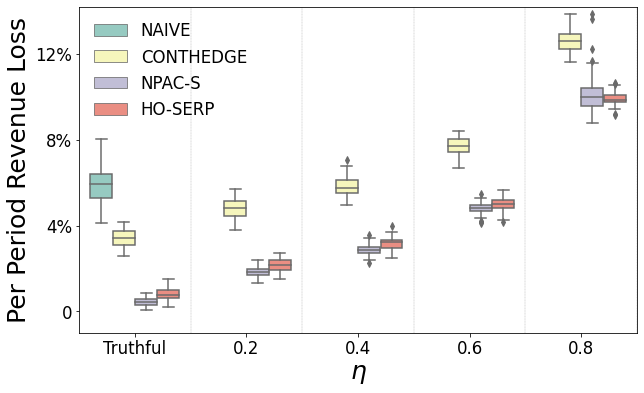}
\caption{\footnotesize \textbf{Performance comparison with baselines.} This figure displays the average per-period revenue loss compared to the benchmark policy (Definition \ref{def:benchmark_policy}). Each box plot corresponds to $n = 50$ trials. \textsc{Naive} is only run for the truthful setting because buyers will have no incentive to bid untruthfully when there is no reserve price. \textsc{ContHedge} is run with``arms'' $\{0, 0.5,1, \dots 10\}$, where each arm corresponds to a reserve price option.}
\label{fig:compare}
\end{figure}

 For comprehensiveness, we also consider the truthful setting by repeating the above valuation generation procedure for another $n =50$ trials and have buyers always submit their true valuations. Finally, for each of the aforementioned trials, we run the NPAC-S as well as other baseline algorithms independently and simply record the realized revenue of each algorithm across all repeated auctions. We report the average per-period revenue loss compared to the benchmark policy (Definition \ref{def:benchmark_policy}) for each algorithm in Figure \ref{fig:compare}. 

% \begin{figure}[!htb]
% \floatbox[{\capbeside\thisfloatsetup{capbesideposition={right,top},capbesidewidth=0.4\textwidth}}]{figure}[\FBwidth]
% {\caption{\footnotesize \textbf{Performance comparison with baselines.} This figure displays the average per-period revenue loss compared to the benchmark policy (Definition \ref{def:benchmark_policy}). Each box plot corresponds to $n = 50$ trials. \textsc{Naive} is only run for the truthful setting because buyers will have no incentive to bid untruthfully when there is no reserve price. \textsc{ContHedge} is run with``arms'' $\{0, 0.5,1, \dots 10\}$, where each arm corresponds to a reserve price option.} \label{fig:compare}}
% {\includegraphics[width=0.5\textwidth]{Incentive}}
% \end{figure}

We observe that our proposed NPAC-S algorithm not only outperforms \textsc{ContHedge} in all settings consistently by a $3\%\sim 4\%$ and \textsc{NAIVE} in the truthful setting by $6\%\sim 7\%$, NPAC-S also generally yields  more stable outcomes as measured by the standard deviation of per-period revenue loss across $n$ trials. Compared to \textsc{HO-SERP}, NPAC-S slightly outperforms \textsc{HO-SERP} in the truthful setting and for $\eta =$ 0.2, 0.4, 0.6. Nevertheless, we point out that our experimental setting inherently favors \textsc{HO-SERP} since performance guarantees of this algorithm relies on the noise distribution being MHR, which is the case for our uniform noise. Moreover, the comparison with \textsc{HO-SERP} also demonstrates the advantages of NPAC-S from a practical viewpoint, since NPAC-S, unlike \textsc{HO-SERP},  sets a single reserve price for all buyers and still matches or improves upon the performance of \textsc{HO-SERP}.

\newpage
\bibliographystyle{informs2014}
\bibliography{ref}

\newpage
\begin{APPENDICES}

\begin{center}
% \noindent\makebox[\linewidth]{\rule{0.8\paperwidth}{0.9pt}}
\vspace{0.8cm}
    \Large Appendices for\\
    \vspace{0.2cm}
    \Large \textbf{Incentive-aware Contextual Pricing with Non-parametric Market Noise}
    \noindent\makebox[\linewidth]{\rule{1\linewidth}{0.9pt}}
\end{center}
\setcounter{page}{1}

% Our Appendix is organized as followed. In Appendix \ref{app:extendreview}, we include an extended literature review that discusses broader related works. In Appendix \ref{add:material}, we include additional material to the main body of the paper (e.g. the NPAC-T policy discussed in Section \ref{sec:lazyNPACS}). Appendix \ref{app:Sec:benchmark} includes the all proofs of the results in Section \ref{Sec:benchmark}. Both Appendices \ref{thm:proof:bound:fullInfoRegret} and \ref{app:Sec:strat} are dedicated to Section \ref{Sec:strat}. In particular, Appendix \ref{thm:proof:bound:fullInfoRegret} proves Theorem \ref{bound:fullInfoRegret}
% which shows a regret bound for the NPAC-T policy against truthful buyers; while Appendix \ref{app:Sec:strat} proves Theorem \ref{thm:strat:Regret}
% which shows a regret bound for the NPAC-S policy against strategic buyers. We put these two appendices in this order because the proof for Theorem \ref{bound:fullInfoRegret} for truthful buyers is simpler, and the proof of Theorem \ref{thm:strat:Regret} for strategic buyers would build on some parts of the proof 
% of Theorem \ref{bound:fullInfoRegret}.

Our Appendix is organized as followed. In Appendix \ref{app:extendreview}, we include an extended literature review that discusses broader related works. Appendix \ref{app:Sec:benchmark} includes the all proofs of the results in Section \ref{Sec:benchmark}.  Appendix  \ref{app:Sec:strat} is dedicated to Section \ref{Sec:strat}. In particular, Appendix  \ref{app:Sec:strat} proves Theorem \ref{thm:strat:Regret}
which shows a regret bound for the NPAC-S policy against strategic buyers. 

\section{Extended Literature Review}
\label{app:extendreview}
There has been a large body of literature that considers the problem of non-contextual dynamic pricing with non-strategic buyers.
\cite{kleinberg2003value} studies repeated non-contextual posted price auctions with a single buyer whose valuations are fixed, drawn from a fixed but unknown distribution, and chosen by an adversary who is oblivious to the seller's algorithm. \cite{den2013simultaneously, besbes2009dynamic,broder2012dynamic} study non-contextual dynamic pricing with demand uncertainty, where they estimate unknown model parameters using estimation techniques such as maximum likelihood. \cite{golrezaei2021bidding} considers a seller repeatedly pricing against a buyer who is subject to budget and return-on-investment (ROI) constraints. 
\cite{cesa2015regret} considers the dynamic pricing problem in non-contextual repeated second-price auctions with multiple buyers whose bids are drawn from some unknown and possibly non-parametric distribution. In addition, they also consider bandit feedback where the seller only observes realized revenues instead of all submitted bids. In their non-contextual setup, the seller's revenue-maximizing price is fixed throughout the entire time horizon, and the key is to approximate this optimal price by estimating the valuation distribution. In our setting, however, the optimal reserve prices are context-dependent, which means the seller is required to estimate (i) the distributional form of valuations and (ii)  buyers' willingness-to-pay that varies in each period according to different contexts.

Another line of research studies the problem of contextual dynamic pricing with non-strategic buyer behavior. \cite{cohen2016feature, lobel2018multidimensional,leme2018contextual} propose learning algorithms based on binary search methods when the context vector is chosen adversarially in each round.  
\cite{chen2018nonparametric} consider the problem where a learner observes contextual features and optimizes an objective by experimenting with a fixed set of decisions. %They present a tree-based non-parametric learning policy that adaptively splits the feature space into smaller bins (hyper-rectangles), and eventually learns near-optimal decisions in each bin. 
Their tree-based non-parametric learning policy is designed to handle very general objectives and not specifically tailored to pricing problems. Thus, in pricing problems, its performance deteriorates as the dimension of the feature vector increases. \cite{javanmard2016dynamic} also considered a contextual pricing problem with an unknown but parametric noise distribution, and uses a maximum likelihood estimator to jointly estimate the mean vector and distributions parameters. \cite{shah2019semi} studied a dynamic pricing problem in repeated posted price mechanisms. They considered a model where the relationship between the expectation of the logarithm of buyer valuation and the contextual features is linear, while the market noise distribution is non-parametric. This logarithmic form of the valuation model allows them to separate the noise term from the context, which makes it possible to independently estimate the noise distribution and expected buyer valuation.  In our setting, however, the context is embedded within the noise distribution, and our estimation errors in the mean vector $\beta$ will propagate into the estimation error in the noise distribution, making the learning task more difficult, compared to that in  \cite{shah2019semi}. 

Finally, our work is also related to  the recent literature within the domain of mechanism design and online learning that adopt methodologies from differential privacy to deal with strategic agents; see, for example,   \cite{mcsherry2007mechanism, mahdian2017incentive, liu2018learning}.

% \section{Additional Material}
% \label{add:material}
% \input{Add_material}

\section{Appendix for Section \ref{Sec:benchmark}: Proof of Proposition \ref{prop:benchmark}}
\label{app:Sec:benchmark}
\label{appsec:prop}
Let $Q_t(\cdot)$ be the distributions of a buyer's valuation when we condition on the feature vector $x_t$.  Further, let $Q_t^-(\cdot)$ be the distribution of  $v_t^-$, which is the second highest valuation at time $t$. Then, we have $Q_t(z) = F(z - \langle \beta, x_t\rangle)$ and  $Q_t^-(z) = F^-(z - \langle \beta, x_t\rangle)$. When $N\geq 2$ and all buyers bid truthfully, according to Equations (\ref{eq:revenue}) , the seller's expected revenue conditioned on $x_t$ by setting reserve price $r_t$ is: 
    \begin{align}
    \begin{aligned}
    \label{eq:prop:1}
        \rev_t(r_t)~ = ~ & \expect\left[\max\{r_t, v_t^-\} \I\{v_t^+ \geq r_t \} ~ | ~ x_t, r_t\right] \\
        ~ = ~ & \expect \left[ r_t \I\{v_t^+ \geq  r_t \geq v_t^-\}  +  v_t^- \I\{v_t^+ \geq v_t^- \geq  r_t\} ~ | ~ x_t, r_t \right]\,,
        \end{aligned}
    \end{align} 
     where  $v_t^+$ is the highest valuation at time $t$. 
    The first term within the expectation, conditioned on $x_t$ and $r_t$, is %\negin{It is not clear how you got the next expression. Please add an explanation. } 
    \begin{align}
    \label{eq:prop:2}
         \expect \left[ r_t \I\{v_t^+ \geq  r_t \geq v_t^-\}   ~ | ~ x_t, r_t  \right] = r_t N \left[Q_t(r_t) \right]^{N-1}[1-Q_t(r_t)]\,,
    \end{align}
    {where we used the fact that $r_t$ is independent of $v_t^+$ and $v_t^-$ since the seller sets reserve price $r_t$ based on only the past history $\mathcal{H}_{t-1}=\{(r_1, v_1,x_1), (r_2, v_2,x_2), \ldots, (r_{t-1}, v_{t-1},x_{t-1})\}$, and both $v_t^+$ and $v_t^-$, conditioned on $x_t$, are independent of the past.} The second term within the expectation of Equation (\ref{eq:prop:1}) is 
    \begin{align}
    \label{eq:prop:3}
         \expect \left[ v_t^- \I\{v_t^+ \geq v_t^- \geq  r_t\}   ~ | ~ x_t, r_t  \right] ~ = ~ &   \expect \left[ v_t^- \I\{ v_t^- \geq  r_t\}   ~ | ~ x_t, r_t \right]\nonumber \\
         ~ = ~ &  \expect \left[ (v_t^- -r_t)\I\{ v_t^-  \geq  r_t\}   ~ | ~ x_t, r_t  \right] + r_t \expect \left[ \I\{ v_t^- \geq  r_t\}   ~ | ~ x_t, r_t  \right]\nonumber \\
         ~ = ~ &   \int_{0}^\infty \prob\left( v_t^- - r_t \geq z\right) dz + r_t\left[1 - Q_t^-(r_t) \right]\nonumber \\
          ~ = ~ &   \int_{r_t}^{\infty} \left[ 1 - Q_t^-(z)\right] dz + r_t\left[1 - Q_t^-(r_t) \right] \nonumber \\
          ~ = ~ &   \expect\left[ v_t^- ~ | ~ x_t, r_t \right] - \int_{0}^{r_t} \left[ 1 - Q_t^-(z)\right] dz + r_t\left[1 - Q_t^-(r_t) \right] \nonumber \\
          ~ = ~ &  \expect\left[ v_t^- ~ | ~ x_t \right] + \int_{0}^{r_t}  Q_t^-(z) dz - r_t Q_t^-(r_t)\,.
    \end{align} 
    {Note that the integration starts from $0$ because all valuations are considered to be positive.} Since $F^-(\Tilde{z}) := NF^{N-1}(\Tilde{z}) - (N-1)F^N(\Tilde{z})$ for any $\Tilde{z}\in\R$, we have
    \begin{align}
    \label{eq:prop:4}
        Q_t^-(r_t) ~ = ~ N \left[Q_t(r_t) \right]^{N-1}[1-Q_t(r_t)]  +  \left[Q_t(r_t) \right]^N\,.
    \end{align}
 Hence, combining Equations (\ref{eq:prop:1}), (\ref{eq:prop:2}), (\ref{eq:prop:3}), and (\ref{eq:prop:4}), we have
    \begin{align*}
        \rev_t(r_t) ~ = ~ & \expect\left[ v_t^- ~ | ~ x_t\right] + \int_{0}^{r_t}  Q_t^-(z) dz -  r_t\left[Q_t(r_t) \right]^N\\
        ~ = ~ &   \expect\left[ v_t^- ~ | ~ x_t \right] + \int_{0}^{r_t}  F^-({z} - \langle \beta, x_t\rangle ) d{z} -  r_t \left[F^+(r_t- \langle \beta, x_t\rangle) \right]\\
        ~ = ~ &  \int_{-\infty}^{\infty} z dF^-(z) + \langle \beta,x_t \rangle  +  \int_{0}^{r_t}  F^-({z} - \langle \beta, x_t\rangle ) d{z} -  r_t \left[F^+(r_t- \langle \beta, x_t\rangle) \right]\,.
    \end{align*}  
    % where the second equality is due to the definition of $G(\cdot)$.
    % Finally, noting $v_t^- =\langle \beta, x_t\rangle + \epsilon_t^- $ and using the fact that $\{ \epsilon_{it} \}_{t\in[T], i\in [N]}$ are i.i.d in $i$ and $t$ implies  $\{ \epsilon_t^- \}_{t\in[T]}$ are i.i.d yields concludes the proof.

% \section{Appendix for Section \ref{Sec:strat} (Truthful Buyers): Proof of Theorem \ref{bound:fullInfoRegret}}
% \label{thm:proof:bound:fullInfoRegret}
% \input{A2_AppendixFullInfo.tex}

% \section{Appendix for Section \ref{Sec:bandit}: Proof of Theorem \ref{banditbound:regret}}
% \newpage
% \section{Appendix for Section \ref{Sec:anombuyer}: Proof of Theorem \ref{thm:anombuyer:Regret}}
% \input{A3_AppendixAnomalousBuyers.tex}

\section{Appendix for Section \ref{Sec:strat}: Proof of Theorem \ref{thm:strat:Regret}}
\label{app:Sec:strat}
\label{appsec:stratbuyers}

We first introduce some definitions that we will extensively rely on throughout our proof of Theorem \ref{thm:strat:Regret}. We start off with the ``good'' events $\xi_{\ell+1}$, $\xi_{\ell+1}^-$ and $\xi_{\ell+1}^+ $ for $\ell \geq 1$ in which the estimates of $\beta$, $F^-$ and $F^+$ are accurate:
\begin{align}
    & \xi_{\ell+1} = \left\{\norm{\widehat{\beta}_{\ell+1} - \beta}_1 \leq \frac{\delta_{\ell}}{x_{\max}} \right\} 
    \label{eq:strat:defxi}\\
    & ~~~~~~~ \text{where}~~  \delta_{\ell}:= \frac{\sqrt{2d\log(|E_\ell|)}\epsilon_{\max} x_{\max}^2}{\lambda_0^2\sqrt{N|E_\ell|}} + \frac{\sqrt{d}\left(NL_\ell a_{\max} + 1 \right)x_{\max}^2}{|E_\ell|\lambda_0^2}
    \label{eq:strat:defdelta} \,,\\
     & \xi_{\ell+1}^- = \left\{\left|\widehat{F}_{\ell+1}^-({z}) - F^-({z})\right| ~ \leq ~ 2N^2\left(\gamma_{\ell} + c_f \delta_{\ell } + \frac{c_f + NL_\ell}{|E_\ell|} \right) \right\} \label{eq:strat:defxi-}\,, \\ 
     &\xi_{\ell+1}^+ = \left\{\left|\widehat{F}_{\ell+1}^+({z}) - F^+({z})\right| ~ \leq ~ N\left(\gamma_{\ell} + c_f \delta_{\ell } + \frac{c_f + NL_\ell}{|E_\ell|} \right) \right\} \label{eq:strat:defxi+}\,,
\end{align}
where $a_{\max}$ is the maximum possible corruption, $\gamma_{\ell} = {\sqrt{\log(|E_\ell|)}}/{\sqrt{2N|E_\ell|}}$, $\lambda_0^2$ is the minimum eigenvalue of covariance matrix $\Sigma$, and $c_f  = \sup_{z \in [-\epsilon_{\max}, \epsilon_{\max}]} f(z) \geq \inf_{z \in [-\epsilon_{\max}, \epsilon_{\max}]} f(z) > 0$. Furthermore, 
$$L_\ell = \frac{\log\left({v_{\max}^2N|E_{\ell}|^4} -1 \right)}{\log({1}/{\eta})} = \mathcal{O}\left(\frac{\log(|E_\ell|)}{\log({1}/{\eta})}\right),$$ where $|E_{\ell}| = T^{1-2^{-\ell}}$ is the length of the $\ell^{th}$ phase.

We also define the event that the number of periods in phase $E_\ell$ during which buyer $i$ submits significantly corrupted bids is bounded by $L_{\ell}$:
\begin{align}
    \mathcal{G}_{i,\ell}:= \left\{\left| \mathcal{S}_{i,\ell}\right| \leq  L_\ell \right\}\,.
\end{align}
Here,  $   \mathcal{S}_{i,\ell} = \left\{t \in E_\ell: |a_{i,t}|\geq \frac{1}{|E_\ell|} \right\}
$  is the set of all periods in phase $E_\ell$ during which buyer $i$ extensively corrupts her bids.

% \begin{align}
%     & \mathcal{S}_\ell = \cup_{i\in[N]} \mathcal{S}_{i,\ell} ~~\text{and}~~  \mathcal{S}_\ell^c = \{t \in E_\ell:  \}
% \end{align}
\bigskip
We are now equipped to show Theorem \ref{thm:strat:Regret} according to the following steps:
\begin{itemize}
    \item [(i)] Decompose the single period regret into $\mathcal{R}_t^{(1)}$ and $\mathcal{R}_t^{(2)}$, where  $\mathcal{R}_t^{(1)}$  bounds the expected revenue loss due to the discrepancy between the actual reserve price $r_t$ and the optimal reserve price $r_t^\star$ and  $\mathcal{R}_t^{(2)}$, which bounds the expected revenue loss due to allocation mismatches.
    Note that  $\mathcal{R}_t^{(1)}$  is a result of the estimation inaccuracies in $\beta$, $F^-$ and $F^+$. 
    
    \item [(ii)] Bound $\mathcal{R}_t^{(1)}$ using Lemmas \ref{lemma:strat:boundedindividualbiglies}, \ref{lemma:strat:bias}, \ref{lemma:strat:boundF-F+estimate}, and \ref{lemma:strat:controlUncert}.
    \item [(iii)]  Bound $\mathcal{R}_t^{(2)}$ using Lemmas \ref{lemma:strat:boundedindividualbiglies} and \ref{lemma:strat:boundedoutcomechanges}.
    \item [(iv)] Sum up $\mathcal{R}_t^{(1)}$ and $\mathcal{R}_t^{(2)}$ to bound the cumulative expected regret over a phase $E_\ell$ and the entire horizon $T$.
\end{itemize}

\bigskip
\textbf{(i) Decomposing single period regret into $\mathcal{R}_t^{(1)}$ and $\mathcal{R}_t^{(2)}$:}
According to the NPAC-S policy detailed in Algorithm \ref{algo:strat}, the expected revenue in period $t$ is given by
\begin{align}
\label{eq:strat:revenue}
    \rev_t(r_t) 
   ~ = ~ & \expect\left[\max\{b_t^-,\widehat{r}_t \}\I\{b_t^+ > \widehat{r}_t\} \I\{\text{no isolation in }t\} + \sum_{i\in[N]} r_t^u\I\{b_{i,t} > r_t^u\}\I\{\text{$i$ is isolated}\} ~ \left.\right| ~ x_t, r_t \right] 
    \,,
\end{align}
where the expectation is taken with respect to $\{ (x_\tau, \epsilon_{i,\tau}, a_{i,\tau})\}_{\tau \in [t], i\in [N]}$ and $\widehat{r}_t,r_t^u$ are defined in Equations (\ref{eq:strat:reserve_price_0}) and (\ref{eq:strat:reserve_price_1}) respectively. Hence, the regret is given by 
\begin{align}
\label{eq:strat:boundR_t}
    \reg_t 
     ~ = ~ & \expect\left[\text{REV}^\star_t - \rev_t(r_t) \right]\nonumber \\
    ~ = ~ & \expect\left[\max\{v_t^-,r_t^\star \}\I\{v_t^+ > r_t^\star\} - \rev_t(r_t) \right]  \nonumber\\
    ~ = ~ & \left(\expect\left[\max\{v_t^-,r_t^\star \}\I\{v_t^+ > r_t^\star\}\right] -  \expect\left[\max\{v_t^-, \widehat{r}_t \}\I\{v_t^+ >  \widehat{r}_t\}\I\{\text{no isolation in }t\}\right]\right) \nonumber\\ 
    & ~~ + \left(\expect\left[\max\{v_t^-, \widehat{r}_t \}\I\{v_t^+ >  \widehat{r}_t\}\I\{\text{no isolation in }t\} - \rev_t(r_t)\right]
     \right)\nonumber\\
    ~ := ~ & \mathcal{R}_t^{(1)} + \mathcal{R}_t^{(2)} \,,
\end{align}
where the expectation is taken with respect the context $x_t \sim \mathcal{D}$ and the randomness in $r_t$; $r_t^\star$ is the optimal reserve price (defined in Equation (\ref{eq:opt_reserve})) if the seller has full knowledge of $F$ and $\beta$; and we defined:
\begin{align}
    & \mathcal{R}_t^{(1)}:= \expect\left[\max\{v_t^-,r_t^\star \}\I\{v_t^+ > r_t^\star\}\right] -  \expect\left[\max\{v_t^-, \widehat{r}_t \}\I\{v_t^+ >  \widehat{r}_t\}\I\{\text{no isolation in }t\}\right] \nonumber \\
    & \mathcal{R}_t^{(2)} := \expect\left[\max\{v_t^-, \widehat{r}_t \}\I\{v_t^+ >  \widehat{r}_t\}\I\{\text{no isolation in }t\} - \rev_t(r_t) \right]
\end{align}

\textbf{(ii) Bounding $\mathcal{R}_t^{(1)}$:}
We start by upper bounding $\mathcal{R}_t^{(1)}$ for a period $t\in E_{\ell +1}$ where $\ell \geq 1$.
% First, according to Algorithm \ref{algo:strat}, when no isolation occurs the seller sets reserve price $\widehat{r}_t$ defined in Equation (\ref{eq:strat:reserve_price_1}), so
% \begin{align}
% \label{eq:strat:boundR_t^11}
%     \expect\left[\max\{v_t^-,r_t \}\I\{v_t^+ > r_t\}\right] ~ \geq ~ & \expect\left[\max\{v_t^-,r_t \}\I\{v_t^+ > r_t\}\I\{\text{no isolation in }t\}\right] \nonumber\\
%     ~ = ~ & \expect\left[\max\{v_t^-, \widehat{r}_t \}\I\{v_t^+ >  \widehat{r}_t\}\I\{\text{no isolation in }t\}\right] \,.
% \end{align}
% Hence, 
\begin{align}
\label{eq:strat:boundR_t^11.5}
    \mathcal{R}_t^{(1)} ~ = ~ & \expect\left[\max\{v_t^-,r_t^\star \}\I\{v_t^+ > r_t^\star\}\right] -  \expect\left[\max\{v_t^-, \widehat{r}_t \}\I\{v_t^+ >  \widehat{r}_t\}\I\{\text{no isolation in }t\}\right] \nonumber \\
    ~ = ~ & \expect\left[\left(\max\{v_t^-,r_t^\star \}\I\{v_t^+ > r_t^\star\} - \max\{v_t^-, \widehat{r}_t \}\I\{v_t^+ >  \widehat{r}_t\}\right)\I\{\text{no isolation in }t\}\right] \nonumber\\
     & ~~ + \expect\left[\max\{v_t^-,r_t^\star \}\I\{v_t^+ > r_t^\star\}\left(1-\I\{\text{no isolation in }t\} \right)\right]  \nonumber \\
      ~ = ~ & \expect\left[\max\{v_t^-,r_t^\star \}\I\{v_t^+ > r_t^\star\} - \max\{v_t^-, \widehat{r}_t \}\I\{v_t^+ >  \widehat{r}_t\}\right]\left(1 - \frac{1}{|E_\ell|} \right) \nonumber\\
     & ~~ + \expect\left[\max\{v_t^-,r_t^\star \}\I\{v_t^+ > r_t^\star\} \right]\cdot \frac{1}{|E_\ell|}  \nonumber \\
     ~ \leq ~ & \expect\left[\max\{v_t^-,r_t^\star \}\I\{v_t^+ > r_t^\star\} - \max\{v_t^-, \widehat{r}_t \}\I\{v_t^+ >  \widehat{r}_t\}\right]  + \frac{v_{\max}}{|E_\ell|}
    %   ~ := ~ & \widetilde{\mathcal{R}}_t^{(1)} + \frac{v_{\max}}{|E_\ell|}
     \,,
\end{align}
where the third equality is because an isolation event is  independent of  any other event, and the final inequality follows from a simple observation that $\max\{v_t^-,r_t^\star \}\I\{v_t^+ > r_t^\star\} \leq v_{\max}$.
% Here, we introduce some additional notation. Let $\expect_{\epsilon_t}[\cdot]$ be the expectation taken with respect to $\{\epsilon_{i,t}\}_{ i\in [N]}$. Similar to the proof of Theorem \ref{bound:fullInfoRegret}, we define $y_t := \langle \beta, x_t \rangle $, $\widehat{y}_t := \langle \widehat{\beta}_\ell, x_t \rangle $. Recall the definition of $\rho_t({r},y,F^{(1)},F^{(2)}) := \int_{0}^{r} F^{(2)}({z}-y)d{z} - r \left[ F^{(1)}({r}-y)\right]$ in Equation (\ref{eq:defrho}).
% Note that $\expect_{\epsilon_t}\left[\max\{v_t^-, r \}\I\{v_t^+ > r\}\right] = \rho_t(r, y_t, F^-, F^+)$ for any $r\in\R$ that does not depend on $\left\{\epsilon_{i,t}\right\}_{i\in [N]}$. Hence, \jason{since $r_t^\star$ only depends on $x_t$ and $\widehat{r}_t$ depends on $x_t$ and $\{(r_\tau, b_\tau,x_\tau)\}_{\tau\in E_{\ell}}$  } $\expect_{\epsilon_t}\left[\max\{v_t^-,r_t^\star \}\I\{v_t^+ > r_t^\star\}\right] = \rho_t(r_t^\star, y_t, F^-, F^+)$ and $\expect_{\epsilon_t}\left[\max\{v_t^-, \widehat{r}_t \}\I\{v_t^+ > \widehat{r}_t\}\right] = \rho_t(\widehat{r}_t, y_t, F^-, F^+)$ according to Proposition \ref{prop:benchmark}.  

For simplicity, we define
\begin{align*}
    \widetilde{\mathcal{R}}_t^{(1)} := \expect\left[\max\{v_t^-,r_t^\star \}\I\{v_t^+ > r_t^\star\} - \max\{v_t^-, \widehat{r}_t \}\I\{v_t^+ >  \widehat{r}_t\} ~ \Big| ~ x_t, \widehat{r}_t \right] \,,
\end{align*}  so Equation (\ref{eq:strat:boundR_t^11.5}) yields
 \begin{align}
\label{eq:strat:singleperiodregret1}
     \mathcal{R}_t^{(1)}  ~ \leq ~ & \expect\left[\widetilde{\mathcal{R}}_t^{(1)} \right] + \frac{v_{\max}}{|E_\ell|}\,,
 \end{align}
where the expectation is taken with respect to the context $x_t$ and reserve price $\widehat{r}_t$. Notice that $\max\{v_t^-,r_t^\star \}\I\{v_t^+ > r_t^\star\} - \max\{v_t^-, \widehat{r}_t \}\I\{v_t^+ >  \widehat{r}_t\}$ is exactly the revenue difference $\rev_t(r_t^\star) - \rev_t(r_t)$ had the seller set reserve prices $r_t^\star$ or $r_t$ when all buyers bid truthfully. Hence,  by applying 
% similar to the Equations (\ref{eq:full:R_t0}) and (\ref{eq:full:reg2}) in the proof of Theorem \ref{bound:fullInfoRegret}, by defining $y_t := \langle \beta, x_t \rangle $, $\widehat{y}_t := \langle \widehat{\beta}_\ell, x_t \rangle $ and $\rho_t({r},y,F^{(1)},F^{(2)}) := \int_{0}^{r} F^{(2)}({z}-y)d{z} - r \left[ F^{(1)}({r}-y)\right]$ (Equation (\ref{eq:defrho})), we can apply
Proposition \ref{prop:benchmark} we obtain
\begin{align*}
     \widetilde{\mathcal{R}}_t^{(1)}  ~=~ & \int_{0}^{r_{t}^{\star}} F^-({z}-\langle \beta, x_t \rangle)d{z} - r_{t}^{\star} \left[ F^+(r_{t}^{\star}-\langle \beta, x_t \rangle)\right] \\
     &\quad  - \int_{0}^{\widehat{r}_t} F^-({z}-\langle \beta, x_t \rangle)d{z} + \widehat{r}_t\left[ F^+(\widehat{r}_t-\langle \beta, x_t \rangle)\right] \,.
\end{align*}
%  where the expectation is taken with respect to $$ $\left\{\epsilon_{i,t}\right\}_{i\in [N]}$.
%  Then, similar to Equation (\ref{eq:full:reg2}) in the proof of Theorem \ref{bound:fullInfoRegret}, by adding and subtracting terms we have, 
{Note that we can apply Proposition \ref{prop:benchmark} because $\widehat{r}_t$ is the reserve price set according to the NPAC-S policy when no isolation occurs, and only depends on the current context $x_t$ and the past $\mathcal{H}_{t-1}=\{(r_1, b_1,x_1), (r_2, b_2,x_2), \ldots, (r_{t-1}, b_{t-1},x_{t-1})\}$.}

By defining $y_t := \langle \beta, x_t \rangle $, $\widehat{y}_t := \langle \widehat{\beta}_\ell, x_t \rangle $ and

\begin{align}
\label{eq:defrho}
    \rho_t({r},y,F^{(1)},F^{(2)}) := \int_{0}^{r} F^{(2)}({z}-y)d{z} - r \left[ F^{(1)}({r}-y)\right]\,,
\end{align}  
we can rewrite $\widetilde{\mathcal{R}}_t^{(1)}$ as the following: 
\begin{align}
\label{eq:strat:boundR_t^12}
\begin{aligned}
   \widetilde{\mathcal{R}}_t^{(1)} ~ = ~ & \expect\left[\max\{v_t^-,r_t^\star \}\I\{v_t^+ > r_t^\star\} - \max\{v_t^-, \widehat{r}_t \}\I\{v_t^+ >  \widehat{r}_t\} ~\Big|~ x_t, \widehat{r}_t \right] \\ 
    ~ = ~ & \rho_t(r_t^\star, y_t, F^-, F^+) - \rho_t(\widehat{r}_t,y_t,F^-, F^+) \\
  ~ = ~ & \rho_t(r_t^\star, y_t, F^-, F^+) - \rho_t(r_t^\star, \widehat{y}_t,F^-, F^+)  \\
  & ~~ + \rho_t(r_t^\star, \widehat{y}_t, F^-, F^+) - \rho_t(r_t^\star, \widehat{y}_t, \widehat{F}_{\ell+1}^-, \widehat{F}_{\ell+1}^+) \\
  & ~~ + \rho_t(r_t^\star, \widehat{y}_t, \widehat{F}_{\ell+1}^-, \widehat{F}_{\ell+1}^+) - \rho_t(\widehat{r}_t, \widehat{y}_t, \widehat{F}_{\ell+1}^-, \widehat{F}_{\ell+1}^+) \\
   & ~~ + 
  \rho_t(\widehat{r}_t, \widehat y_t, \widehat{F}_{\ell+1}^-, \widehat{F}_{\ell+1}^+) - \rho_t(\widehat{r}_t, \widehat{y}_t, F^-, F^+)  \\
   & ~~ + \rho_t(\widehat{r}_t, \widehat{y}_t , F^-, F^+) - \rho_t(\widehat{r}_t,y_t,F^-, F^+) \,. 
 \end{aligned}
\end{align}

We now invoke  Lemma \ref{lemma:strat:controlUncert}, where we show that when events $\xi_{\ell+1}$, $\xi_{\ell+1}^-$ and $\xi_{\ell+1}^+$ (see definition in Equation (\ref{eq:strat:defxi}),(\ref{eq:strat:defdelta}), (\ref{eq:strat:defxi-}) and (\ref{eq:strat:defxi+}) ) happen for some phase $\ell \geq 1$, we have for $r \in \{r_t^\star, \widehat{r}_t\}$,
\begin{enumerate}
    \item [(i)] $\left|\rho_t(r, y_t, F^-, F^+) - \rho_t(r, \widehat{y}_t, F^-, F^+)\right| ~ \leq ~ 3r c_f N^2  \delta_{\ell}   $ ~~ a.s. 
    \item [(ii)] $\left|\rho_t(r, \widehat{y}_t , F^-, F^+) - \rho_t(r, \widehat{y}_t, \widehat{F}_{\ell+1}^-, \widehat{F}_{\ell+1}^+) \right|~ \leq ~  3rN^2\left(\gamma_{\ell} + c_f \delta_{\ell } + \frac{c_f + L_\ell}{|E_\ell|} \right)$ ~~  a.s.
\end{enumerate}
 Note that the first inequality bounds the impact of errors $\beta$ and the second bounds the impact of errors in the distributions.  Applying these bounds in (\ref{eq:strat:boundR_t^12}), we get
\begin{align}
   \widetilde{\mathcal{R}}_t^{(1)} \cdot \I\left\{\xi_{\ell+1} \cap \xi_{\ell+1}^- \cap \xi_{\ell+1}^+ \right\}  ~ \leq ~ &  3(r_t^\star + \widehat{r}_t) c_f N^2 \delta_{\ell} 
    \nonumber \\ 
    & ~~ + 3(r_t^\star + \widehat{r}_t)N^2\left(\gamma_{\ell} + c_f \delta_{\ell } + \frac{c_f + L_\ell}{|E_\ell|} \right)\nonumber \\
    & ~~ + \rho_t(r_t^\star, \widehat{y}_t, \widehat{F}_{\ell+1}^-, \widehat{F}_{\ell+1}^+) - \rho_t(\widehat{r}_t, \widehat{y}_t, \widehat{F}_{\ell+1}^-, \widehat{F}_{\ell+1}^+)  \,. \label{eq:strat:boundR_t^13}
\end{align}

We recall that the seller's pricing decision $\widehat{r}_t$  when no isolation occurs is defined in Equation (\ref{eq:strat:reserve_price_1}), and realize that in fact $\widehat{r}_t = \arg\max_{r\in(0, v_{\max}]}  \rho_t(r,\widehat{y}_t, \widehat{F}_{\ell+1}^-, \widehat{F}_{\ell+1}^+)$. So, by the optimality of $\widehat{r}_t$ and $r_t^\star \leq v_{\max}$, we obtain the fact that $\rho_t(r_t^\star, \widehat{y}_t, \widehat{F}_{\ell+1}^-, \widehat{F}_{\ell+1}^+) - \rho_t(\widehat{r}_t, \widehat{y}_t, \widehat{F}_{\ell+1}^-, \widehat{F}_{\ell+1}^+)  \leq 0$. Using this inequality in (\ref{eq:strat:boundR_t^13}), we get
\begin{align}
\label{eq:strat:singleperiodregret2}
   & \widetilde{\mathcal{R}}_t^{(1)} \cdot \I\left\{\xi_{\ell+1} \cap \xi_{\ell+1}^- \cap \xi_{\ell+1}^+ \right\}   \nonumber \\
    ~ \leq ~ & 6v_{\max} c_f N^2 \delta_{\ell} +  6v_{\max}N^2\left(\gamma_{\ell} + c_f \delta_{\ell } + \frac{c_f + L_\ell}{|E_\ell|} \right) \nonumber \\
    ~ = ~ & 12v_{\max} c_f N^2 \delta_{\ell} +6v_{\max}N^2\left(\frac{\sqrt{\log(|E_\ell|)}}{\sqrt{2N|E_\ell|}}  + \frac{ c_f + L_\ell}{|E_\ell|} \right) \nonumber\\
    ~ = ~ & {12v_{\max} c_f N^2 \delta_{\ell} + \frac{6v_{\max}\sqrt{N^3\log(|E_\ell|)}}{\sqrt{2E_{\ell}}} + \frac{6v_{\max}N^2(c_f + L_\ell)}{|E_\ell|}} %\\
    %~ < ~ & \old{12v_{\max} c_f N^2 \delta_{\ell} +\frac{4v_{\max}\sqrt{N^3}}{\sqrt{|E_\ell|}}\left({\sqrt{\log(|E_\ell|)} + c_f + L_\ell}\right)}
    \,,
\end{align}
where we used the fact that $r_t^\star, \widehat{r}_t \leq v_{\max}$ in the inequality. %\new{(Delete this), and $|E_\ell| >\sqrt{2|E_\ell|} > \sqrt{|E_\ell|}$ as well as $|E_\ell| \geq |E_1| = \sqrt{T} \geq \sqrt{N}$ in the second inequality.}
Note that $L_\ell = {\log\left({v_{\max}^2N|E_{\ell}|^4} -1 \right)}/{\log(\frac{1}{\eta})} = \mathcal{O}\left(\log(T)/\log({1}/{\eta})\right)$, since we recall that $|E_\ell| = T^{1-2^{-\ell}}$.

To complete the bound for $\mathcal{R}_t^{(1)} $ in period $t \in E_{\ell + 1}$, we continue to bound Equation (\ref{eq:strat:singleperiodregret1}):
 \begin{align}
 \label{eq:strat:singleperiodregret3}
     \mathcal{R}_t^{(1)}  ~ \leq ~ & \expect\left[\widetilde{\mathcal{R}}_t^{(1)} \right] + \frac{v_{\max}}{|E_\ell|}\nonumber \\
      ~ = ~ & \expect\left[\widetilde{\mathcal{R}}_t^{(1)} \cdot \I\left\{\xi_{\ell+1} \cap \xi_{\ell+1}^- \cap \xi_{\ell+1}^+ \right\} \right] + \expect\left[\widetilde{\mathcal{R}}_t^{(1)} \cdot \I\left\{\xi_{\ell+1}^c \cup \left(\xi_{\ell+1}^-\right)^c \cup \left(\xi_{\ell+1}^+\right)^c \right\} \right] + \frac{v_{\max}}{|E_\ell|}\nonumber \\
     ~ \leq ~ & \expect\left[\widetilde{\mathcal{R}}_t^{(1)} \cdot \I\left\{\xi_{\ell+1} \cap \xi_{\ell+1}^- \cap \xi_{\ell+1}^+ \right\} \right] + v_{\max}\prob\left(\xi_{\ell+1}^c \cup \left(\xi_{\ell+1}^-\right)^c \cup \left(\xi_{\ell+1}^+\right)^c  \right)+ \frac{v_{\max}}{|E_\ell|}\nonumber \\
     ~ \leq ~ & {
       12v_{\max} c_f N^2 \delta_{\ell} + \frac{6v_{\max}\sqrt{N^3\log(|E_\ell|)}}{\sqrt{2E_{\ell}}} + \frac{v_{\max}\left(6N^2(c_f + L_\ell)+ 9N + 15d + 9\right)}{|E_\ell|}  }\,,
    %   ~ \leq ~ &  12v_{\max} c_f N^2 \delta_{\ell} + \old{ \frac{4v_{\max}\sqrt{N^3}}{\sqrt{|E_\ell|}}\left({\sqrt{\log(|E_\ell|)} + c_f + L_\ell}\right)} + \frac{v_{\max}(9N + 15d + 9) }{|E_\ell|}  \,,
 \end{align}
where the second inequality follows from a simple observation that $\widetilde{\mathcal{R}}_t^{(1)} \leq v_{\max}$ almost surely, and the third inequality uses Equation (\ref{eq:strat:singleperiodregret2}) and Lemma \ref{lemma:strat:boundprobabilities1}, which shows $\prob\left(\xi_{\ell+1}^c \cup \left(\xi_{\ell+1}^-\right)^c \cup \left(\xi_{\ell+1}^+\right)^c \right) ~ \leq ~ (9N + 15d + 8)/|E_\ell|$,

\textbf{(iii) Bounding $\mathcal{R}_t^{(2)}$:}
So far, we have bounded $ \mathcal{R}_t^{(1)}$ for $t\in E_{\ell +1}$ ($\ell \geq 1$), and will move on to bound $ \mathcal{R}_t^{(2)}$ defined in Equation (\ref{eq:strat:boundR_t}) for $t\in E_{\ell}$ for any $\ell \geq 1$ . 
% First of all, in light of the definition of seller's revenue in Equation \ref{eq:revenue}, under the NPAC-S policy described in Algorithm \ref{algo:strat}, we have
% \begin{align}
%     \rev_t(r_t) 
%   ~ = ~ & \expect\left[\max\{b_t^-,\widehat{r}_t \}\I\{b_t^+ > \widehat{r}_t\} \I\{\text{no isolation in }t\} + \sum_{i\in[N]} r_t^u\I\{b_{i,t} > r_t^u\}\I\{\text{$i$ is isolated}\}  ~ \Big| ~ x_t, r_t \right] 
%     \,.
% \end{align}
We define 
\begin{align}
\label{def:exludinghighestbidandval}
    b_{-i,t}^+ = \max_{j\neq i} b_{j,t} ~~ \text{ and } ~~ v_{-i,t}^+ = \max_{j\neq i} v_{j,t}\,,
\end{align}
which represent the highest bid excluding that of buyer $i$, and the highest valuation excluding that of buyer $i$, respectively. We then have
\begin{align}
\label{eq:strat:boundR_t^21}
\mathcal{R}_t^{(2)} ~ = ~ &\expect\left[\max\{v_t^-, \widehat{r}_t \}\I\{v_t^+ >  \widehat{r}_t\}\I\{\text{no isolation in }t\}  - \rev_t(r_t)\right]  \nonumber \\
    ~ \leq ~ & \expect\left[\max\{v_t^-, \widehat{r}_t \}\I\{v_t^+ >  \widehat{r}_t\}\I\{\text{no isolation in }t\}\right]  - \expect\left[\max\{b_t^-,r_t \}\I\{b_t^+ > \widehat{r}_t\}\I\{\text{no isolation in }t\}\right] \nonumber\\
     ~ = ~ & \left( \expect\left[\max\{v_t^-, \widehat{r}_t \}\I\{v_t^+ >  \widehat{r}_t\}\right] - \expect\left[\max\{b_t^-,\widehat{r}_t \}\I\{b_t^+ > \widehat{r}_t\}\right]\right)\cdot \left( 1 - \frac{1}{|E_\ell|}\right)  \nonumber\\
    ~ < ~ & \expect\left[\max\{v_t^-,\widehat{r}_t \}\I\{v_t^+ > \widehat{r}_t\}\right] - \expect\left[\max\{b_t^-,\widehat{r}_t \}\I\{b_t^+ > \widehat{r}_t\}\right]  \nonumber\\
     ~ = ~ & \sum_{i\in[N]} \expect\left[\max\{v_t^-,\widehat{r}_t \}\I\{v_{i,t} > \max\{v_{-i,t}^+, \widehat{r}_t\}\} - \max\{b_t^-,\widehat{r}_t \}\I\{b_{i,t} > \max\{b_{-i,t}^+ \widehat{r}_t\}\}\right]\nonumber\\
      ~ = ~ & {\sum_{i\in[N]} \expect\left[\max\{v_t^-,\widehat{r}_t \}\I\{ \max\{v_{-i,t}^+, \widehat{r}_t\}< v_{i,t} <\max\{b_{-i,t}^+ \widehat{r}_t\}\} \right] }\nonumber\\
      & ~~ { - \sum_{i\in[N]}\expect\left[\max\{v_t^-,\widehat{r}_t \}\I\{ \max\{b_{-i,t}^+ \widehat{r}_t\} < v_{i,t} <\max\{v_{-i,t}^+, \widehat{r}_t\} \}\right]} \nonumber\\
      & ~~ {+ \sum_{i\in[N]}\expect\left[\max\{v_t^-,\widehat{r}_t \}\I\{v_{i,t} > \max\{b_{-i,t}^+ \widehat{r}_t\}\} - \max\{b_t^-,\widehat{r}_t \}\I\{b_{i,t} > \max\{b_{-i,t}^+ \widehat{r}_t\}\}\right]} \nonumber\\
       ~ \leq ~ & \sum_{i\in[N]} \expect\left[\max\{v_t^-,\widehat{r}_t \}\I\{ \max\{v_{-i,t}^+, \widehat{r}_t\}< v_{i,t} <\max\{b_{-i,t}^+ \widehat{r}_t\}\} \right]  \nonumber\\
      & ~~ + \sum_{i\in[N]}\expect\left[\max\{v_t^-,\widehat{r}_t \}\I\{v_{i,t} > \max\{b_{-i,t}^+ \widehat{r}_t\}\} - \max\{b_t^-,\widehat{r}_t \}\I\{b_{i,t} > \max\{b_{-i,t}^+ \widehat{r}_t\}\}\right] \nonumber\\
      ~ \leq ~ & \sum_{i\in[N]} v_{\max}\expect\left[\I\{ \max\{v_{-i,t}^+,\widehat{r}_t\}< v_{i,t} <\max\{b_{-i,t}^+ \widehat{r}_t\}\} \right]\nonumber\\
      & ~~ + \sum_{i\in[N]}\expect\left[\max\{v_t^-,\widehat{r}_t \}\I\{v_{i,t} > \max\{b_{-i,t}^+ \widehat{r}_t\}\} - \max\{b_t^-,\widehat{r}_t \}\I\{b_{i,t} > \max\{b_{-i,t}^+ \widehat{r}_t\}\}\right]
      \,,
\end{align}
where the first inequality follows from Equation (\ref{eq:strat:revenue}); the third inequality is due to the fact that $\sum_{i\in[N]}\expect\left[\max\{v_t^-,\widehat{r}_t \}\I\{ \max\{b_{-i,t}^+ \widehat{r}_t\} < v_{i,t} <\max\{v_{-i,t}^+, \widehat{r}_t\} \}\right] \geq 0$; and the last inequality holds because $\max\{v_t^-,\widehat{r}_t \} \leq v_{\max}$. To continue the bound for Equation (\ref{eq:strat:boundR_t^21}), we use the definition of $\mathcal{B}_{i,\ell} := \mathcal{B}_{i,\ell}^s \cup \mathcal{B}_{i,\ell}^o$ in Lemma \ref{lemma:strat:boundedoutcomechanges}, where 
\begin{align}
    & \mathcal{B}_{i,\ell}^s = \left\{t\in E_\ell: \I\left\{v_{i,t} > \{b_{-i,t}^+, \widehat{r}_t\} \right\} = 1 ~ , ~ \I\left\{b_{i,t} > \{b_{-i,t}^+, \widehat{r}_t\} \right\} = 0 \right\}  \nonumber\\
    & \mathcal{B}_{i,\ell}^o = \left\{t\in E_\ell: \I\left\{v_{i,t} > \{b_{-i,t}^+, \widehat{r}_t\} \right\} = 0 ~ , ~  \I\left\{b_{i,t} > \{b_{-i,t}^+, \widehat{r}_t\} \right\} = 1 \right\}  \nonumber \,.
\end{align}
Here, $\mathcal{B}_{i,\ell}^s$ represents the periods during which buyer $i$ could have won the auction had she bid truthfully but in reality lost since she shaded her bid (allocation mismatch due to shading), while $\mathcal{B}_{i,\ell}^o$ represents the periods when buyer $i$ would have lost the auction had she bid truthfully, but instead won the item due to overbidding (allocation mismatch due to overbidding). Hence, for any period $t\in  E_{\ell}/\mathcal{B}_{i,\ell} =  \left\{t\in E_\ell: \I\left\{v_{i,t} >\{b_{-i,t}^+, \widehat{r}_t\} \right\} = \I\left\{b_{i,t} > \{b_{-i,t}^+, \widehat{r}_t\} \right\} \right\}$ (which means in period $t \in E_{\ell}/\mathcal{B}_{i,\ell}$ the outcome for buyer $i$ would not have changed even if she bid truthfully), we have $\I\{v_{i,t} > \max\{b_{-i,t}^+,\widehat{r}_t\}\} = \I\{b_{i,t} > \max\{b_{-i,t}^+, \widehat{r}_t\}\}$. Therefore, defining $\mathcal{B}_{\ell} := \cup_{i\in[N]}\mathcal{B}_{i,\ell}$, we have
\begin{align}
 & \mathcal{R}_t^{(2)} \I\{t\in E_{\ell}/\mathcal{B}_{\ell}\} \nonumber \\
  ~ \leq ~ & \sum_{i\in[N]} v_{\max}\expect\left[\I\{ \max\{v_{-i,t}^+,\widehat{r}_t\}< v_{i,t} <\max\{b_{-i,t}^+ \widehat{r}_t\}\} \right]\nonumber\\
      & ~~ + \sum_{i\in[N]}\expect\left[\max\{v_t^-,\widehat{r}_t \}\I\{v_{i,t} > \max\{b_{-i,t}^+ \widehat{r}_t\}\} - \max\{b_t^-,\widehat{r}_t \}\I\{b_{i,t} > \max\{b_{-i,t}^+ \widehat{r}_t\}\}\right]  \I\{t\in E_{\ell}/\mathcal{B}_{\ell}\} \nonumber\\
    ~ = ~ & \sum_{i\in[N]} v_{\max}\expect\left[\I\{ \max\{v_{-i,t}^+, \widehat{r}_t\}< v_{i,t} <\max\{b_{-i,t}^+ \widehat{r}_t\}\} \right] \nonumber\\
    & ~~ + \sum_{i\in[N]} \expect\left[\left(\max\{v_t^-,\widehat{r}_t \} - \max\{b_t^-,\widehat{r}_t \}\right)\I\{b_{i,t} > \max\{b_{-i,t}^+, \widehat{r}_t\}\}\right] \nonumber \\
    ~ \leq ~ & \sum_{i\in[N]}  v_{\max}\expect\left[\I\{ \max\{v_{-i,t}^+, \widehat{r}_t\}< v_{i,t} <\max\{b_{-i,t}^+, \widehat{r}_t\}\} \right]  + \expect\left[\max\{v_t^-,\widehat{r}_t \} - \max\{b_t^-,\widehat{r}_t \}\right]  \nonumber \\
    ~ \leq ~ & \sum_{i\in[N]}  v_{\max}\expect\left[\I\{ \max\{v_{-i,t}^+, \widehat{r}_t\}< v_{i,t} <\max\{b_{-i,t}^+, \widehat{r}_t\}\} \right]  + \expect\left[ \left(v_t^- - b_t^- \right)^+\right]  \nonumber
      \,.
\end{align}
{The first inequality follows from Equation (\ref{eq:strat:boundR_t^21}); the first equality follows from the fact that $t\in E_{\ell}/\mathcal{B}_{\ell}$; the second inequality holds because $\sum_{i\in [N]}\I\{b_{i,t} > \max\{b_{-i,t}^+ \widehat{r}_t\}\} ~ \leq ~  \sum_{i\in [N]}\I\{b_{i,t} >b_{-i,t}^+\}\} = 1$;} the third inequality applies the fact that $\max\{a,c\} -\max\{b,c\} \leq (a - b)^+$ for any $a,b,c \in \R$. Denoting $i^\star := \arg \max_{i\in[N]} v_{i,t}$, we have 
\begin{align*}
    &\sum_{i\in[N]}  v_{\max}\expect\left[\I\{ \max\{v_{-i,t}^+, \widehat{r}_t\}< v_{i,t} <\max\{b_{-i,t}^+, \widehat{r}_t\}\} \right]  \\
    ~ = ~ & v_{\max}\expect\left[\I\{ \max\{v_{-i^\star,t}^+, \widehat{r}_t\}< v_{i^\star,t} <\max\{b_{-i^\star,t}^+, \widehat{r}_t\}\} \right]\end{align*} since $\I\{ \max\{v_{-i,t}^+, \widehat{r}_t\}< v_{i,t}\} = 0$ if $i\neq i^\star$. Therefore 
\begin{align}
\label{eq:strat:boundR_t^22}
\mathcal{R}_t^{(2)} \I\{t\in E_{\ell}/\mathcal{B}_{\ell}\} 
    ~ \leq ~ v_{\max}\expect\left[\I\{ \max\{v_{-i^\star,t}^+, \widehat{r}_t\}< v_{i^\star,t} <\max\{b_{-i^\star,t}^+, \widehat{r}_t\}\} \right] + \expect\left[ \left(v_t^- - b_t^- \right)^+\right]
      \,,
\end{align}

To bound the first term in Equation (\ref{eq:strat:boundR_t^22}), we again evoke the inequality $\max\{a,c\} -\max\{b,c\} = (a - b)^+$ for any $a,b,c \in \R$ and get $  \max\{b_{-i^\star,t}^+, \widehat{r}_t\} - \max\{v_{-i^\star,t}^+, \widehat{r}_t\} \leq  \left(b_{-i^\star,t}^+ - v_{-i^\star,t}^+ \right)^+$. Hence,
\begin{align}
\label{eq:strat:boundR_t^23}
    & \expect\left[\I\{ \max\{v_{-i^\star,t}^+, \widehat{r}_t\} < v_{i^\star,t} <\max\{b_{-i^\star,t}^+, \widehat{r}_t\}\} \right] \nonumber \\
    ~ \leq ~ &  \expect\left[\I\{ \max\{b_{-i^\star,t}^+, \widehat{r}_t\} - \left(b_{-i^\star,t}^+ - v_{-i^\star,t}^+ \right)^+ < v_{i^\star,t} <\max\{b_{-i^\star,t}^+, \widehat{r}_t\}\} \right] \nonumber \\
    ~ = ~ &  \expect\left[\expect\left[\I\{ \max\{b_{-i^\star,t}^+, \widehat{r}_t\} - \left(b_{-i^\star,t}^+ - v_{-i,t}^+ \right)^+ < v_{i^\star,t} <\max\{b_{-i^\star,t}^+, \widehat{r}_t\}\}  ~ \Big| ~ b_{-i^\star,t}^+ , v_{-i^\star,t}^+ \right] \right] \nonumber \\
    ~ = ~ &  \expect\left[\int_{\max\{b_{-i^\star,t}^+, \widehat{r}_t\} -  \left(b_{-i^\star,t}^+ - v_{-i^\star,t}^+ \right)^+ - \langle \beta, x_t \rangle}^{\max\{b_{-i^\star,t}^+, \widehat{r}_t\}- \langle \beta, x_t \rangle } f({z})d{z} \right] \nonumber \\
     ~ \leq ~ &  c_f  \expect\left[\left(b_{-i^\star,t}^+ - v_{-i^\star,t}^+ \right)^+ \right] \,.
\end{align}
{Now, set $j\in[N]$ such that $b_{-i^\star,t}^+ = b_{j,t}$ ($j\neq i^\star$), i.e. $j$ is the highest bidder among all buyers excluding $i^\star$. Then $b_{-i^\star,t}^+ - v_{-i^\star,t}^+ = b_{j,t} - v_{-i^\star,t}^+ \leq  b_{j,t} - v_{j,t} = -a_{j,t}$, where the inequality follows from the fact that $v_{-i^\star,t}^+$ is the highest valuation among all buyers excluding $i^\star$ (which includes $j$ as $j\neq i^\star$).} Therefore,  continuing the bound in Equation (\ref{eq:strat:boundR_t^23}), we have
\begin{align}
\label{eq:strat:boundR_t^24}
    \expect\left[\I\{ \max\{v_{-i^\star,t}^+ , \widehat{r}_t\} < v_{i^\star,t} <\max\{b_{-i^\star,t}^+, \widehat{r}_t\}\} \right] ~ \leq ~ c_f (-a_{j,t})^+  ~ \leq ~ c_f \sum_{i\in [N]} (-a_{i,t})^+ \,.
\end{align}

To bound the second term in Equation (\ref{eq:strat:boundR_t^22}), namely $\expect\left[ \left(v_t^- - b_t^- \right)^+\right]$, without loss of generality assume $v_{1,t} \geq v_{2,t} \geq \dots \geq v_{N,t}$. Hence $v_t^- = v_{2,t}$. If $b_{2,t} \leq b_t^-$, we have $v_t^- - b_t^- \leq v_{2,t} - b_{2,t} = a_{2,t}$. Otherwise if $b_{2,t} > b_t^-$, then buyer $2$ submitted the highest bid, so $b_{i,t} \leq b_t^-$ for any $i \neq 2$ and thus,   $v_t^- - b_t^- \leq v_{1,t} - b_t^-  \leq v_{1,t} - b_{1,t} = a_{1,t}$. Hence,
\begin{align}{}
\label{eq:strat:boundR_t^25}
   \expect\left[ \left(v_t^- - b_t^- \right)^+\right] \leq  \max_{j\in [N]}(a_{j,t})^+ \leq \sum_{j\in [N]} (a_{j,t})^+\,.
\end{align}

 Finally, combining Equations (\ref{eq:strat:boundR_t^22}), (\ref{eq:strat:boundR_t^24}), and (\ref{eq:strat:boundR_t^25}), we have for any $t\in E_{\ell}$ and $\ell \geq 1$
 \begin{align}
 \label{eq:strat:boundR_t^26}
    \mathcal{R}_t^{(2)} \I\{t\in E_{\ell}/\mathcal{B}_{\ell}\} \leq v_{\max} c_f \sum_{i\in [N]} (-a_{i,t})^+ + \sum_{i\in [N]} (a_{i,t})^+  \leq  (v_{\max} c_f + 1) \sum_{i\in [N]} |a_{i,t}|
\end{align}
  \textbf{iv. Bounding Cumulative Regret:} We now bound the cumulative expected regret in a phase $E_{\ell + 1}$ ($\ell \geq 1$) via first bounding $\sum_{t\in E_{\ell+1}}  \mathcal{R}_t^{(1)}$ and $\sum_{t\in E_{\ell+1}}  \mathcal{R}_t^{(2)}$ respectively. 
 \begin{align}
 \label{eq:strat:boundregretinphase1}
     & \sum_{t\in E_{\ell+1}} \mathcal{R}_t^{(1)} \nonumber\\
    ~ \leq ~ & \sum_{t\in E_{\ell+1}} \left(   12v_{\max} c_f N^2 \delta_{\ell} + \frac{6v_{\max}\sqrt{N^3\log(|E_\ell|)}}{\sqrt{2E_{\ell}}} + \frac{v_{\max}\left(6N^2(c_f + L_\ell)+ 9N + 15d + 9\right)}{|E_\ell|}  \right) \nonumber \\
    ~ = ~ &  |E_{\ell+1}|\left( 12v_{\max} c_f N^2 \delta_{\ell} + \frac{6v_{\max}\sqrt{N^3\log(|E_\ell|)}}{\sqrt{2E_{\ell}}} + \frac{v_{\max}\left(6N^2(c_f + L_\ell)+ 9N + 15d + 9\right)}{|E_\ell|}  \right) \nonumber\\ 
     ~ = ~ &  |E_{\ell+1}|\cdot \frac{3v_{\max}\sqrt{2N^3\log(|E_{\ell}|)}}{\sqrt{|E_{\ell}|}} \left(\frac{4c_f \epsilon_{\max} x_{\max}^2 \sqrt{d}}{\lambda_0^2} +1 \right) \nonumber\\
      & ~~ + \frac{|E_{\ell+1}|}{|E_\ell|}\left( \frac{12v_{\max} c_f N^2\sqrt{d}\left(NL_\ell a_{\max} + 1 \right)x_{\max}^2}{\lambda_0^2} + {v_{\max}\left(6N^2(c_f + L_\ell)+ 9N + 15d + 9\right)} \right)\nonumber \\
      ~ \leq ~ & c_{1}^{1}c_f \sqrt{dTN^3\log(|E_{\ell}|)} + c_2^2 c_f \sqrt{d}N^3 L_\ell T^{\frac{1}{4}} \nonumber \\
     ~ \leq ~ & \ce c_f \sqrt{dN^3 \log(|E_{\ell}|) }\left( \sqrt{T} + \frac{\sqrt{N^3\log(|E_{\ell}|)}T^{\frac{1}{4}}}{\log\left(1/\eta\right)} \right) 
     \,,
 \end{align}
for some absolute constants $c_{1}^{1}, c_{1}^{2}, \ce > 0$. The first inequality follows from Equation (\ref{eq:strat:singleperiodregret3}).  In the second equality, we then used the definition of $\delta_{\ell}= \frac{\sqrt{2d\log(|E_\ell|)}\epsilon_{\max} x_{\max}^2}{\lambda_0^2\sqrt{N|E_\ell|}} + \frac{\sqrt{d}\left(NL_\ell a_{\max} + 1 \right)x_{\max}^2}{|E_\ell|\lambda_0^2}$, defined in Equation (\ref{eq:strat:defdelta}). In the second inequality, we relied on the construction of the length of phases in Algorithm \ref{algo:strat}, {i.e. $|E_\ell| = T^{1-2^{-\ell}}$ so that $|E_{\ell+1}|/\sqrt{|E_\ell|} = \sqrt{T}$ and $|E_{\ell+1}|/|E_\ell| = T^{2^{-(\ell+1)}} \leq T^{\frac{1}{4}}$}. The last inequality follows from the fact that $L_\ell = {\log\left({v_{\max}^2N|E_{\ell}|^4} -1 \right)}/{\log(\frac{1}{\eta})}$.
 
 On the other hand, to bound $\sum_{t\in E_{\ell+1}}  \mathcal{R}_t^{(2)}$, we again utilize the definition of $\mathcal{B}_{i,\ell} := \mathcal{B}_{i,\ell}^s \cup \mathcal{B}_{i,\ell}^o$ and $\mathcal{B}_{\ell} := \cup_{i\in[N]} \mathcal{B}_{i,\ell}$ where $\mathcal{B}_{i,\ell}^s$ and $ \mathcal{B}_{i,\ell}^o$ are defined in Equation (\ref{eq:strat:boundedoutcomechanges0}) of Lemma \ref{lemma:strat:boundedoutcomechanges}. Denote $K_{\ell+1} :=  2L_{\ell+1} + 4c_f + 8\log(|E_{\ell+1}|)$. Then, we have
%  $L_\ell = {\log\left({v_{\max}^2N|E_{\ell}|^4} -1 \right)}/{\log(\frac{1}{\eta})} = \mathcal{O}\left(\log(T)/\log({1}/{\eta})\right)$. 
 \begin{align}
   \label{eq:strat:boundregretinphase2}
      \sum_{t\in E_{\ell+1}}  \mathcal{R}_t^{(2)}
    ~ = ~ &  \expect\left[\sum_{t\in\mathcal{B}_{\ell+1}}  \mathcal{R}_t^{(2)}\right] + \expect\left[\sum_{t\in E_{\ell+1}/\mathcal{B}_{\ell+1}}  \mathcal{R}_t^{(2)} \right]\nonumber\\
    ~ \leq ~ &  v_{\max}\expect\left[|\mathcal{B}_{\ell+1}|\cdot \I\{ |\mathcal{B}_{\ell+1}| \leq N K_{\ell+1}\right] +  v_{\max} \expect\left[|\mathcal{B}_{\ell+1}| \cdot \I\{ |\mathcal{B}_{\ell+1}| > NK_{\ell+1}\}\right] \nonumber \\
    & ~~ + (v_{\max} c_f + 1)\expect\left[\sum_{t\in E_{\ell+1}/\mathcal{B}_{\ell+1}}  \sum_{i\in [N]} |a_{i,t}| \right]\nonumber \\
    ~ \leq ~ &  v_{\max}N K_{\ell+1} +  v_{\max}|E_{\ell + 1}| \cdot \prob\left(|\mathcal{B}_{\ell+1}| > N K_{\ell+1}\right) + (v_{\max} c_f + 1)\expect\left[\sum_{t\in E_{\ell+1}/\mathcal{B}_{\ell+1}}  \sum_{i\in [N]} |a_{i,t}| \right]\nonumber \\
     ~ \leq ~ &  v_{\max}N K_{\ell+1} +  4v_{\max}N+ (v_{\max} c_f + 1)\expect\left[\sum_{t\in E_{\ell+1}/\mathcal{B}_{\ell+1}}  \sum_{i\in [N]} |a_{i,t}| \right] \nonumber \\
     ~ \leq ~ &  v_{\max}N (K_{\ell+1}+4)  + (v_{\max} c_f + 1)\expect\left[\sum_{t\in E_{\ell+1}}  \sum_{i\in [N]} |a_{i,t}| \right] \,,
 \end{align}
 where the first inequality follows from Equation (\ref{eq:strat:boundR_t^26}) and uses the fact that $\mathcal{R}_t^{(2)}\leq v_{\max}$; the second inequality is because $|\mathcal{B}_{\ell+1}|\leq |E_{\ell+1}|$; the third inequality applies Lemma \ref{lemma:strat:boundedoutcomechanges} which shows $\prob\left(|\mathcal{B}_{i,\ell+1}| > K_{\ell+1}\right) \leq 4/|E_{\ell+1}|$, {and hence $\prob\left(|\mathcal{B}_{\ell+1}| \leq N K_{\ell+1}\right) \geq  \prob\left(\cap_{i\in[N]}\left\{|\mathcal{B}_{i,\ell+1}| \leq K_{\ell+1} \right\}\right) \geq 1 - 4N/|E_{\ell+1}|$}.  To bound $\expect\left[\sum_{t\in E_{\ell+1}} \sum_{i\in [N]} |a_{i,t}| \right]$, we recall $\mathcal{S}_{\ell+1} := \cup_{i\in[N]}\mathcal{S}_{i,\ell+1}$ where $\mathcal{S}_{i,\ell+1} $ is defined in Equation (\ref{eq:strat:defindividualbigcorruptions}), and consider the following
 \begin{align}
  \label{eq:strat:boundregretinphase3}
      \expect\left[\sum_{t\in E_{\ell+1}} \sum_{i\in [N]} |a_{i,t}| \right] 
     ~ \leq ~ &  \expect\left[\sum_{t\in\mathcal{S}_{\ell+1}} \sum_{i\in [N]} |a_{i,t}| \right] +  \expect\left[\sum_{t\in E_{\ell+1}/\mathcal{S}_{\ell+1}} \sum_{i\in [N]} \frac{1}{|E_{\ell+1}|}\right]\nonumber \\
     ~ \leq ~ & N a_{\max}\expect\left[|\mathcal{S}_{\ell+1}|\right] +  N \nonumber  \\
      ~ = ~ & 
      N a_{\max}\expect\left[|\mathcal{S}_{\ell+1}| \cdot \left(\I\{|\mathcal{S}_{\ell+1}| \leq N L_{\ell+1} \} + \I\{ |\mathcal{S}_{\ell+1}| > NL_{\ell+1} \} \right)\right] +  N \nonumber\\
     ~ \leq ~ & N a_{\max}\left( NL_{\ell+1} + |E_{\ell+1}| \cdot \prob\left(|\mathcal{S}_{\ell+1}| > NL_{\ell+1} \right)\right)  +  N\nonumber \\
     ~ \leq ~ & {N^2 a_{\max}\left( L_{\ell+1} + 1\right)  +  N} \,,
 \end{align}
 where the first inequality holds because $|a_{i,t}| \leq 1/ |E_{\ell+1}|$ for all $t\in E_{\ell+1}/\mathcal{S}_{\ell+1}$ and the fourth inequality follows from Lemma \ref{lemma:strat:boundedindividualbiglies} that shows $\prob\left(\left| \mathcal{S}_{i,\ell+1}\right| > L_{\ell+1} \right) \leq {1}/{|E_{\ell+1}|}$, 
 {which implies $\prob\left(\left| \mathcal{S}_{\ell+1}\right| \leq NL_{\ell+1} \right) \geq \prob\left(\cap_{i\in[N]} \left\{\left| \mathcal{S}_{i,\ell+1}\right| \leq L_{\ell+1} \right\}\right) \geq 1 - {N}/{|E_{\ell+1}|}$}.
 
 Hence, Equations (\ref{eq:strat:boundregretinphase2}) and (\ref{eq:strat:boundregretinphase3}) show that $\sum_{t\in E_{\ell+1}}  \mathcal{R}_t^{(2)}$ is upper bounded as
 \begin{align}
     \sum_{t\in E_{\ell+1}}  \mathcal{R}_t^{(2)} ~ \leq ~ &
     v_{\max}N (K_\ell+4)  + (v_{\max} c_f + 1)\left( N^2 a_{\max}\left( L_{\ell+1} + 1\right)  +  N\right) \nonumber \\
      ~ \leq ~ & \ct c_f N^2 \cdot \frac{\log(|E_{\ell+1}|)}{\log\left(1/\eta\right)} \,,
 \end{align}
 {for some absolute constant $\ct > 0$.} Combining this with the upper bound $${\ce c_f \sqrt{dN^3 \log(|E_{\ell}|) }\left( \sqrt{T} + \frac{\sqrt{N^3\log(|E_{\ell}|)}T^{\frac{1}{4}}}{\log\left(1/\eta\right)} \right)}$$
 shown in Equation (\ref{eq:strat:boundregretinphase1}), the expected cumulative regret in phase $E_{\ell+1}$ is 
  {
 \begin{align*}
       \sum_{t\in E_{\ell+1}} \reg_t  ~ \leq ~ \cth c_f \sqrt{dN^3 \log(T) }\left( \sqrt{T} + \frac{\sqrt{N^3 \log(T)}T^{\frac{1}{4}}}{\log\left(1/\eta\right)} \right)\,,
 \end{align*}}
%  \old{
%  \begin{align*}
%       \sum_{t\in E_{\ell+1}} \reg_t  \leq \frac{\cth c_f d N^3 \log(T)\sqrt{T}}{\log({1}/{\eta})}\,,
%  \end{align*}}
 for some absolute constant $\cth > 0$. Finally, since the total number of phases is upper bounded by $\lceil \log\log(T) \rceil +1$, the cumulative expected regret over the entire horizon $T$ is
  {
  \begin{align*}
      \reg(T)  ~ \leq ~ & v_{\max}|E_1| + \sum_{\ell = 2}^{\lceil \log\log(T) \rceil} \cth c_f \sqrt{dN^3 \log(T) }\left( \sqrt{T} + \frac{\sqrt{N^3 \log(T)}T^{\frac{1}{4}}}{\log\left(1/\eta\right)} \right)\\
     ~ = ~ & \mathcal{O}\left(c_f \sqrt{dN^3 \log(T) }\cdot \log\left(\log(T)\right) \left( \sqrt{T} + \frac{\sqrt{N^3 \log(T)}T^{\frac{1}{4}}}{\log\left(1/\eta\right)} \right)\right)\,.
 \end{align*}}
%  \old{
%   \begin{align*}
%       \reg(T)  ~ \leq ~ & v_{\max}|E_1| + \sum_{\ell = 2}^{\cz\lceil \log\log(T) \rceil +1} \cth c_f d L_{\ell}N^3 \log(T)\sqrt{T}\\
%      ~ = ~ & \mathcal{O}\left( \frac{c_f dN^3 \log\left(\log(T)\right) \log(T)\sqrt{T}}{\log({1}/{\eta})}\right)\,.
%  \end{align*}}

\subsection{Proof of Lemma \ref{lemma:strat:boundedindividualbiglies}}
\label{appsec:strat:lem1}
% We first characterize the utility $U_{i,t}(b)$ for buyer $i$ submitting a bid of value $b$ under three scenarios, namely when no isolation occurs, when $j\neq i$ is isolated, and when $i$ is isolated. Note that this utility function assumes that bids of other (independent) buyers and the reserve price is fixed. We denote the utility of each scenario as  $U_{i,t}^{\text{NT}}$, $U_{i,t}^{\text{Tj}}$ and  $U_{i,t}^{\text{Ti}}$ respectively, and note that 
According to the definitions of the cumulative discounted utility defined in Equation (\ref{eq:strat:utility}) and the NPAC-S policy in Algorithm \ref{algo:strat}, buyer $i$'s utility for submitting a bid $b\in [0,v_{\max}]$ in period $t\in[T]$ conditioning on $v_{i,t}, b_{-i,t}^+, r_t$ is given by
\begin{align}
\label{eq:strat:utility1}
    u_{i,t}(b) = 
    \begin{cases}
    \left(v_{i,t} - \max\{r_t,  b_{-i,t}^+\}\right)\I\{b > \max\{ r_t, b_{-i,t}^+ \}\} & \text{no isolation}\\
     \left(v_{i,t} - r_t\right)\I\{b > r_t\} & \text{$i$ is isolated}\\
     0  & \text{$j\neq i$ is isolated}
    \end{cases}\,,
\end{align}
where $b_{-i,t}^+$ is the highest bid excluding that of buyer $i$, and the reserve price $r_t= \widehat{r}_t \I\{\text{no isolation in }t\}  +  r_t^u(1-  \I\{\text{no isolation in }t\})$ ($\widehat{r}_t$ and $r_t^u$ are defined in Equations (\ref{eq:strat:reserve_price_0}) and (\ref{eq:strat:reserve_price_1}) of the NPAC-S policy respectively). Note that $u_{i,t}(b)$ is a random variable that depends on the $x_t, \{\epsilon_{i,t}\}_{i\in[N]}, b_{-i,t}^+$ and $r_t$. The undiscounted utility loss $u_{i,t}^-$ for buyer $i$ if he submits a bid $b_{i,t}$ compared to bidding truthfully is $u_{i,t}^- = u_{i,t}(v_{i,t}) - u_{i,t}(b_{i,t})$. 

Now, when any buyer $j\neq i$ is isolated, the utility for buyer $i$ is always 0 regardless of what he submits, so there is no utility loss due to bidding behaviour. We now consider the scenarios when no isolation occurs and when buyer $i$ is isolated, respectively, using the definition of utility in Equation (\ref{eq:strat:utility}). 
\begin{itemize}
    \item [] \textbf{No isolation occurs:}
    % The undiscoutnutility for submitting any bid $b\in \R$ is $ \left(v_{i,t} - \max\{r_t, b_{-i,t}^+\}\right)\I\{b > \max\{ r_t, b_{-i,t}^+\}\} $ for any given $r_t$ and $b_{-i, t}^+$, where $ b_{-i,t}^+$ is the highest bid among bids from buyers other than $i$. Hence, 
    The undiscounted utility loss for bidding untruthfully is
    \begin{align}
    \label{eq:strat:notarget}
         u_{i,t}^-\I\{\text{no isolation in }t\}
        ~ = ~ & \left(u_{i,t}(v_{i,t}) - u_{i,t}(b_{i,t})\right)\I\{\text{no isolation in }t\} \nonumber \\
       ~ = ~ & \left(v_{i,t} - \max\{r_t,  b_{-i,t}^+\}\right)\I\{v_{i,t} > \max\{ r_t, b_{-i,t}^+ \}\} \nonumber\\
       & ~~~ - \left(v_{i,t} - \max\{r_t,  b_{-i,t}^+\}\right)\I\{b_{i,t} > \max\{ r_t, b_{-i,t}^+\}\} \nonumber\\
       ~ = ~ & \left|v_{i,t} - \max\{r_t,  b_{-i,t}^+\}\right|\I\{v_{i,t} > \max\{ r_t, b_{-i,t}^+ \}> b_{i,t}\} \nonumber \\ 
       & ~~~ + \left|v_{i,t} - \max\{r_t,  b_{-i,t}^+\}\right|\I\{v_{i,t} < \max\{ r_t, b_{-i,t}^+ \}< b_{i,t}\} \nonumber\\
       ~ \geq ~ & 0 \,.
    \end{align}
    \item [] \textbf{Isolating buyer $i$:} The undiscounted utility for submitting any bid $b\in \R$ for any given $r_t$ is $ \left(v_{i,t} - r_t \right)\I\{b >  r_t\} $. Hence, 
    \begin{align}
    \label{eq:strat:analyzetargeti}
         u_{i,t}^- \I\{\text{$i$ is isolated}\}
         ~ = ~ & \left(u_{i,t}(v_{i,t}) - u_{i,t}(b_{i,t})\right)\I\{\text{$i$ is isolated}\} \nonumber \\
       ~ = ~ & \left(v_{i,t} - r_t\right)\I\{v_{i,t} > r_t\} - \left(v_{i,t} - r_t\right)\I\{ b_{i,t} > r_t\}\nonumber\\
       ~ = ~ & \left(v_{i,t} - r_t \right)\I\{v_{i,t} > r_t > b_{i,t}\}  + \left(-v_{i,t} + r_t\right)\I\{v_{i,t} < r_t < b_{i,t}\}  \,.
    \end{align}
\end{itemize}

The NPAC-S policy offers a price $r_t$  drawn from Uniform$(0,v_{\max})$ to the isolated buyer $i$ with probability $1/|E_{\ell}|$, where $i$ is chosen uniformly among all buyers. So, the expected utility loss $u_{i,t}^-$ for a buyer $i\in[N]$ conditioned on the fact that the buyer lies by an amount of $a_{i,t}$ is 
\begin{align}
\label{eq:strat:lowerboundconditionalutilityloss}
    & \expect[u_{i,t}^- ~|~ a_{i,t}] \nonumber\\
      ~ = ~ & \expect[u_{i,t}^- \I\{\text{$i$ is isolated}\} +  u_{i,t}^- \I\{\text{no isolation in }t\} ~|~ a_{i,t}] \nonumber\\
      ~ \geq ~ & \expect[u_{i,t}^- \I\{\text{$i$ is isolated}\} ~|~ a_{i,t}] \nonumber\\
      ~ = ~ &\frac{1}{N|E_{\ell}|} \expect\left[ \left(v_{i,t} - r_t \right)\I\{v_{i,t} > r_t >b_t\} + \left(-v_{i,t} + r_t\right)\I\{b_t < r_t < v_{i,t}\} ~|~ a_{i,t} \right] \nonumber \\
    %   ~ = ~ &   \frac{\eta^t }{N |E_{\ell}|} \expect[\left(v_{i,t} - r_t\right)\I\{ r_t \leq b_{i,t} < v_{i,t}\}  | b_{i,t}, q_{i,t},v_{i,t}]] \nonumber\\
       ~ = ~ & \frac{1}{v_{\max}N|E_{\ell}|}\expect\left[\expect\left[ \int_{v_{i,t} - a_{i,t}}^{v_{i,t}}(v_{i,t} - r)dr  + \int_{v_{i,t} }^{v_{i,t} + a_{i,t}} (-v_{i,t} + r)dr ~ \Big|~ a_{i,t}, v_{i,t}\right]~ \Big|~ a_{i,t}\right]\nonumber \\ 
        ~ = ~ & \frac{\left(a_{i,t}\right)^2}{v_{\max}N|E_{\ell}|} \,.
    % ~ \geq ~ & \frac{\eta^t (1-q_{i,t})}{v_{\max}N|E_{\ell}|}\int_{b_{i,t}}^{v_{i,t}}(v_{i,t} - r)dr \nonumber \\ 
    % ~ = ~ &  
    %  \frac{\eta^t (1-q_{i,t})}{2v_{\max}N|E_{\ell}|}\int_{v_{i,t} - a^{(k)}}^{v_{i,t}}(v_{i,t} - r)dr \nonumber\\
    %  ~ = ~ & \frac{\eta^t \left(a^{(k)}\right)^2(1-q_{i,t})}{2v_{\max}N|E_{\ell}|}\,.
\end{align}
The first inequality follows from $ u_{i,t}^- \I\{\text{$i$ is isolated}\} \geq 0$ as demonstrated in Equation (\ref{eq:strat:notarget}). Now we lower bound the total expected utility loss in phase $E_{\ell}$. First, by Equations (\ref{eq:strat:notarget}) and (\ref{eq:strat:analyzetargeti}), we know that $u_{i,t}^-\geq 0$ for $\forall i,t$. Therefore, denoting $s_{\ell+1}$ as the first period of phase $E_{\ell+1}$, for any $\tilde{z} >0$ we have
\begin{align}
\label{eq:strat:lowerboundutilityloss}
    \expect\left[\sum_{t\in E_{\ell}} \eta^t u_{i,t}^- \right] ~ \geq ~ & \expect\left[\sum_{t\in \mathcal{S}_{i,\ell}} \eta^t u_{i,t}^-\right]\nonumber \\
     ~ \geq ~ & \expect\left[\sum_{t\in \mathcal{S}_{i,\ell}}\eta^t u_{i,t}^- \I\{\left| \mathcal{S}_{i,\ell}\right| \geq \tilde{z} \}\right]  \nonumber\\
      ~ = ~ & \expect\left[ \expect\left[\sum_{t\in \mathcal{S}_{i,\ell}} \eta^t u_{i,t}^-  ~\Big|~ \{a_{i,t}\}_{t\in E_{\ell}}  \right]  \I\{\left| \mathcal{S}_{i,\ell}\right| \geq \tilde{z} \}\right]\nonumber \\ 
      ~ \geq ~ & \expect\left[ \sum_{t\in \mathcal{S}_{i,\ell}} \frac{\eta^t }{v_{\max}N|E_{\ell}|^3} \cdot \I\{\left| \mathcal{S}_{i,\ell}\right| \geq \tilde{z} \}\right]\nonumber \\
      ~ \geq ~ & \expect\left[ \sum_{t = s_{\ell+1}-\left| \mathcal{S}_{i,\ell}\right| }^{s_{\ell+1}-1} \frac{\eta^t }{v_{\max}N|E_{\ell}|^3} \cdot \I\{\left| \mathcal{S}_{i,\ell}\right| \geq \tilde{z} \}\right]\nonumber\\
       ~ \geq ~ & \expect\left[ \sum_{t = s_{\ell+1}-\tilde{z} }^{s_{\ell+1}-1} \frac{\eta^t }{v_{\max}N|E_{\ell}|^3} \cdot \I\{\left| \mathcal{S}_{i,\ell}\right| \geq \tilde{z} \}\right]\nonumber\\
        ~ = ~ &   \frac{\eta^{s_{\ell+1}}\left( 1 - \eta^{-\tilde{z}}\right) }{(1-\eta) v_{\max}N|E_{\ell}|^3}  \prob\left(\left| \mathcal{S}_{i,\ell}\right| \geq \tilde{z} \right)\,,
\end{align}
where  the first equality holds because $|\mathcal{S}_{i,\ell}| = \sum_{t\in E_{\ell}}\I\{a_{i,t} > 1/ E_{\ell}\}$ is a function of $\{a_{i,t}\}_{t\in E_{\ell}}$; the third inequality follows from Equation (\ref{eq:strat:lowerboundconditionalutilityloss}) and $a_{i,t} \geq {1}/{|E_\ell|}$ for any $t\in \mathcal{S}_{i,\ell}$; and the fourth inequality is because $\eta \in (0,1)$. 

Furthermore, corrupting a bid at time $t\in E_\ell$ will only impact the prices offered by the seller in future phases, i.e., phase $\ell+1, \ell+2, \dots$, so the utility gain due to lying in phase $\ell$, denoted as $U_{i,\ell}^+$ is upper bounded by $v_{\max}\sum_{t \geq s_{\ell+1}} \eta^t = v_{\max}\eta^{s_{\ell+1}}/(1 - \eta)$. Since the buyer is utility maximizing, the net utility gain due to lying in phase $\ell$ should be greater than $0$, otherwise the buyer can choose to always bid $0$ in phase $\ell$ which is equivalent to not participating in the auctions. Hence, 
\begin{align*}
    \expect\left[U_{i,\ell}^+ - \sum_{t\in E_{\ell}}\eta^t u_{i,t}^-\right]\geq 0\,.
\end{align*}
Combining this with $U_{i,\ell}^+ \leq v_{\max}\eta^{s_{\ell+1}}/(1 - \eta)$ and the lower bound for $ \expect\left[\sum_{t\in E_{\ell}} u_{i,t}^- \right] $ shown in Equation (\ref{eq:strat:lowerboundutilityloss}), we have
\begin{align*}
    \frac{v_{\max}\eta^{s_{\ell+1}}}{1 - \eta} \geq  \frac{\eta^{s_{\ell+1}}\left( 1 - \eta^{-\tilde{z}}\right)}{(1-\eta) v_{\max}N|E_{\ell}|^3}  \prob\left(\left| \mathcal{S}_{i,\ell}\right| \geq \tilde{z} \right)\,,
\end{align*}
which holds for any $\tilde{z}>0$. Taking $\tilde{z} = \log\left(v_{\max}^2N|E_{\ell}|^4 -1 \right)/\log(1/\eta)$ and by rearranging terms, the inequality above yields
\begin{align*}
    \prob\left(\left| \mathcal{S}_{i,\ell}\right| \geq \frac{\log\left({v_{\max}^2N|E_{\ell}|^4} -1 \right)}{\log(\frac{1}{\eta})} \right) \leq \frac{1}{|E_{\ell}|}\,.
\end{align*}
\qed

\subsection{Proof of Lemma \ref{lemma:strat:boundedoutcomechanges}}
\label{appsec:strat:lem2}
Defining $\mathcal{H}_{i,t} := \{(b_{-i,\tau}^+, \widehat{r}_\tau, x_\tau)\}_{\tau \in [t]}$, we have \begin{align}
\label{eq:strat:boundedoutcomechanges1}
    & \expect\left[ \I\{t \in (E_\ell/ \mathcal{S}_{i,\ell}) \cap  \mathcal{B}_{i,\ell}^s \} ~ | ~ \mathcal{H}_{i,t}\right] \nonumber \\
   ~ = ~ & \prob\left(t \in (E_\ell/ \mathcal{S}_{i,\ell}) \cap  \mathcal{B}_{i,\ell}^s ~ | ~ \mathcal{H}_{i,t} \right) \nonumber \\
   ~ = ~ & \prob\left( v_{i,t} \geq \max\{b_{-i,t}^+, \widehat{r}_t\}~ , ~ b_{i,t} < \max\{b_{-i,t}^+, \widehat{r}_t\} ~ , ~ a_{i,t} \in (0, 1/|E_\ell|) ~ | ~ \mathcal{H}_{i,t}\right) \nonumber \\
   ~ = ~ & \prob\left( \max\{b_{-i,t}^+, \widehat{r}_t\} - \langle x_t, \beta\rangle \leq \epsilon_{i,t} \leq  \max\{b_{-i,t}^+, \widehat{r}_t\} - \langle x_t, \beta\rangle + a_{i,t} ~ , ~ a_{i,t} \in (0, 1/|E_\ell|) ~ | ~ \mathcal{H}_{i,t}\right) \nonumber \\
   ~ \leq ~ & \prob\left( \max\{b_{-i,t}^+, \widehat{r}_t\} - \langle x_t, \beta\rangle  \leq \epsilon_{i,t} \leq  \max\{b_{-i,t}^+, \widehat{r}_t\} - \langle x_t, \beta\rangle + 1/|E_\ell| ~ | ~ \mathcal{H}_{i,t}\right) \nonumber \\
    ~ = ~ & \expect\left[\int_{\max\{b_{-i,t}^+, \widehat{r}_t\} - \langle x_t, \beta\rangle}^{\max\{b_{-i,t}^+, \widehat{r}_t\} - \langle x_t, \beta\rangle + 1/|E_\ell|} f({z}) d{z} ~ \Big| ~ \mathcal{H}_{i,t}\right]\nonumber \\
    ~ \leq ~ & \frac{c_f}{|E_\ell|}\,.
\end{align}
The last inequality uses the fact that $c_f = \sup_{\tilde{z}\in [-\epsilon_{\max},\epsilon_{\max}]} f(\tilde{z})$. 

Define $\zeta_t = \I\{t \in (E_\ell/ \mathcal{S}_{i,\ell})\cap  \mathcal{B}_{i,\ell}^s \}$ and $\phi_t =  \expect\left[ \I\{t \in (E_\ell/ \mathcal{S}_{i,\ell})\cap  \mathcal{B}_{i,\ell}^s \} ~ | ~ \mathcal{H}_{i,t}\right]$. Then $\expect[\zeta_t ~ | ~ \mathcal{H}_{i,t}] = \phi_t$, which implies $\expect[\zeta_t - \phi_t ~ | ~ \sum_{\tau < t}\zeta_\tau, \sum_{\tau < t}\phi_\tau] =   \expect\left[\expect\left[\zeta_t - \phi_t ~ | ~ \mathcal{H}_{i,t} \right] ~ | ~ \sum_{\tau < t}\zeta_\tau, \sum_{\tau < t}\phi_\tau\right] = 0$. Hence, in the context of the multiplicative Azuma inequality described in Lemma \ref{multiplicativeAzuma}, by setting $z_{1,t} = \zeta_t$, $z_{2,t} = \phi_t$, $\tilde{\gamma} =1/2 $ and $A = 2\log(|E_\ell|)$ we have $|z_{1,t}-z_{2,t}| \leq 1$
\begin{align}
\label{eq:strat:boundedoutcomechanges}
    \prob\left(\frac{1}{2}\sum_{t\in E_\ell} \zeta_t \geq \sum_{t\in E_\ell} \phi_t + 2\log(|E_\ell|) \right) \leq \exp\left( - \log(|E_\ell|)\right) \,.
\end{align}
Now, according to Equation (\ref{eq:strat:boundedoutcomechanges1}), we have $\phi_t\leq c_f/|E_\ell|$, so $\sum_{t\in E_\ell} \phi_t\leq c_f$. Moreover, $| (E_\ell/\mathcal{S}_{i,\ell} )\cap  \mathcal{B}_{i,\ell}^s| = \sum_{t\in E_\ell} \zeta_t$. Hence, following Equation (\ref{eq:strat:boundedoutcomechanges}), we have
\begin{align}
\label{eq:strat:boundedoutcomechanges2}
      \prob\left(| (E_\ell/\mathcal{S}_{i,\ell} )\cap  \mathcal{B}_{i,\ell}^s| \geq 2c_f + 4\log(|E_\ell|) \right) ~ \leq ~ & \prob\left(\frac{1}{2}\sum_{t\in E_\ell} \zeta_t \geq \sum_{t\in E_\ell} \phi_t + 2\log(|E_\ell|) \right) \nonumber \nonumber \\
      ~ \leq ~ & \exp\left( -\log(|E_\ell|)\right) = \frac{1}{|E_\ell|}\,.
\end{align}
When the event $\mathcal{G}_{i,t} = \left\{\left| \mathcal{S}_{i,\ell}\right| \leq L_\ell  \right\}$ occurs, where $L_\ell = {\log\left({v_{\max}^2N|E_{\ell}|^4} -1 \right)}/{\log({1}/{\eta})}$, we have $|\mathcal{B}_{i,\ell}^s| \leq \left| \mathcal{S}_{i,\ell}\right| + | (E_\ell/\mathcal{S}_{i,\ell} )\cap  \mathcal{B}_{i,\ell}^s| \leq L_\ell + | (E_\ell/\mathcal{S}_{i,\ell} )\cap  \mathcal{B}_{i,\ell}^s|$. Therefore when  event $\mathcal{G}_{i,t}$ occurs,
\begin{align*}
    & \prob\left(|\mathcal{B}_{i,\ell}^s| \leq L_\ell + 2c_f + 4\log(|E_\ell|)  \right) \\
    ~ \geq ~ &  \prob\left(\left\{|\mathcal{B}_{i,\ell}^s| \leq L_\ell + 2c_f + 4\log(|E_\ell|) \right\} ~ \bigcap ~ \mathcal{G}_{i,t} \right) \\
      ~ \geq ~ &  \prob\left(\left\{| (E_\ell/\mathcal{S}_{i,\ell} )\cap  \mathcal{B}_{i,\ell}^s| \leq  2c_f + 4\log(|E_\ell|) \right\} ~ \bigcap ~ \mathcal{G}_{i,t} \right) \\
      ~ \geq ~ & 1 - \prob\left(| (E_\ell/\mathcal{S}_{i,\ell} )\cap  \mathcal{B}_{i,\ell}^s| \geq 2c_f + 4\log(|E_\ell|) \right) - \prob\left( \mathcal{G}_{i,t}^c \right) \\
      ~ \geq ~ & 1 -  \frac{2}{|E_\ell|}\,.
\end{align*}
The second inequality follows from $|\mathcal{B}_{i,\ell}^s| \leq L_\ell + | (E_\ell/\mathcal{S}_{i,\ell} )\cap  \mathcal{B}_{i,\ell}^s|$ when the event $\mathcal{G}_{i,t}$ occurs; the third inequality applies the union bound, and the final inequality follows from Equation (\ref{eq:strat:boundedoutcomechanges2}) and Lemma \ref{lemma:strat:boundedindividualbiglies}.

Similarly, we can show the same probability upper bound for $|\mathcal{B}_{i,\ell}^o|$. Finally, using the fact that $\mathcal{B}_{i,\ell} = \mathcal{B}_{i,\ell}^s \cup \mathcal{B}_{i,\ell}^o$ and applying a union bound would yield the desired expression.
\qed

\subsection{Other Lemmas for proving Theorem \ref{thm:strat:Regret}} 
\label{appsec:strat:otherlemmas}

\begin{lemma}[Bounding Estimation Errors in $\beta$]\label{lemma:strat:bias} 
% Assume that the events $\mathcal{G}_{i,\ell}= \left\{\left| \mathcal{S}_{i,\ell}\right| \leq L_\ell  \right\}$ hold for any $i\in [N]$.
For any phase $E_\ell$ and $\gamma > 0$, we have
\begin{align*}
  & \prob\left( \norm{\widehat{\beta}_{\ell+1} - \beta}_1 \leq {\gamma} +  \frac{d\left(NL_\ell a_{\max} + 1 \right)x_{\max}}{|E_\ell|\lambda_0^2}  \right) \\
  ~ \geq ~ & 1 -  2d\exp\left( -\frac{N\gamma^2\lambda_0^4 |E_\ell| }{2{\epsilon_{\max}}^2x_{\max}^2 d}\right)  -  d\exp\left( -\frac{|E_\ell|\lambda_0^2}{8x_{\max}^2}\right) - \frac{N}{|E_\ell|}\,,
\end{align*}{ where $\lambda_0^2$ is the minimum eigenvalue of the covariance matrix $\Sigma$, $\widehat{\beta}_{\ell+1}$ is defined in Equation (\ref{eq:strat:betaestimate}), and $L_\ell = {\log\left({v_{\max}^2N|E_{\ell}|^4} -1 \right)}/{\log({1}/{\eta})}$.} Furthermore, setting $\gamma =  \frac{\sqrt{2d\log(|E_\ell|)}\epsilon_{\max} x_{\max}}{\lambda_0^2\sqrt{N|E_\ell|}}$ and denoting $\delta_\ell = \gamma \cdot x_{\max}+ \frac{d\left(N L_\ell a_{\max} + 1 \right)x_{\max}^2}{|E_\ell|\lambda_0^2}$, we have
\begin{align*}
    \prob\left( \norm{\widehat{\beta}_{\ell+1} - \beta}_1 \leq \frac{\delta_\ell}{x_{\max}} \right) \geq 1 - \frac{2d + N }{|E_\ell|}-  d\exp\left( -\frac{|E_\ell|\lambda_0^2}{8x_{\max}^2}\right)\,.
\end{align*}
\end{lemma}

\textit{Proof of Lemma \ref{lemma:strat:bias}.}
% {The proof of this lemma is based on several modifications to that of Lemma \ref{lemma:full:bias} to resolve the issues that arise when estimating $\beta$ in the presence of corrupted bids submitted by buyers. }

The proof of Lemma \ref{lemma:strat:bias} is inspired by Lemma EC.7.2  in \cite{bastani2015online}, but here we made substantial modifications to resolve the issues that arise when estimating $\beta$ in the presence of corrupted bids submitted by buyers.

First, recall that the smallest eigenvalue $\lambda_0^2$ of the covariance matrix $\Sigma$ of $x\sim \mathcal{D}$ is greater than $0$. Since the second moment matrix $\expect[x_t x_t^\top] = \Sigma + \expect[x] \expect[x]^\top$, we know that the smallest eigenvalue of $\expect[x_t x_t^\top]$ is at least $\lambda_0^2 > 0$. 
% {Recall that the smallest eigenvalue $\lambda_0^2$ of the covariance matrix $\Sigma$ of $x\sim \mathcal{D}$ is greater than $0$, and as argued in the proof of Lemma \ref{lemma:full:bias}, we note 
% that the smallest eigenvalue of $\expect[x_t x_t^\top]$ is at least $\lambda_0^2 > 0$.}
% Since the second moment matrix $\expect[x_t x_t^\top] = \Sigma + \expect[x] \expect[x]^\top$, we know that the smallest eigenvalue of $\expect[x_t x_t^\top]$ is at least $\lambda_0^2 > 0$.
We denote the design matrix of all the features in phase $E_\ell$ as $X\in \R^{ |E_\ell|\times d}$, and $\Bar{\epsilon}_\tau = \frac{\sum_{i\in[N]}\epsilon_{i,\tau}}{N}$ for $\forall \tau \in E_\ell$. 
 
 We first consider the case where the smallest eigenvalue of the second moment matrix $\lambda_{\min}\left( X^\top  X/|E_\ell| \right) \geq \lambda_0^2/2$, which implies that $(X^\top X)^{-1}$ exists and $(X^\top X)^{-1} = (X^\top X)^\dag$.
% Later, we show that with high probability, $\lambda_{\min}\left( X^\top  X/|E_\ell|\right) \geq \lambda_0^2/2$.
By the definition $b_{i,t} = v_{i,t} - a_{i,t}$, and the definition of $\Bar{b}_\tau$ for any $\tau \in [T]$ in Equation (\ref{eq:strat:betaestimate}) we have
 \begin{align}
     \widehat{\beta}_{\ell+1} = \left(X^\top X \right)^{-1} X^\top  \left(\begin{matrix}\Bar{b}_1\\ \vdots\\ \Bar{b}_t \nonumber \\ \end{matrix}\right) &= \left(X^\top X \right)^{-1} X^\top  \left(\begin{matrix} \frac{\sum_{i\in[N]} v_{i,1} - a_{i,1}}{N}\\ \vdots\\ \frac{\sum_{i\in[N]} v_{i,t} - a_{i,t}}{N}\\ \end{matrix}\right)  \nonumber\\
     & =\beta +\left(X^\top X \right)^{-1} X^\top   \left(\begin{matrix} \frac{\sum_{i\in[N]} \epsilon_{i,1} - a_{i,1}}{N}\\ \vdots\\ \frac{\sum_{i\in[N]} \epsilon_{i,t} - a_{i,t}}{N}\\ \end{matrix}\right)  \nonumber\\
     & = \beta +\left(X^\top X \right)^{-1} X^\top \left(\bar{\mathcal{E}} - A\right)\label{eq:full_biasboundbetahat}\,,
 \end{align}
 where $\bar{\mathcal{E}}$  is the column vector consisting of all $\Bar{\epsilon}_\tau := \frac{\sum_{i\in[N]} \epsilon_{i,\tau}}{N}$, and $A$ is the column vector consisting of all $\Bar{a}_\tau := \frac{\sum_{i\in[N]} a_{i,\tau}}{N}$ for $\forall \tau \in [t]$. Therefore, 
\begin{align}\label{eq:strat:biasbound2norm}
    \norm{\widehat{\beta}_{\ell+1} - \beta}_2 ~ = ~ & \norm{\left(X^\top X \right)^{-1} X^\top \left(\bar{\mathcal{E}} - A \right) }_2 \nonumber\\
    ~ \leq ~ &  \frac{1}{|E_\ell|\lambda_0^2}\left( \norm{X^\top  \bar{\mathcal{E}}}_2+ \norm{X^\top A}_2 \right)\,.
\end{align}
Denote $X^j$ as the $j$th column of $X$, i.e. the 
$j$th row of $X^\top $ for $j = 1,2\dots d$, we now bound $\norm{X^\top  \bar{\mathcal{E}}}_2$ and $ \norm{X^\top A}_2$ separately. First, since $ \norm{X^\top  \bar{\mathcal{E}}}_2^2 = \sum_{j\in[d]}  \left| \bar{\mathcal{E}}^\top  X^j  \right|^2 $, for any $\gamma > 0$,
\begin{align}
\label{eq:strat:biasbound2norm_2}
     \bigcap_{j \in [d]} \left\{ \left| \bar{\mathcal{E}}^\top  X^j  \right|  \leq \frac{|E_\ell|\lambda_0^2\gamma }{\sqrt{d}} \right\} \subseteq \left\{ \frac{1}{|E_\ell|\lambda_0^2}\cdot  \norm{X^\top  \bar{\mathcal{E}}}_2 \leq \gamma \right\}\,.
\end{align}
We observe that $ \bar{\mathcal{E}}^\top  X^j = \frac{\sum_{\tau\in E_\ell}\sum_{i\in [N]}\epsilon_{i,\tau}X_{\tau j}}{N}$, where all $\epsilon_{i,\tau}X_{\tau j}$ are $0$-mean and ${\epsilon_{\max}} x_{\max}$-subgaussion random variables. {Therefore by Hoeffding's inequality, for any $\tilde{\gamma} > 0$}
\begin{align}
\label{eq:strat:biasbound2norm_3}
    & \prob\left( \left|N\bar{\mathcal{E}}^\top  X^j \right| \leq \Tilde{\gamma} \right) \geq 1 - 2\exp\left( -\frac{ \Tilde{\gamma}^2}{2{\epsilon_{\max}}^2x_{\max}^2 |E_\ell|N}\right) \,.
\end{align}
Replacing $\tilde{\gamma}$ with $N |E_\ell|\lambda_0^2\gamma /\sqrt{d}$ and using Equation (\ref{eq:strat:biasbound2norm_2}) yields:
\begin{align}
\label{eq:strat:biasbound2norm_4}
     \prob\left(\left\{ \frac{1}{|E_\ell|\lambda_0^2}\cdot  \norm{X^\top  \bar{\mathcal{E}}}_2 \leq \gamma \right\}\right)
    ~ \geq ~ &  \prob\left(\bigcap_{j \in [d]} \left\{ \left| \bar{\mathcal{E}}^\top  X^j  \right|  \leq \frac{|E_\ell|\lambda_0^2\gamma }{\sqrt{d}} \right\}\right) \nonumber\\
    ~\geq ~ & 1 - \sum_{j\in [d]} \prob\left( \left| \bar{\mathcal{E}}^\top  X^j  \right| > \frac{|E_\ell|\lambda_0^2\gamma }{\sqrt{d}} \right) \nonumber \\
     ~\geq ~ & 1 -  2d\exp\left( -\frac{N\gamma^2\lambda_0^4 |E_\ell| }{2{\epsilon_{\max}}^2x_{\max}^2 d}\right)\,,
\end{align}
where the first inequality follows from Equation (\ref{eq:strat:biasbound2norm_2}), the second inequality applies the union bound, and the last inequality follows from Equation (\ref{eq:strat:biasbound2norm_3}).

{In the following, we show a high probability bound for $ \norm{X^\top A}_2^2$ by using the fact that $|a_{i,t}|\leq 1/|E_\ell|$ for any $t\in E_\ell/ \mathcal{S}_{i,\ell}$, where $\mathcal{S}_{i,\ell}  = \left\{t \in E_\ell: |a_{i,t}| > {1}/{|E_\ell|} \right\}$, and $\mathcal{S}_{i,\ell} \leq L_{\ell}$ with high probability.}

Recall the event $\mathcal{G}_{i,\ell}= \left\{\left| \mathcal{S}_{i,\ell} \right| \leq L_{\ell}\right\}$, and in Lemma \ref{lemma:strat:boundedindividualbiglies} we showed that
$ \prob\left( \mathcal{G}_{i,\ell}^c\right) = \prob\left(\left| \mathcal{S}_{i,\ell}\right| > L_\ell \right) \leq \frac{1}{|E_{\ell}|}$. We now bound $ \norm{X^\top A}_2$ under the occurrence of $\mathcal{G}_{i,\ell}$ for all $i$.
\begin{align}
\label{eq:strat:biasboundnormA_1}
    \norm{X^\top A}_2^2 = \sum_{j\in[d]}  \left| A^\top X^j \right|^2 ~ = ~ & \sum_{j\in[d]} \left( \frac{\sum_{\tau\in E_\ell}\sum_{i\in [N]}a_{i,\tau}X_{\tau j}}{N}\right)^2 \nonumber \\
    ~ \leq ~ & \sum_{j\in[d]} \left(\frac{\sum_{\tau\in E_\ell}\sum_{i\in [N]}|a_{i,\tau}| x_{\max}}{N} \right)^2 \,.
\end{align}
For periods in $S_\ell := \cup_{i\in [N]} \mathcal{S}_{i,\ell} $, we have,
\begin{align}
\label{eq:strat:biasboundnormA_2}
    \frac{\sum_{\tau\in S_\ell}\sum_{i\in [N]}|a_{i,\tau}| x_{\max}}{N} ~ \leq ~ \sum_{\tau\in S_\ell}a_{\max}x_{\max} ~ \leq ~ N L_\ell a_{\max}x_{\max}
    \,,
\end{align}
where the last inequality holds because events $\mathcal{G}_{i,\ell}$ occurs for all $i$. On the other hand, recall that $|a_{i,t}|\geq 1/|E_\ell|$ for any $i$ and $t\in \mathcal{S}_{i,\ell}$. Hence, $|a_{i,t}|\leq 1/|E_\ell|$ for periods in $ E_\ell/ \mathcal{S}_\ell$, 
\begin{align}
\label{eq:strat:biasboundnormA_3}
   \frac{\sum_{\tau\in E_\ell/ \mathcal{S}_\ell}\sum_{i\in [N]}|a_{i,\tau}| x_{\max}}{N} ~ \leq ~ \sum_{\tau\in E_\ell/ \mathcal{S}_\ell}\frac{x_{\max}}{|E_{\ell}|} ~ \leq ~ \sum_{\tau\in E_\ell}\frac{x_{\max}}{|E_{\ell}|} = x_{\max}\,.
\end{align}
Combining Equations (\ref{eq:strat:biasboundnormA_1}), (\ref{eq:strat:biasboundnormA_2}), and (\ref{eq:strat:biasboundnormA_3}), we have
\begin{align}
\label{eq:strat:biasboundnormA_4}
    \norm{X^\top A}_2~ \leq ~ \sqrt{d\left(\frac{\sum_{\tau\in [t]}\sum_{i\in [N]}|a_{i,\tau}| x_{\max}}{N}\right)^2} ~ \leq ~ \sqrt{d}\left( N L_\ell a_{\max} + 1\right)x_{\max} \,.
\end{align}

Now it only remains to show $\lambda_{\min}\left(X^\top  X/|E_{\ell}|\right) \geq \lambda_0^2/2$ with high probability, which can be achieved by applying Lemma \ref{matrixchernoff}. In the context of this lemma, we consider the sequence of random matrices $\{ x_\tau x_\tau^\top / |E_{\ell}|\}_{\tau\in[E_{\ell}]}$, and note that $X^\top X/|E_{\ell}| = \sum_{\tau\in E_{\ell}} (x_\tau x_\tau^\top /|E_{\ell}|)$.  We first  upper bound the maximum eigenvalue of $x_\tau x_\tau^\top /|E_{\ell}|$, namely $\lambda_{\max}\left(x_\tau x_\tau^\top /|E_{\ell}|\right) $ for any $\tau \in E_{\ell}$ by 
\begin{align}
% \label{eq:boundmaxeigenvalue}
    \lambda_{\max}\left(\frac{x_\tau x_\tau^\top}{|E_{\ell}|} \right)  = \max_{\norm{z}_2 =1} z^\top \frac{x_\tau x_\tau^\top}{|E_{\ell}|} z \leq \frac{1}{|E_{\ell}|} \max_{\norm{z}_2 =1}(x^\top z)^2 \leq \frac{x_{\max}^2}{|E_{\ell}|} \nonumber \,.
\end{align}
 This allows us to apply the matrix Chernoff bound in Lemma \ref{matrixchernoff} (setting $\Bar{\gamma} = 1/2$ in the lemma) and get
\begin{align}
\label{eq:strat:matrixchernoff}
   \prob\left(\lambda_{\min}\left(\frac{X^\top X}{|E_{\ell}|} \right) \geq  \frac{\lambda_0^2}{2} \right) ~ \geq ~ & \prob\left(\lambda_{\min}\left(\frac{X^\top X}{|E_{\ell}|} \right)  \geq ~ \frac{1}{2}\lambda_{\min}\left(\expect\left[\frac{X^\top X}{|E_{\ell}|}\right]\right)\right) \nonumber\\
   ~ \geq ~ &  1- d\exp\left( -\frac{|E_{\ell}|\lambda_0^2}{8x_{\max}^2}\right)\,,
\end{align} 
where  the first inequality follows from the fact that $\lambda_{\min}\left(\expect[X^\top X/|E_{\ell}|]\right) \geq \lambda_0^2$.

% {Now, following the same arguments of Equation (\ref{eq:full:matrixchernoff}) in the proof of Lemma \ref{lemma:full:bias}, but by replacing $t$ with $|E_\ell|$, we have 
% \begin{align}
% \label{eq:strat:matrixchernoff}
%   \prob\left(\lambda_{\min}\left(\frac{X^\top X}{|E_\ell|} \right) \geq  \frac{\lambda_0^2}{2} \right) ~ \geq ~ 1- d\exp\left( -\frac{|E_\ell|\lambda_0^2}{8x_{\max}^2}\right)\,.
% \end{align} }
Putting everything together, we get 
\begin{align*}
     & \prob\left( \norm{\widehat{\beta}_{\ell+1} - \beta}_1  \leq \gamma + \frac{\sqrt{d}\left(NL_\ell a_{\max} + 1 \right)x_{\max}}{|E_\ell|\lambda_0^2} \right) \nonumber\\
     ~\geq ~&\prob\left( \norm{\widehat{\beta}_{\ell+1} - \beta}_2 \leq \gamma + \frac{\sqrt{d} \left(N L_\ell a_{\max} + 1 \right)x_{\max}}{|E_\ell|\lambda_0^2} \right) \\
      ~\geq ~ &
    \prob\left(\left\{ \frac{1}{|E_\ell|\lambda_0^2}\left( \norm{X^\top  \bar{\mathcal{E}}}_2+ \norm{X^\top A}_2 \right) \leq \gamma + \frac{\sqrt{d}\left(NL_\ell a_{\max} + 1 \right)x_{\max}}{|E_\ell|\lambda_0^2}\right\}  \bigcap \left\{\lambda_{\min}\left(\frac{X^\top X}{|E_\ell|} \right) \geq   \frac{\lambda_0^2}{2}\right\}\right) \\
     ~\geq ~ &
    \prob\left(\left\{ \frac{1}{|E_\ell|\lambda_0^2}\norm{X^\top  \bar{\mathcal{E}}}_2\leq \gamma\right\} ~ \bigcap ~ \left(\bigcap_{i\in [N]} \mathcal{G}_{i,\ell}\right) ~ \bigcap ~ \left\{\lambda_{\min}\left(\frac{X^\top X}{|E_\ell|}  \right) \geq   \frac{\lambda_0^2}{2}\right\}\right) \\
    %  ~ = ~ &
    % \prob\left(\left\{ \frac{1}{|E_\ell|\lambda_0^2}\norm{X^\top  \bar{\mathcal{E}}}_2\leq \gamma\right\}\right) -  \prob\left(\left\{ \frac{1}{|E_\ell|\lambda_0^2}\norm{X^\top  \bar{\mathcal{E}}}_2\leq \gamma\right\}  \bigcap \left\{\lambda_{\min}\left(\frac{X^\top X}{|E_\ell|} \right) \leq   \frac{\lambda_0^2}{2}\right\}\right) \\
     ~\geq ~ & 1 - 
    \prob\left(\left\{ \frac{1}{|E_\ell|\lambda_0^2}\norm{X^\top  \bar{\mathcal{E}}}_2 > \gamma\right\}\right)  -\sum_{i\in[N]} \prob\left(\mathcal{G}_{i,\ell}^c \right) -  \prob\left(\left\{\lambda_{\min}\left(\frac{X^\top X}{|E_\ell|} \right) \leq   \frac{\lambda_0^2}{2}\right\}\right) \\
    % ~\geq ~ &  \prob\left(\bigcap_{j \in [d]} \left\{ \left| \bar{\mathcal{E}}^\top  X^j  \right|  \leq \frac{t\lambda_0^2\gamma }{\sqrt{d}} \right\}  \bigcap \left\{\lambda_{\min}\left(\frac{X^\top X}{t} \right) \geq  \frac{\lambda_0^2}{2}\right\}\right) \\
    % ~\geq ~ & 1 - \sum_{j\in [d]} \prob\left( \left| \bar{\mathcal{E}}^\top  X^j  \right| > \frac{t\lambda_0^2\gamma }{\sqrt{d}} \right)  - \prob\left( \lambda_{\min}\left(\frac{X^\top X}{t} \right) < \frac{\lambda_0^2}{2} \right) \\
     ~\geq ~ & 1 -  2d\exp\left( -\frac{N\gamma^2\lambda_0^4 |E_\ell| }{2{\epsilon_{\max}}^2x_{\max}^2 d}\right)  - \frac{N}{|E_\ell|}-  d\exp\left( -\frac{|E_\ell|\lambda_0^2}{8x_{\max}^2}\right) \,.
\end{align*}
{The first inequality follows from the fact that $\norm{{z}}_1 \leq \norm{{z}}_2$ for any vector ${z}$; the second inequality follows from Equation (\ref{eq:strat:biasbound2norm}); the third inequality follows from Equation (\ref{eq:strat:biasboundnormA_4}) when the event $\cap_{i\in[N]}\mathcal{G}_{i,\ell} $ occurs; the fourth inequality applies a simple union bound; and the final inequality follows from Equations (\ref{eq:strat:biasbound2norm_4}), (\ref{eq:strat:matrixchernoff}) and Lemma \ref{lemma:strat:boundedindividualbiglies}.}
 \endproof

\begin{lemma}[Bounding Estimation Error in $F^-$ and $F^+$]
\label{lemma:strat:boundF-F+estimate}
Define $\tilde{\sigma}_t$ to be the sigma algebra generated by all $\{x_\tau , a_{i,\tau}, \epsilon_{i,\tau}\}_{i\in [N],\tau \in[t]} $. Then, for any $\tilde{\sigma}_t$-measurable random variable ${z}$ and $\gamma > 0$, we have 
\begin{align*}
    &  \prob\left(\left|\widehat{F}_{\ell+1}^-({z}) - F^-({z})\right|\leq 2N^2 {z}_\ell\right)
  ~ \geq ~ 1 - 4\exp\left( -2N|E_\ell|\gamma^2\right) - \frac{4(d + N)}{|E_\ell|}-  2d\exp\left( -\frac{|E_\ell|\lambda_0^2}{8x_{\max}^2}\right)\, \\
  & \prob\left(\left|\widehat{F}_{\ell+1}^+({z}) - F^+({z})\right|\leq N{z}_\ell \right)
  ~ \geq ~  1 - 4\exp\left( -2N|E_\ell|\gamma^2\right) - \frac{4(d + N)}{|E_\ell|}-  2d\exp\left( -\frac{|E_\ell|\lambda_0^2}{8x_{\max}^2}\right)\,,
\end{align*}
where $ {z}_\ell := \gamma + c_f \delta_{\ell } + {(c_f + L_\ell)}/{|E_\ell|} $,  $c_f = \sup_{\tilde{z}\in [-\epsilon_{\max},\epsilon_{\max}]} f(\tilde{z})$, $\delta_\ell$ is defined in Equation (\ref{eq:strat:defdelta}), and $L_\ell = {\log\left({v_{\max}^2N|E_{\ell}|^4} -1 \right)}/{\log({1}/{\eta})}$.
\end{lemma}
\textit{Proof of Lemma \ref{lemma:strat:boundF-F+estimate}.}
We first bound the error in the estimate of $F$, namely $\left|\widehat{F}_{\ell+1}({z}) - F({z})\right|$. Then, we use the relationship $F^-({z}) = NF^{N-1}({z}) - (N-1)F^N({z})$ and $F^+({z}) = F^N({z})$, as well as the definition of $\widehat{F}_{\ell+1}^-({z})$ and $\widehat{F}_{\ell+1}^+({z})$ in Equation (\ref{eq:strat:F-F+estimate}) to show the desired probability bounds.

% Since events $\mathcal{G}_{i,\ell}= \left\{\left| \mathcal{S}_{i,\ell}\right| \leq  L_\ell \right\}$ hold for all $i\in[N]$, we know that $|\mathcal{S}_\ell| \leq NL_\ell$.

We first upper and lower bound $\widehat{F}_{\ell+1}^-({z})$ for any $z\in \R$. Recall the event $\mathcal{S}_{i,\ell}  = \left\{t \in E_\ell: |a_{i,t}|\geq {1}/{|E_\ell|} \right\}$ and in Lemma \ref{lemma:strat:boundedindividualbiglies} we showed that $\prob\left(\left| \mathcal{S}_{i,\ell}\right| > L_\ell \right) \leq {1}/{|E_{\ell}|}$. Hence, for any $i\in[N]$, we have $|a_{i,t}|\leq 1/|E_\ell|$ for all periods $\tau \in E_\ell/ \mathcal{S}_{i,\ell}$, so 
\begin{align}
\label{eq:strat:boundF-estimate1}
        %  \widehat{F}_{\ell+1}({z}) 
        % ~ = ~ \frac{1}{|E_{\ell}|} 
        & \sum_{\tau \in E_{\ell}} \I\left\{ b_{{i},\tau} -\langle  \widehat{\beta}_{\ell+1}, x_\tau \rangle \leq {z} \right\} \nonumber \\
         ~ = ~ &  \left(\sum_{\tau \in E_\ell/ \mathcal{S}_{i,\ell} } \I\left\{ b_{{i},\tau} -\langle  \widehat{\beta}_{\ell+1}, x_\tau \rangle \leq {z} \right\} + \sum_{\tau \in \mathcal{S}_{i,\ell} } \I\left\{ v_{{i},\tau} -\langle  \widehat{\beta}_{\ell+1}, x_\tau \rangle \leq {z} \right\} \right) \nonumber \\
         & ~~ + \left(\sum_{\tau \in \mathcal{S}_{i,\ell}} \I\left\{ b_{{i},\tau} -\langle  \widehat{\beta}_{\ell+1}, x_\tau \rangle \leq {z} \right\} - \sum_{\tau \in \mathcal{S}_{i,\ell}} \I\left\{ v_{{i},\tau} -\langle  \widehat{\beta}_{\ell+1}, x_\tau \rangle \leq {z} \right\}  \right)
        \,.
     \end{align}
Consider the sum in first the parenthesis of Equation (\ref{eq:strat:boundF-estimate1}) and note that  $b_{{i},\tau}  =  v_{{i},\tau} -a_{{i},\tau} =  \langle  \beta, x_\tau \rangle + \epsilon_{{i},\tau} -a_{{i},\tau} $. Since $|a_{i,\tau}|\leq 1/|E_\ell|$ for any $i\in[N]$ and $\tau\in E_\ell/ \mathcal{S}_{i,\ell}$, 
\begin{align}
 \label{eq:strat:boundF-estimate1.5}
 \langle  \beta, x_\tau \rangle +  \epsilon_{{i},\tau}  - \frac{1}{|E_\ell|} ~ \leq ~  b_{{i},\tau} ~ \leq ~ \langle  \beta, x_\tau \rangle +  \epsilon_{{i},\tau}  + \frac{1}{|E_\ell|} , ~~~ \forall \tau \in E_\ell/ \mathcal{S}_{i,\ell}\,.
\end{align}

% Let $\mathcal{S}_\ell := \cup_{i\in [N]} \mathcal{S}_{i,\ell} $, and by definition we have $|a_{i,t}|\geq 1/|E_\ell|$ for any $i$ and $t\in \mathcal{S}_{i,\ell}$. Hence, $|a_{i,t}|\leq 1/|E_\ell|$ for periods in $E_\ell/ \mathcal{S}_{i,\ell} := E_\ell/ \mathcal{S}_{i,\ell}$. 
% , and events $\mathcal{G}_{i,\ell}= \left\{\left| \mathcal{S}_{i,\ell}\right| \leq  L_\ell \right\}$ hold for any $i\in [N]$ where $\delta_\ell$ is defined in Equation (\ref{eq:strat:defdelta}).
Now, assume that the event $\xi_{\ell+1} =\left\{\norm{\widehat{\beta}_{\ell+1} - \beta}_1 \leq {\delta_{\ell}}/{x_{\max}}\right\}$ holds. Therefore, we can upper bound the sum in first the parenthesis of Equation (\ref{eq:strat:boundF-estimate1}) as
\begin{align}
 \label{eq:strat:boundF-estimate2}
        &  \sum_{\tau \in E_\ell/ \mathcal{S}_{i,\ell} } \I\left\{  b_{{i},\tau} -\langle  \widehat{\beta}_{\ell+1}, x_\tau \rangle \leq {z} \right\} + \sum_{\tau \in \mathcal{S}_{i,\ell} } \I\left\{ v_{{i},\tau} -\langle  \widehat{\beta}_{\ell+1}, x_\tau \rangle \leq {z} \right\} \nonumber \\
          ~ \leq ~ &  \sum_{\tau \in E_\ell/ \mathcal{S}_{i,\ell}} \I\left\{ \epsilon_{{i},\tau} \leq {z} + \langle  \widehat{\beta}_{\ell+1} - \beta, x_\tau \rangle + \frac{1}{|E_\ell|} \right\} + \sum_{\tau \in \mathcal{S}_{i,\ell}} \I\left\{ \epsilon_{{i},\tau} \leq {z} + \langle  \widehat{\beta}_{\ell+1} - \beta, x_\tau \rangle + \frac{1}{|E_\ell|} \right\} \nonumber\\
            ~ = ~ &  \sum_{\tau \in E_\ell} \I\left\{ \epsilon_{{i},\tau} \leq {z} + \langle  \widehat{\beta}_{\ell+1} - \beta, x_\tau \rangle + \frac{1}{|E_\ell|} \right\}\nonumber \\
          ~ \leq ~ &  \sum_{\tau \in E_\ell} \I\left\{ \epsilon_{{i},\tau} \leq {z} + \delta_{\ell}  + \frac{1}{|E_\ell|} \right\}
        \,,
     \end{align}
 where the first equality follows from $v_{i,\tau} = \langle \beta, x_\tau\rangle + \epsilon_{i,\tau}$ and $b_{i,\tau} = v_{i,\tau} - a_{i,\tau}$; the first inequality follows Equation (\ref{eq:strat:boundF-estimate1.5}); and the final inequality is due to the occurrence of the event $\xi_{\ell+1} = \left\{\norm{\widehat{\beta}_{\ell+1} - \beta}_1 \leq {\delta_{\ell}}/{x_{\max}} \right\}$. Similarly, we can also lower bound the sum in the first parenthesis of Equation (\ref{eq:strat:boundF-estimate1}):
 \begin{align}
 \label{eq:strat:boundF-estimate3}
         \sum_{\tau \in E_\ell/ \mathcal{S}_{i,\ell} } \I\left\{ b_{{i},\tau} -\langle  \widehat{\beta}_{\ell+1}, x_\tau \rangle \leq {z} \right\} + \sum_{\tau \in \mathcal{S}_{i,\ell}} \I\left\{ b_{{i},\tau} -\langle  \widehat{\beta}_{\ell+1}, x_\tau \rangle \leq {z} \right\} 
          ~ \geq ~  \sum_{\tau \in E_\ell} \I\left\{ \epsilon_{{i},\tau} \leq {z} - \delta_{\ell} - \frac{1}{|E_\ell|} \right\}
        \,.
\end{align}
 
 Furthermore, assuming events $\mathcal{G}_{i,\ell}= \left\{\left| \mathcal{S}_{i,\ell}\right| \leq  L_\ell \right\}$ hold for all $i\in[N]$, we can simply upper bound and lower bound the expression in the second parenthesis of Equation (\ref{eq:strat:boundF-estimate1}):
% $|\mathcal{S}_{i,\ell}| \leq  L_\ell$: 
 \begin{align}
 \label{eq:strat:boundF-estimate4}
    - L_\ell ~ \leq ~ \sum_{\tau \in \mathcal{S}_{i,\ell}} \I\left\{ b_{{i},\tau} -\langle  \widehat{\beta}_{\ell+1}, x_\tau \rangle \leq {z} \right\} - \sum_{\tau \in \mathcal{S}_{i,\ell}} \I\left\{ v_{{i},\tau} -\langle  \widehat{\beta}_{\ell+1}, x_\tau \rangle \leq {z} \right\} ~ \leq ~ L_\ell  \,.
 \end{align}
 Combining Equations (\ref{eq:strat:boundF-estimate1}), (\ref{eq:strat:boundF-estimate2}), (\ref{eq:strat:boundF-estimate3}), (\ref{eq:strat:boundF-estimate4}), and using the definition
 $$\widehat{F}_{\ell+1}({z}) = \frac{1}{N|E_{\ell}|} 
  \sum_{{i} \in [N]}\sum_{\tau \in E_{\ell}} \I\left\{ b_{{i},\tau} -\langle  \widehat{\beta}_{\ell+1}, x_\tau \rangle \leq {z} \right\},$$ under the occurrence of events $\xi_{\ell+1}$, and $\mathcal{G}_{i,\ell}$ for all $i
  \in[N]$, we have
\begin{align}
 \label{eq:strat:boundF-estimate5}
   & \frac{1}{N|E_{\ell}|}  \sum_{{i} \in [N]} \sum_{\tau \in E_{\ell}} \I\left\{ \epsilon_{{i},\tau} \leq {z} - \delta_{\ell} - \frac{1}{|E_\ell|}\right\} - \frac{L_\ell}{|E_\ell|} ~ \leq ~ 
    \widehat{F}_{\ell+1}({z})  ~~ \text{and}\nonumber\\ 
   &  \widehat{F}_{\ell+1}({z}) ~ \leq ~ \frac{1}{N|E_\ell|} \sum_{{i} \in [N]}\sum_{\tau \in E_{\ell}} \I\left\{ \epsilon_{{i},\tau} ~ \leq ~ {z} +  \delta_{\ell} + \frac{1}{|E_\ell|} \right\} + \frac{L_\ell}{|E_\ell|} \,. 
\end{align}

Now, for any $\gamma > 0$,  
\begin{align}
 \label{eq:strat:boundF-estimate6}
    & \prob\left( F\left({z} - \delta_{\ell} - \frac{1}{|E_\ell|}\right) - \widehat{F}_{\ell+1}({z})
     \leq  \gamma  + \frac{L_\ell}{|E_\ell|}\right)\nonumber\\
      ~ \geq ~ & \prob\left(\left\{ F\left({z} - \delta_{\ell} - \frac{1}{|E_\ell|}\right) - \widehat{F}_{\ell+1}({z})
     \leq  \gamma  + \frac{L_\ell}{|E_\ell|} \right\} ~ \bigcap ~ \xi_{\ell+1} ~ \bigcap ~ \left(\bigcap_{i\in[N]}\mathcal{G}_{i,\ell}\right)\right)\nonumber\\
      ~ \geq ~ & \prob\left(\left\{ F\left({z} - \delta_{\ell} - \frac{1}{|E_\ell|}\right) - \frac{1}{N|E_\ell|} \sum_{{i} \in [N]} \sum_{\tau \in E_\ell} \I\left\{ \epsilon_{{i},\tau} \leq{z}  - \delta_{\ell} - \frac{1}{|E_\ell|} \right\}  \leq \gamma \right\} ~ \bigcap ~ \xi_{\ell+1} ~ \bigcap ~ \left(\bigcap_{i\in[N]}\mathcal{G}_{i,\ell}\right)\right)\nonumber\\
    % ~ \geq ~ &  \prob\left( F\left({z} - \delta_{\ell} - \frac{1}{|E_\ell|}\right) - \frac{1}{N|E_\ell|} \sum_{{i} \in [N]} \sum_{\tau \in E_\ell} \I\left\{ \epsilon_{{i},\tau} \leq{z}  - \delta_{\ell} - \frac{1}{|E_\ell|} \right\}  \leq \gamma \right) \nonumber \\
     ~ \geq ~ & \prob\left(\left\{ \sup_{\Tilde{{z}} \in \R} \left|F(\Tilde{{z}}) -  \frac{1}{N|E_\ell|}  \sum_{{i} \in [N]} \sum_{\tau \in E_\ell} \I\left\{ \epsilon_{{i},\tau} \leq \Tilde{{z}} \right\} \right| \leq \gamma \right\} ~ \bigcap ~ \xi_{\ell+1} ~ \bigcap ~ \left(\bigcap_{i\in[N]}\mathcal{G}_{i,\ell}\right)\right)\nonumber\\
%   ~ \geq ~&~  \prob\left( \sup_{\Tilde{{z}} \in \R}
% \left|F(\Tilde{{z}}) -  \frac{1}{N|E_\ell|}  \sum_{{i} \in [N]} \sum_{\tau \in E_\ell} \I\left\{ \epsilon_{{i},\tau} \leq \Tilde{{z}} \right\} \right| \leq \gamma \right) \nonumber \\
    ~ \geq ~ & 1 - \prob\left(\left\{ \sup_{\Tilde{{z}} \in \R} \left|F(\Tilde{{z}}) -  \frac{1}{N|E_\ell|}  \sum_{{i} \in [N]} \sum_{\tau \in E_\ell} \I\left\{ \epsilon_{{i},\tau} \leq \Tilde{{z}} \right\} \right| > \gamma \right\} \right) -\prob\left(\xi_{\ell+1}^{c}\right) - \sum_{i\in[N]}\prob\left(\mathcal{G}_{i,\ell}^{c}\right)\nonumber\\
    ~ \geq ~ & 1 - 2\exp\left( -2N|E_\ell|\gamma^2\right) - \left(\frac{2d + N }{|E_\ell|} +  d\exp\left( -\frac{|E_\ell|\lambda_0^2}{8x_{\max}^2}\right)\right) - \frac{N}{|E_{\ell}|} \nonumber \\
     ~ = ~ & 1 - 2\exp\left( -2N|E_\ell|\gamma^2\right) - \frac{2(d + N)}{|E_\ell|}-  d\exp\left( -\frac{|E_\ell|\lambda_0^2}{8x_{\max}^2}\right)\,,
\end{align}
where the second inequality follows from Equation (\ref{eq:strat:boundF-estimate5}), the fourth inequality uses the union bound, and the final inequality follows from the DKW inequality (Theorem \ref{DKW}), Lemma \ref{lemma:strat:bias}, and Lemma \ref{lemma:strat:boundedindividualbiglies}. We note that we can apply the DKW inequality because   $\{\epsilon_{{i},\tau}\}_{\tau\in E_\ell, i\in[N]}$ are $N |E_\ell|$ i.i.d. realizations of noise variables. According to the Lipschitz property of $F$ shown in Lemma \ref{lemma:FF-F+Lipschitz}, $|F({z}  - \delta_{\ell} - 1/|E_\ell|) - F({z})| \leq c_f(\delta_{\ell}+ 1/|E_\ell| )$ for $\forall {z} \in \R$. Hence, combining this with Equation (\ref{eq:strat:boundF-estimate6}), yields
\begin{align}
 \label{eq:strat:boundF-estimate7}
    & \prob\left( F({z}) - \widehat{F}_{\ell+1}({z}) ~ \leq ~ \gamma + c_f\left(\delta_{\ell}+\frac{1}{|E_\ell|} \right) + \frac{L_\ell}{|E_\ell|} \right)\nonumber \\ 
    ~ \geq ~ &  \prob\left(  F\left({z} - \delta_{\ell} - \frac{1}{|E_\ell|}\right) - \widehat{F}_{\ell+1}({z})
    ~ \leq ~  \gamma  + \frac{L_\ell}{|E_\ell|}\right) \nonumber\\
  ~ \geq ~ &  1 - 2\exp\left( -2N|E_\ell|\gamma^2\right) - \frac{2(d + N)}{|E_\ell|}-  d\exp\left( -\frac{|E_\ell|\lambda_0^2}{8x_{\max}^2}\right)\,.
\end{align}
Similarly, $|F({z}  + \delta_{\ell} + 1/|E_\ell|) - F({z})| \leq c_f (\delta_{\ell}+ 1/|E_\ell| )$ for $\forall z\in \R$, so we can show
\begin{align}
 \label{eq:strat:boundF-estimate8}
    & \prob\left(\widehat{F}_{\ell+1}({z}) - F({z})~ \leq ~ \gamma + c_f  \left(\delta_{\ell}+\frac{1}{|E_\ell|} \right) + \frac{L_\ell}{|E_\ell|} \right) \nonumber \\ 
    ~ \geq ~ &  \prob\left(  \widehat{F}_{\ell+1}({z}) -   F\left({z}+  \delta_{\ell} + \frac{1}{|E_\ell|}\right)
  ~ \leq ~ \gamma  + \frac{L_\ell}{|E_\ell|}\right) \nonumber\\
  ~ \geq ~ &  1 - 2\exp\left( -2N|E_\ell|\gamma^2\right) - \frac{2(d + N)}{|E_\ell|}-  d\exp\left( -\frac{|E_\ell|\lambda_0^2}{8x_{\max}^2}\right)\,.
\end{align}
Combining Equations (\ref{eq:strat:boundF-estimate7}) and (\ref{eq:strat:boundF-estimate8}) using a union bound yields
\begin{align}
 \label{eq:strat:boundF-estimate9}
    & \prob\left(\left|\widehat{F}_{\ell+1}({z}) - F({z})\right|\leq \gamma + c_f \delta_{\ell } + \frac{c_f + L_\ell}{|E_\ell|} \right) \nonumber \\
  ~ \geq ~ &  1 - 4\exp\left( -2N|E_\ell|\gamma^2\right) - \frac{4(d + N)}{|E_\ell|}-  2d\exp\left( -\frac{|E_\ell|\lambda_0^2}{8x_{\max}^2}\right) \,.
\end{align}

Finally, we now bound $|\widehat{F}_t^-(z) - F^-(z)|$ and $|\widehat{F}_t^+(z) - F^+(z)|$ using the fact that $F^-({z}) = NF^{N-1}({z}) - (N-1)F^N({z})$ and $F^+({z}) = F^N({z})$.
 \begin{align} 
  \label{eq:strat:boundF-estimate10}
      |\widehat{F}_{\ell+1}^-({z}) - F^-({z})| 
    ~ = ~ &  \left|N\widehat{F}_{\ell+1}^{N-1}(z) - (N-1)\widehat{F}_{\ell+1}^N(z) - \left(NF^{N-1}({z}) - (N-1)F^N({z})\right)\right| \nonumber \\
     ~ \leq ~ &  N\left|\widehat{F}_{\ell+1}^{N-1}({z}) - F^{N-1}({z}) \right| + (N-1)\left| \widehat{F}_{\ell+1}^N({z}) - F^N({z})\right| \nonumber \\
    ~ = ~ &  N\left|\left(\widehat{F}_{\ell+1} ({z}) - F({z})\right)\left(\sum_{n = 1}^{N-1} \left(\widehat{F}_{\ell+1}({z})\right)^{n-1}\left(F({z})\right)^{N -1 - n} \right) \right|\nonumber \\
    & ~~~ + (N-1)\left|\left(\widehat{F}_{\ell+1} ({z}) - F({z})\right)\left(\sum_{n = 1}^{N} \left(\widehat{F}_{\ell+1}({z})\right)^{n-1}\left(F({z})\right)^{N - n} \right) \right| \nonumber \\ 
    ~ \leq ~ & N(N-1) \left|\widehat{F}_{\ell+1}({z}) - F({z}) \right| + (N-1)N\left |\widehat{F}_{\ell+1}({z}) - F({z})\right| \nonumber\\
    ~ < ~ & 2N^2 \left|\widehat{F}_{\ell+1}({z}) - F({z})\right| \,.
\end{align}
 The second equality uses $a^m - b^m = (a-b)\left(\sum_{n = 1}^m a^{n-1} b^{m - n} \right)$ for any integer $m \geq 2$. The second inequality follows from $\widehat{F}_{\ell+1}({z}), F({z}) \in [0,1]$ for $\forall {z} 
 \in \R$. Combining Equations (\ref{eq:strat:boundF-estimate9}) and (\ref{eq:strat:boundF-estimate10}), we get 
\begin{align*}
    & \prob\left(\left|\widehat{F}_{\ell+1}^-({z}) - F^-({z})\right|\leq 2N^2\left(\gamma + c_f \delta_{\ell } + \frac{c_f + L_\ell}{|E_\ell|} \right) \right) \nonumber \\
  ~ \geq ~ & 1 - 4\exp\left( -2N|E_\ell|\gamma^2\right) - \frac{4(d + N)}{|E_\ell|}-  2d\exp\left( -\frac{|E_\ell|\lambda_0^2}{8x_{\max}^2}\right)\,.
\end{align*}
 The probability bound for $\left|\widehat{F}_{\ell+1}^-({z}) - F^-({z})\right|$ can be shown in a similar fashion by noting that similar to Equation (\ref{eq:strat:boundF-estimate10}) we can show $ |\widehat{F}_{\ell+1}^+({z}) - F^+({z})| < N\left|\widehat{F}_{\ell+1}({z}) - F({z})\right|$.
\endproof

 \begin{lemma}[Bounding the Impact of Estimation Errors on Revenue]  \label{lemma:strat:controlUncert}
We assume that the events $\xi_{\ell+1} = \left\{\norm{\widehat{\beta}_{\ell+1} - \beta}_1 \leq \frac{\delta_{\ell}}{x_{\max}} \right\}$, $\xi_{\ell+1}^- = \left\{\left|\widehat{F}_{\ell+1}^-({z}) - F^-({z})\right|\leq 2N^2\left(\gamma_{\ell} + c_f \delta_{\ell } + \frac{c_f + L_\ell}{|E_\ell|} \right) \right\}$ and $\xi_{\ell+1}^+ = \left\{\left|\widehat{F}_{\ell+1}^+({z}) - F^+({z})\right|\leq N\left(\gamma_{\ell} + c_f \delta_{\ell } + \frac{c_f + L_\ell}{|E_\ell|} \right)\right\}$ occur for some phase $\ell \geq 1$, where ${z} \in \R$, $\gamma_{\ell} =  \sqrt{\log(|E_\ell|)}/\sqrt{2N|E_\ell|}$, and $\delta_{\ell}$ is defined in Equation (\ref{eq:strat:defdelta}). Hence for any ${r} \in \{r_t^\star,~ r_t\}$ where $t\in E_{\ell+1}$ we have the following:
\begin{enumerate}
    \item [(i)] $\left|\rho_t(r, y_t, F^-, F^+) - \rho_t(r, \widehat{y}_t, F^-, F^+)\right| ~ \leq ~ 3r c_f N^2  \delta_{\ell}   $ ~~ a.s. 
    \item [(ii)] $\left|\rho_t(r, \widehat{y}_t , F^-, F^+) - \rho_t(r, \widehat{y}_t, \widehat{F}_{\ell+1}^-, \widehat{F}_{\ell+1}^+) \right|~ \leq ~  3rN^2\left(\gamma_{\ell} + c_f \delta_{\ell } + \frac{c_f + L_\ell}{|E_\ell|} \right)$ ~~  a.s.
\end{enumerate}
where $y_t = \langle \beta, x_t \rangle$, $\widehat{y}_t = \langle  \widehat{\beta}_{\ell+1}, x_t \rangle$, $\widehat{\beta}_{\ell+1},\widehat{F}_{\ell+1}^-, \widehat{F}_{\ell+1}^+$ are defined in Equations (\ref{eq:strat:betaestimate}) and (\ref{eq:strat:F-F+estimate}).  The function $\rho_t$ is defined in Equation (\ref{eq:defrho}).
\end{lemma}

\textit{Proof of Lemma \ref{lemma:strat:controlUncert}.}
\textbf{Part (i)} We consider the following:
 \begin{align*}
     &\left|\rho_t(r, y_t, F^-, F^+) - \rho_t(r, \widehat{y}_t, F^-, F^+)\right| \\
     ~ = ~ & \left| \int_{0}^r \left[F^-({z}-y_t )-F^-({z}- \widehat{y}_t ) \right]d{z} - r \left[ F^+(r-y_t)-F^+(r- \widehat{y}_t)\right] \right| \\
     ~ \leq ~  &  \int_{0}^r \left|F^-({z}-y_t )-F^-({z}- \widehat{y}_t ) \right| d{z} + r \left| F^+(r-y_t)-F^+(r- \widehat{y}_t)\right| \\
     ~ \leq ~  &  \int_{0}^r 2c_f N^2 |y_t - 
     \widehat{y}_t| d{z} + r c_f N|y_t - 
     \widehat{y}_t|\\
     ~ \leq ~  &  \int_{0}^r 2c_f N^2 \left(\norm{\widehat{\beta}_{\ell+1} - \beta}_1 x_{\max}\right)d{z} + r c_f N \norm{\widehat{\beta}_{\ell+1} - \beta}_1 x_{\max } \\
     ~ \leq ~  & 3r c_f N^2  \delta_{\ell} \,.
 \end{align*} 
 The first equality follows from definition of $\rho_t$ in Equation (\ref{eq:defrho}), and the second inequality applies the Lipschitz property of $F^-$ and $F^+$ using Lemma \ref{lemma:FF-F+Lipschitz}.
 The third inequality follows from Cauchy's inequality: $|y_t - \widehat{y}_t| = |\langle \widehat{\beta}_{\ell+1} - \beta, x_t \rangle| \leq \norm{\widehat{\beta}_{\ell+1} - \beta}_1 x_{\max}$, and the last inequality follows from the occurrence of $\xi_{\ell+1}$ and $N\geq 1$.

\textbf{Part (ii)} Similar to part (i), we have 
 \begin{align*}
     &\left|\rho_t(r,\widehat{y}_t, F^-, F^+) - \rho_t(r, \widehat{y}_t, \widehat{F}_{\ell+1}^-, \widehat{F}_{\ell+1}^+)\right| \\
     ~ = ~ & \left| \int_{0}^r \left[F^-({z}-\widehat{y}_t )- \widehat{F}_{\ell+1}^-({z}- \widehat{y}_t ) \right]d{z} - r \left[ F^+(r-\widehat{y}_t)- \widehat{F}_{\ell+1}^+(r- \widehat{y}_t)\right] \right| \\
     ~ \leq ~ & \int_{0}^r \left|F^-({z}-\widehat{y}_t )- \widehat{F}_{\ell+1}^-({z}- \widehat{y}_t ) \right| d{z} + r \left| F^+(r-\widehat{y}_t)- \widehat{F}_{\ell+1}^+(r- \widehat{y}_t)\right| \\
    ~ \leq ~ &  3rN^2\left(\gamma_{\ell} + c_f \delta_{\ell } + \frac{c_f + L_\ell}{|E_\ell|} \right)\,,
    % ~ \leq ~ &  3rN^2\left(\gamma_{\ell} + c_f \delta_{\ell } + \frac{c_f + L_\ell}{|E_\ell|} \right)
    % \,,
 \end{align*}
 where the last inequality follows from the occurrence of events $\xi_{\ell+1}^-$ and $\xi_{\ell+1}^+$ and $N\geq 1$. 
\endproof

\begin{lemma}[Bounding probabilities]  \label{lemma:strat:boundprobabilities1}
The probability that not all events $\xi_{\ell+1}$, $\xi_{\ell+1}^-$ and $\xi_{\ell+1}^+ $ occur for some phase $\ell \geq 1$ is bounded as
\begin{align*}
    \prob\left(\xi_{\ell+1}^c \cup \left(\xi_{\ell+1}^-\right)^c \cup \left(\xi_{\ell+1}^+\right)^c \right) ~ \leq ~ \frac{9N + 15d + 8}{|E_\ell|}\,,
\end{align*}
where the events $\xi_{\ell+1}$, $\xi_{\ell+1}^-$ and $\xi_{\ell+1}^+ $ are defined in Equations (\ref{eq:strat:defxi}), (\ref{eq:strat:defxi-}), and (\ref{eq:strat:defxi+}) respectively.
\end{lemma}
\textit{Proof of Lemma \ref{lemma:strat:boundprobabilities1}.}
% First of all, recall $\mathcal{G}_{i,\ell} = \left\{\left| \mathcal{S}_{i,\ell}\right| \leq L_\ell  \right\}$ and define $\mathcal{G}_{\ell}:= \cap_{i\in[N]} \mathcal{G}_{i,\ell}$. By Lemma \ref{lemma:strat:boundedindividualbiglies}, we have

% \begin{align}
% \label{eq:strat:eventsnotoccur0}
%     \prob\left(\mathcal{G}_{\ell}^c \right) ~ \leq ~ \sum_{i\in[N]}\prob\left(\mathcal{G}_{i,\ell}^c \right) ~ = ~ \sum_{i\in[N]}\prob\left( \left|\mathcal{S}_{i,\ell}\right| > L_\ell \right) ~ \leq ~ \frac{N}{|E_\ell|}\,.
% \end{align}

We first bound the probability of $\xi_{\ell+1}^c$, and then proceed to bound the   the probability of $ \left(\xi_{\ell+1}^-\right)^c$ and $\left(\xi_{\ell+1}^+\right)^c$.

Recall that $\xi_{\ell+1} = \left\{\norm{\widehat{\beta}_{\ell+1} - \beta}_1 \leq \frac{\delta_{\ell}}{x_{\max}} \right\}$. Then,
\begin{align}
\label{eq:strat:eventsnotoccur1}
     \prob\left( \xi_{\ell+1}^c \right) ~ \leq ~ &
      \frac{2d + N }{|E_\ell|} + d\exp\left( -\frac{|E_\ell|\lambda_0^2}{8x_{\max}^2}\right) \nonumber\\
    %  \prob\left( \xi_{\ell+1}^c \cap  \mathcal{G}_{\ell} \right) + \prob\left(\mathcal{G}_{\ell}^c \right) \nonumber\\
    % ~ \leq ~ & 2d\exp\left(- \frac{N\left(\frac{\sqrt{2d\log(|E_\ell|)}\epsilon_{\max} x_{\max}}{\lambda_0^2\sqrt{N|E_\ell|}}\right)^2 | E_\ell|\lambda_0^4}{2d\epsilon_{\max}^2 x_{\max}^2}\right) +  d\exp\left( -\frac{|E_\ell|\lambda_0^2}{8x_{\max}^2}\right)  + \frac{N}{|E_\ell|} \nonumber\\
    ~ \leq ~ &   \frac{2d + N }{|E_\ell|} +  d\exp\left( -\frac{\log(|E_\ell|) T^{\frac{1}{4}}\lambda_0^2}{8x_{\max}^2}\right) \nonumber\\
    ~ \leq ~ &  \frac{N + 3d}{|E_\ell|} 
    \,,
\end{align}
where the first inequality follows from Lemma \ref{lemma:strat:bias} by taking $\gamma = {\sqrt{2d\log(|E_\ell|)}\epsilon_{\max} x_{\max}}/\left({\lambda_0^2\sqrt{N|E_\ell|}}\right)$; the second inequality uses the fact that $|E_\ell| \geq |E_1| = \sqrt{T}$, ~ $T \geq \max\left \{ \left(\frac{8x_{\max}^2}{\lambda_0^2}\right)^4, 9 \right\}$, which implies $|E_\ell| \geq \log(|E_\ell|) \sqrt{|E_\ell|} \geq T^{\frac{1}{4}} \log(|E_\ell|) $. Note that here we used the fact that $\sqrt{x} \geq \log(x)$ for all $x\geq 9$.

We now bound the probability of $\left(\xi_{\ell+1}^-\right)^c$:
\begin{align}\label{eq:strat:eventsnotoccur2}
     \prob\left(\left(\xi_{\ell+1}^-\right)^c \right) ~ \leq ~ & 
    %  4\exp\left( -2N|E_\ell|\gamma^2\right) + \frac{2(d + N)}{|E_\ell|} +  2d\exp\left( -\frac{|E_\ell|\lambda_0^2}{8x_{\max}^2}\right)\nonumber \\
     4\exp\left( -2N|E_\ell|\cdot \left(\frac{\sqrt{\log(|E_\ell|)}}{\sqrt{2N|E_\ell|}}\right)^2\right) + \frac{4(d + N)}{|E_\ell|} +  2d\exp\left( -\frac{|E_\ell|\lambda_0^2}{8x_{\max}^2}\right) \nonumber\\
    %  \prob\left( \left(\xi_{\ell+1}^-\right)^c \cap \xi_{\ell+1} \cap  \mathcal{G}_{\ell} \right) + \prob\left( \xi_{\ell+1}^c \right) + \prob\left( \mathcal{G}_{\ell}^c \right)\nonumber\\
    % ~ \leq ~ & 4\exp\left( -2N|E_\ell|\cdot \left(\frac{\sqrt{\log(|E_\ell|)}}{\sqrt{2N|E_\ell|}}\right)^2\right) + \frac{N + 3d}{|E_\ell|}  + \frac{N}{|E_\ell|} \nonumber\\
    ~ \leq ~ &  \frac{2(2N + 3d + 2)}{|E_\ell|} 
    \,,
\end{align}
where the first inequality follows from Lemma \ref{lemma:strat:boundF-F+estimate} by taking $\gamma = \gamma_\ell = {\sqrt{\log(|E_\ell|)}}/{\sqrt{2N|E_\ell|}}$, and the last inequality again uses the fact that $|E_\ell| \geq \log(|E_\ell|) \sqrt{|E_\ell|} \geq T^{\frac{1}{4}} \log(|E_\ell|) $ when $T \geq \max\left \{ \left(\frac{8x_{\max}^2}{\lambda_0^2}\right)^4, 9 \right\}$ .

Similarly, we can bound the probability of $\left(\xi_{\ell+1}^+\right)^c$:
\begin{align}\label{eq:strat:eventsnotoccur3}
     \prob\left(\left(\xi_{\ell+1}^+\right)^c \right)
    ~ \leq ~ &  \frac{2(2N + 3d + 2)}{|E_\ell|} \,,
\end{align}

Finally, combining Equations (\ref{eq:strat:eventsnotoccur1}), (\ref{eq:strat:eventsnotoccur2}) and (\ref{eq:strat:eventsnotoccur3}), we have 
\begin{align}
    \prob\left(\xi_{\ell+1}^c \cup \left(\xi_{\ell+1}^-\right)^c \cup \left(\xi_{\ell+1}^+\right)^c \right) ~ \leq ~  \prob\left(\xi_{\ell+1}^c\right) +  \prob\left( \left(\xi_{\ell+1}^-\right)^c  \right) +  \prob\left(\left(\xi_{\ell+1}^+\right)^c \right) ~ \leq ~ \frac{9N + 15d + 8}{|E_\ell|} \nonumber\,.
\end{align}
\endproof

\begin{lemma}[Lipschitz Property for $F$, $F^-$ and $F^+$]\label{lemma:FF-F+Lipschitz}
The following hold for any ${z}_1, {z}_2 \in \R$:
\begin{enumerate}
    \item [(i)]  $|F({z}_1) -  F({z}_2)| \leq c_f |{z}_1-{z}_2 |$.
    \item [(ii)] $|F^-({z}_1) -  F^-({z}_2)| \leq 2c_f N^2|{z}_1-{z}_2 |$.
    \item [(iii)]$|F^+({z}_1) -  F^+({z}_2)| \leq c_f N|{z}_1-{z}_2 |$.
\end{enumerate}
Here,  $0<c_f = \sup_{{z}\in [-\epsilon_{\max},\epsilon_{\max}]}f(z)$.
\end{lemma}

\textit{Proof of Lemma \ref{lemma:FF-F+Lipschitz}.}
Without loss of generality, we assume ${z}_1 < {z}_2$. Note that $F({z}) = 0$ for $\forall {z}\in (-\infty, -\epsilon_{\max}]$, and $F({z}) = 1$ for $\forall {z}\in [\epsilon_{\max}, \infty)$.

\textbf{Part (i)}
We consider the following cases:
\begin{enumerate}\label{eq:LipschitzF}
    \item [Case 1:]  $\left({z}_1 < {z}_2 \leq - \epsilon_{\max} ~ \text{ or } ~ \epsilon_{\max} \leq {z}_1 < {z}_2 \right)$: $|F({z}_2) - F({z}_1)| = 0 \leq c_f|{z}_2 - {z}_1|$.
     \item [Case 2:]  $\left(-\epsilon_{\max} < {z}_1 < {z}_2 < \epsilon_{\max}\right)$: By the mean value theorem,  $|F({z}_2) - F({z}_1)| = f({\tilde{z}}) |{z}_2 - {z}_1| < c_f |{z}_2 - {z}_1|$, where $\tilde{z} \in ({z}_1, {z}_2)$.
    \item [Case 3:]  $\left( {z}_1 \leq -\epsilon_{\max} < {z}_2 < \epsilon_{\max}\right)$: We have $|{z}_2 - (-\epsilon_{\max})| = {z}_2  - (-\epsilon_{\max}) \leq {z}_2 - {z}_1$ and  $F({z}_1) = F(-\epsilon_{\max}) = 0$. Hence $|F({z}_2) - F({z}_1)| = |F({z}_2) - F(-\epsilon_{\max})| = f({\tilde{z}}) |{z}_2 - (-\epsilon_{\max})| \leq c_f|{z}_2 - {z}_1|$, where $\tilde{z} \in (-\epsilon_{\max}, {z}_2)$ by the mean value theorem. 
    \item [Case  4]  $\left(-\epsilon_{\max} < {z}_1 <  \epsilon_{\max} \leq {z}_2 \right)$: We have $|\epsilon_{\max}- {z}_1| = \epsilon_{\max}- {z}_1 \leq  {z}_2 - {z}_1$ and $F({z}_2) = F(\epsilon_{\max}) =1 $ . Hence $|F({z}_2) - F({z}_1)| = |F(\epsilon_{\max}) - F({z}_1) | = f({\tilde{z}}) |\epsilon_{\max}- {z}_1|  \leq c_f |{z}_2 - {z}_1|$, where $\tilde{z} \in ({z}_1, \epsilon_{\max})$ by the mean value theorem. 
\end{enumerate}

\textbf{Part (ii) \& (iii)}
We recall that $F^-({z}) = NF^{N-1}({z}) - (N-1)F^N({z})$ and $F^+({z}) = F^N({z})$, so
 \begin{align} 
    &  |F^-({z}_2) - F^-({z}_1)| \nonumber \\
    ~ = ~ &  \left|NF^{N-1}({z}_2)- (N-1)F^{N}({z}_2)- \left(NF^{N-1}({z}_1)- (N-1)F^{N}({z}_1)\right)\right| \nonumber \\
     ~ \leq ~ &  N\left|F^{N-1}({z}_2) - F^{N-1}({z}_1) \right| + (N-1)\left| F^{N}({z}_2) -F^{N}({z}_1)\right| \nonumber \\
    ~ = ~ &  N\left|\left(F({z}_2) - F({z}_1)\right)\left(\sum_{n = 1}^{N-1} \left(F({z}_2)\right)^{n-1}\left(F({z}_1)\right)^{N -1 - n} \right) \right|\nonumber \\
    & ~~~ + (N-1)\left|\left(F({z}_2) - F({z}_1)\right)\left(\sum_{n = 1}^{N} \left(F({z}_2)\right)^{n-1}\left(F({z}_1)\right)^{N - n} \right) \right| \nonumber \\ 
    ~ \leq ~ & N(N-1) \left|F({z}_2) - F({z}_1) \right| + (N-1)N\left |F({z}_2) - F({z}_1)\right| \nonumber\\
    ~ < ~ & 2N^2 c_f |{z}_2 - {z}_1| \nonumber \,.
\end{align}
 The second equality uses $a^m - b^m = (a-b)\left(\sum_{n = 1}^m a^{n-1} b^{m - n} \right)$ for any $a,b \in \R$ and integer $m \geq 2$. The second inequality follows from $F({z}) \in [0,1]$ for $\forall {z} 
 \in \R$. The final inequality follows from the Lipschitz property of $F$ shown in part (i). Following the same arguments, we can also show that $|F^+({z}_2) - F^+({z}_1)| \leq  c_f N|{z}_2 - {z}_1|$. 
 \endproof

\section{Supplementary Lemmas}
% \begin{lemma}[Lemma 2.1 in \cite{cesa2015regret}]\label{banditbound:F+}
% For $\forall x\in [0,1]$ and $\forall y\in (0,1)$, if $|G(x) - y| \leq \eta $ for some $\eta > 0$, then $ |x- G^{-1}(y)|\leq \frac{2\eta}{\sqrt{1-y}}$.
% \end{lemma}
% \begin{remark}
% Lemma \ref{banditbound:F+} allows us to bound the error in this estimate of $F^+$ using our estimation error for $F^-$ via considering $x = F^+(\cdot) = G^{-1}(F^-(\cdot))$ and $y = \widehat{F}_t^-(\cdot)$:
% \end{remark}

\begin{lemma}[Dvoretzky-Kiefer-Wolfowitz Inequality (\cite{dvoretzky1956asymptotic})]\label{DKW}
Let $Z_1,Z_2,\dots Z_n$ be i.i.d. random variables with cumulative distribution function $F$, and denote the associated empirical distribution function as 
\begin{align}
    \widehat{F}({z}) = \frac{1}{n}\sum_{i=1}^n\I\{Z_i\leq z\} \quad, z\in \mathbb{R}\,.
\end{align}
Then, for any $\Bar{\gamma} > 0$, 
\begin{align}
    \prob\left(\sup_{z\in \R}\left| \widehat{F}(z) - F(z)\right| \leq \Bar{\gamma}\right) \geq 1- 2\exp\left(-2n\Bar{\gamma}^2 \right)\,.
\end{align}
\end{lemma}

\begin{lemma}[Matrix Chernoff Bound (\cite{tropp2015introduction})]\label{matrixchernoff}
Consider a finite sequence of independent, random matrices $\{Z_k \in \R^d\}_{k\in [K]}$. Assume that $0\leq \lambda_{\min}(Z_k) $ and $ \lambda_{\max}(Z_k)\leq B $ for any $k$. Denote $Y = \sum_{k\in[K]} Z_k$, $\mu_{\min} = \lambda_{\min}(\expect[Y])$, and $\mu_{\max} = \lambda_{\max}(\expect[Y])$. Then for $\forall \Bar{\gamma} \in (0,1)$,
\begin{align*}
    \prob\left( \lambda_{\min}(Y) \leq \Bar{\gamma} \mu_{\min} \right) \leq d \exp\left(-\frac{(1-\Bar{\gamma})^2 \mu_{\min}}{2B} \right) \,.
\end{align*}
\end{lemma}

 \begin{lemma} [Multiplicative Azuma Inequality(\cite{koufogiannakis2014nearly})]
\label{multiplicativeAzuma}
Let $Z_1 = \sum_{\tau \in [\widetilde{T}]} z_{1,\tau}$ and $Z_2 = \sum_{\tau \in [\widetilde{T}]} z_{2,\tau}$ be sums of non-negative random variables, where $\widetilde{T}$ is a random stopping time with a finite expectation, and, for all $\tau \in [\widetilde{T}]$, $|z_{1,\tau} - z_{2,\tau}| \leq 1$ and $\expect\left[(z_{1,\tau} - z_{2,\tau} )~ \big|~ \sum_{s < \tau} z_{1,s}, \sum_{s<\tau} z_{2,s}\right]\leq 0$. Let $\Tilde{\gamma} \in [0,1]$ and $A \in \R$. Then,
\begin{align*}
    \prob\left((1-\Tilde{\gamma}) Z_1 \geq Z_2 + A \right) \leq \exp\left(-\Tilde{\gamma} A\right)
\end{align*}
\end{lemma}
\end{APPENDICES}

\end{document}